\documentclass[10pt,journal,compsoc]{IEEEtran}

\usepackage{enumitem}
\usepackage{bm}

\ifCLASSOPTIONcompsoc
  \usepackage[nocompress]{cite}
\else
  \usepackage{cite}
\fi

\usepackage{graphicx}
\usepackage{amsmath}
\usepackage{amssymb}
\usepackage{bbm}

\usepackage{algorithmic}
\usepackage{algorithm}
\usepackage{makecell}
\usepackage{tikz}
\usepackage{forest}
\usetikzlibrary{trees,positioning,shapes,shadows,arrows.meta}

\ifCLASSOPTIONcompsoc
 \usepackage[caption=false,font=footnotesize,labelfont=sf,textfont=sf]{subfig}
\else
 \usepackage[caption=false,font=footnotesize]{subfig}
\fi

\usepackage{url}

\usepackage{booktabs} % for table
\usepackage{multirow} % for table
\usepackage{xspace}
\usepackage{xcolor}

\usepackage{hyperref}
\hypersetup{hypertex=true,
            colorlinks=true,
            linkcolor=black,
            anchorcolor=black,
            citecolor=black}

\newenvironment{packed_itemize}{
	\begin{itemize}[leftmargin=*]
		\setlength{\itemsep}{3pt}
		\setlength{\parskip}{-0.5pt}
		\setlength{\parsep}{-8pt}
	}{\end{itemize}}

\definecolor{ngreen}{HTML}{D5E8D4}
\definecolor{nblue}{HTML}{DAE8FC}
\definecolor{npurple}{HTML}{E1D5E7}

\begin{document}

\title{A Survey on Efficient Inference for Large Language Models}

% Authors must not appear in the submitted version. They should be hidden
% as long as the tmlr package is used without the [accepted] or [preprint] options.
% Non-anonymous submissions will be rejected without review.

\author{Zixuan Zhou*,
  Xuefei Ning*,
  Ke Hong*,
  Tianyu Fu,
  Jiaming Xu,
  Shiyao Li,\\
  Yuming Lou,
  Luning Wang,
  Zhihang Yuan,
  Xiuhong Li,
  Shengen Yan,
  Guohao Dai,\\
  Xiao-Ping Zhang~\IEEEmembership{Fellow,~IEEE},
  Huazhong Yang~\IEEEmembership{Fellow,~IEEE},
  Yuhan Dong,
  Yu~Wang~\IEEEmembership{Fellow,~IEEE}% <-this % stops a space
\IEEEcompsocitemizethanks{\IEEEcompsocthanksitem Z. Zhou, K. Hong, T. Fu, S. Li, L. Wang are with Infinigence-AI and the Department
of Electronic Engineering, Tsinghua University, China.\\
E-mail: zhouzx21@mails.tsinghua.edu.cn (Z. Zhou)
\IEEEcompsocthanksitem X. Ning, Y. Lou, H. Yang, Y. Wang are with the Department of Electronic Engineering, Tsinghua University, China. \\
E-mail: foxdoraame@gmail.com (X. Ning), yu-wang@tsinghua.edu.cn (Y. Wang)
\IEEEcompsocthanksitem J. Xu, G. Dai are with 
Infinigence-AI and the Department of Electronic Engineering, Shanghai Jiaotong University, China.\\
E-mail: daiguohao@sjtu.edu.cn (G. Dai)
\IEEEcompsocthanksitem X.-P. Zhang, Y. Dong are with Tsinghua Shenzhen International Graduate School. \\
E-mail: xpzhang@ieee.org (X.-P. Zhang), dongyuhan@sz.tsinghua.edu.cn (Y. Dong)
\IEEEcompsocthanksitem Z. Yuan, S. Yan are with Infinigence-AI. 
\IEEEcompsocthanksitem X. Li is with Peking University.
\IEEEcompsocthanksitem Corresponding authors: Yu Wang, Xuefei Ning, Guohao Dai.
\IEEEcompsocthanksitem *Equal contribution.
}}% <-this % stops an unwanted space
%\thanks{$^*$ Corresponding author.}}

% The \author macro works with any number of authors. Use \AND 
% to separate the names and addresses of multiple authors.

\newcommand{\fix}{\marginpar{FIX}}
\newcommand{\new}{\marginpar{NEW}}

\def\month{MM}  % Insert correct month for camera-ready version
\def\year{YYYY} % Insert correct year for camera-ready version
\def\openreview{\url{https://openreview.net/forum?id=XXXX}} % Insert correct link to OpenReview for camera-ready version

\maketitle

\begin{abstract}
%or the serving systems that require low latency and high throughput. 
Large Language Models (LLMs) have attracted extensive attention due to their remarkable performance across various tasks. 
However, the substantial computational and memory requirements of LLM inference pose challenges for deployment in resource-constrained scenarios.
Efforts within the field have been directed towards developing techniques aimed at enhancing the efficiency of LLM inference. This paper presents a comprehensive survey of the existing literature on efficient LLM inference.
We start by analyzing the primary causes of the inefficient LLM inference, i.e., the large model size, the quadratic-complexity attention operation, and the auto-regressive decoding approach. Then, we introduce a comprehensive taxonomy that organizes the current literature into data-level, model-level, and system-level optimization.
Moreover, the paper includes comparative experiments on representative methods within critical sub-fields to provide quantitative insights. Last but not least, we provide some knowledge summary and discuss future research directions.
%Compared with the existing surveys, our survey encompasses a broader level and scope of research topics and studies in the field of efficient LLM inference, especially some of the most advanced sub-fields, such as output organization techniques and new architecture design. 

%Moreover, our survey provides in-depth analysis and practical suggestions for users for some critical optimization techniques with extensive experiments and results. 
% more level
% We hope our survey can not only offer instructions for practitioners to deploy effective LLMs but also serve as a roadmap for researchers to conduct further studies.

\end{abstract}

% \newpage
% \tableofcontents
% \newpage

%[done] \todo{@zzx memory cost->usage}
\section{Introduction}

% introduce background, multiple optimization level

%\subsection{Motivation}

Large Language Models (LLMs) have garnered substantial attention from both academia and industry in recent years. The field of LLMs has experienced notable growth and significant achievements. Numerous open-source LLMs have emerged, including the GPT-series (GPT-1~\cite{radford2018improving}, GPT-2~\cite{radford2019language}, and GPT-3~\cite{brown2020language}), OPT~\cite{zhang2022opt}, LLaMA-series (LLaMA~\cite{touvron2023llama}, LLaMA 2~\cite{touvron2023llama}, Baichuan 2~\cite{yang2023baichuan}, Vicuna~\cite{chiang2023vicuna}, LongChat~\cite{li2023long}), BLOOM~\cite{workshop2022bloom}, FALCON~\cite{almazrouei2023falcon}, GLM~\cite{du2021glm}, and Mistral~\cite{jiang2024mixtral}, which are used for both academic research and commercial purposes. The success of LLMs stems from their robust capability in handling diverse tasks such as neural language understanding (NLU), neural language generation (NLG), reasoning~\cite{yang2023harnessing,wei2022chain}, and code generation~\cite{chen2021evaluating}, consequently enabling impactful applications like ChatGPT, Copilot, and Bing. There is a growing belief~\cite{bubeck2023sparks} that the rise and achievements of LLMs signify a significant stride towards Artificial General Intelligence (AGI) for humanity.
%\todo{give more description of the impacts of LLMs}

\begin{figure}[htb]
    \centering
    \includegraphics[width=1.0\linewidth]{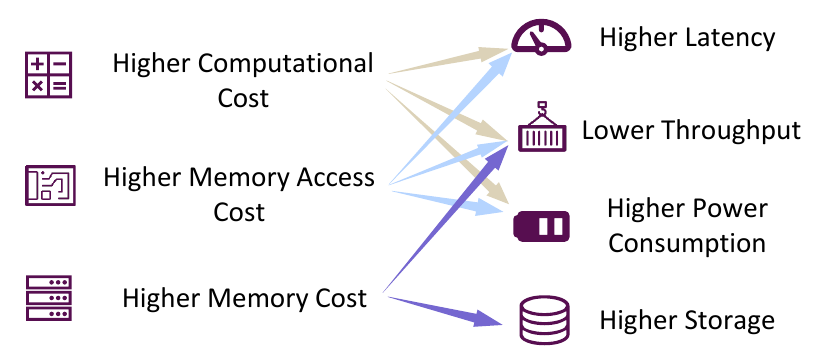}
    \caption{The challenges of LLM deployment.}
    \label{fig:llm_deploy}
\end{figure}

However, the deployment of LLMs is not always going smoothly. 
As shown in Fig.~\ref{fig:llm_deploy}, LLMs typically demand higher computational cost, memory access cost and memory usage in their inference process (we will analyse the root causes in the Sec.~\ref{sec:efficiency}), which deteriorates the efficiency indicators (e.g., latency, throughput, power consumption and storage) in the resource-constrained scenarios. This poses challenges for the application of LLMs in both edge and cloud scenarios. For example, the immense storage requirements render the deployment of a 70-billion-parameter model impractical on personal laptops for tasks such as development assistance. Additionally, the low throughput would result in significant costs if LLMs are used for every search engine request, leading to a considerable reduction in the profits of the search engine.

Fortunately, a substantial array of techniques has been proposed to enable efficient inference for LLMs. To gain a comprehensive understanding of existing studies and inspire further research, this survey employs a hierarchical classification and systematic summarization of the current landscape of efficient LLM inference. Specifically, we categorize relevant studies into three levels: data-level optimization, model-level optimization, and system-level optimization (refer to Sec.~\ref{sec:taxonomy} for elaboration). Moreover, we conduct experimental analyses on representative methods within critical sub-fields to consolidate knowledge, offer practical recommendations, and provide guidance for future research endeavors.

\begin{table}[htb]
\caption{Comparison of existing surveys.}
\label{tab:survey_comparison}
\begin{center}
\resizebox{0.48\textwidth}{!}
{
\begin{tabular}{c|cccc}
\toprule
\multirow{2}{*}[-1ex]{Survey} & \multicolumn{3}{c}{Optimization Levels} & \multirow{2}{*}[-0.5ex]{\makecell[c]{Experimental \\ Analysis}} \\ 
\cmidrule(lr){2-4} 
& Data-level & Model-level & System-level & \\
\midrule
\cite{zhu2023survey,park2024comprehensive,wang2024model} & & \checkmark & &  \\
\cite{tang2024survey} & & \checkmark & & \checkmark \\
\cite{ding2023efficiency} & \checkmark & \checkmark & &  \\
\cite{miao2023towards} & & \checkmark & \checkmark &  \\
\cite{wan2023efficient, xu2024survey} & \checkmark & \checkmark & \checkmark &  \\
Ours & \checkmark & \checkmark & \checkmark & \checkmark \\
\bottomrule
\end{tabular}
}
\end{center}
\end{table}

% Currently, there exist some surveys~\cite{zhu2023survey,park2024comprehensive,wang2024model,ding2023efficiency,miao2023towards,wan2023efficient} that also focus on the field of efficient LLMs, yet they still have room for improvements. Zhu et al.~\cite{zhu2023survey}, Park et al.~\cite{park2024comprehensive} and Wang et al.~\cite{wang2024model} focus on the model compression techniques, a sub-field of model-level optimization, for LLMs. Ding et al.~\cite{ding2023efficiency} review the efficiency research from the perspectives of data and model architecture. While Miao et al.~\cite{miao2023towards} focus on the efficient LLM inference from a machine learning system (MLSys) research perspective. Compared with them, our survey covers a more comprehensive research scope, including three levels of optimization (i.e., data-level, model-level and system-level) with more up-to-date studies. Wan et al.~\cite{wan2023efficient} and Xu et al.~\cite{xu2024survey} also provide a comprehensive review of efficient LLM research. But compared with their work, we further conduct some experiments, and provide practical knowledge and suggestions based on the experimental analysis for some critical sub-fields (e.g., model quantization, serving system). We summarize the comparison in Tab.~\ref{tab:survey_comparison}.

Currently, several surveys~\cite{zhu2023survey,park2024comprehensive,wang2024model,ding2023efficiency,miao2023towards,wan2023efficient,tang2024survey} have been conducted in the field of efficient LLMs. These surveys primarily focus on different aspects of LLM efficiency but offer opportunities for further improvement. Zhu et al.~\cite{zhu2023survey}, Park et al.~\cite{park2024comprehensive}, Wang et al.~\cite{wang2024model} and Tang et al.~\cite{tang2024survey} concentrate on model compression techniques within model-level optimization. Ding et al.~\cite{ding2023efficiency} center on efficiency research considering both data and model architecture perspectives. Miao et al.~\cite{miao2023towards} approach efficient LLM inference from a machine learning system (MLSys) research perspective. 
In contrast, our survey provides a more comprehensive research scope, addressing optimization at three levels: data-level, model-level, and system-level, with the inclusion of recent advancements. While Wan et al.~\cite{wan2023efficient} and Xu et al.~\cite{xu2024survey} also deliver comprehensive review of efficient LLM research, our work extends by incorporating comparative experiments and offering practical insights and recommendations based on experimental analyses in several critical sub-fields like model quantization and serving systems. A comparison of these surveys is summarized in Table~\ref{tab:survey_comparison}.

% \todo{把negtive改成我们positive}

% \todo{give a review of the existing surveys}.

The remainder of this survey is organized as follows: Sec.~\ref{sec:preliminary} introduces the basic concept and knowledge about LLMs and presents a detailed analysis of the efficiency bottlenecks during the inference process of LLMs. Sec.~\ref{sec:taxonomy} demonstrates our taxonomy. Sec.~\ref{sec:data-level-opt} to Sec.~\ref{sec:system-level-opt} respectively present and discuss studies on efficiency optimization at three distinct levels. Sec.~\ref{sec:discussion} offers broader discussions for several key application scenarios. Sec.~\ref{sec:conclusion} concludes the key contributions provided by this survey.

%\subsection{Background and Efficiency Analysis}
\section{Preliminaries}
\label{sec:preliminary}

\subsection{Transformer-Style LLMs}
\label{sec:transformer}

Language modeling, as the fundamental function of language models (LMs), involves modeling the likelihood of the word sequence and predicting the distribution of subsequent words. Over recent years, researchers have discovered that scaling up language models not only enhances their language modeling ability but also engenders emergent capabilities for tackling more intricate tasks beyond conventional NLP tasks~\cite{zhao2023survey}. These scaled-up language models are referred to as large language models (LLMs). 

The mainstream LLMs are designed based on the Transformer architecture~\cite{vaswani2017attention}. Specifically, a typical Transformer architecture is composed of several stacked Transformer blocks. Typically, a Transformer block consists of a Multi-Head Self-Attention (MHSA) block, a Feed Forward Network (FFN), and a LayerNorm (LN) operation. For each block, it receives the output features of the previous one as the input, and passes the features through each sub-module to obtain the output. Specially, before the first block, a tokenizer is used to convert the original input sentence into a sequence of tokens, and a following embedding layer serves to convert the tokens into the input features. Then, the additional position embeddings are added into the input features to encode the sequential order of each input token. 

The core concept of the Transformer architecture is the self-attention mechanism, which is adopted in the MHSA block. Specifically, denoted the input features as $X=[x_1, x_2, ..., x_n]$, the MHSA block applies linear projection to them and obtains a set of queries Q, keys K and values V as Eq.~\ref{eq:mhsa_linear}: 
\begin{equation}
    Q_i=XW^{Q_i}, K_i=XW^{K_i}, V_i=XW^{V_i},
    \label{eq:mhsa_linear}
\end{equation}
where $W^{Q_i}$, $W^{K_i}$ and $W^{V_i}$ are the projection matrices corresponding to the $i$-th attention head. Then the self-attention operation is applied to each tuple of $(Q_i, K_i, V_i)$ and get the feature of the $i$-th attention head $Z_i$ as Eq.~\ref{eq:mhsa_attention}: 
\begin{equation}
    Z_i={\rm Attention}(Q_i, K_i, V_i)={\rm Softmax}(\frac{Q_iK_i^T}{\sqrt{d_k}})V_i, 
    \label{eq:mhsa_attention}
\end{equation}
where $d_k$ is the dimension of the queries (keys). Note that the self-attention operation contains the matrix multiplication operation, its computation complexity is quadratic in the input length. Finally, the MHSA block concatenates the features of all the attention heads and applies a linear projection to them to form its output $Z$ as Eq.~\ref{eq:mhsa_concat}: 
\begin{equation}
    Z={\rm Concat}(Z_1, Z_2, ..., Z_h)W^O, 
    \label{eq:mhsa_concat}
\end{equation}
where $W_O$ is the projection matrix. As can be seen, the self-attention mechanism allows the model to identify the importance of different input parts regardless of the distance, and thus can capture the long-range dependencies and complex relationships in the input sentence.

Another important module in the Transformer block is the FFN. Typically, FFN is placed after the MHSA block and consists of two linear transformation layers with a non-linear activation function. It receives the output features $X$ from the MHSA block and processes them as Eq~\ref{eq:ffn}: 
\begin{equation}
    {\rm FFN}(X) = W_2\sigma(W_1X), 
    \label{eq:ffn}
\end{equation}
where $W_1$ and $W_2$ denote the weight matrices of the two linear layers, and $\sigma(\cdot)$ denotes the activation function. 
% In the previous study, it has been proved that FFN plays an important role in slowing the expressive power degradation of the Transformer model w.r.t. the model depth. 

\begin{figure}[h]
    \centering
    \includegraphics[width=1.0\linewidth]{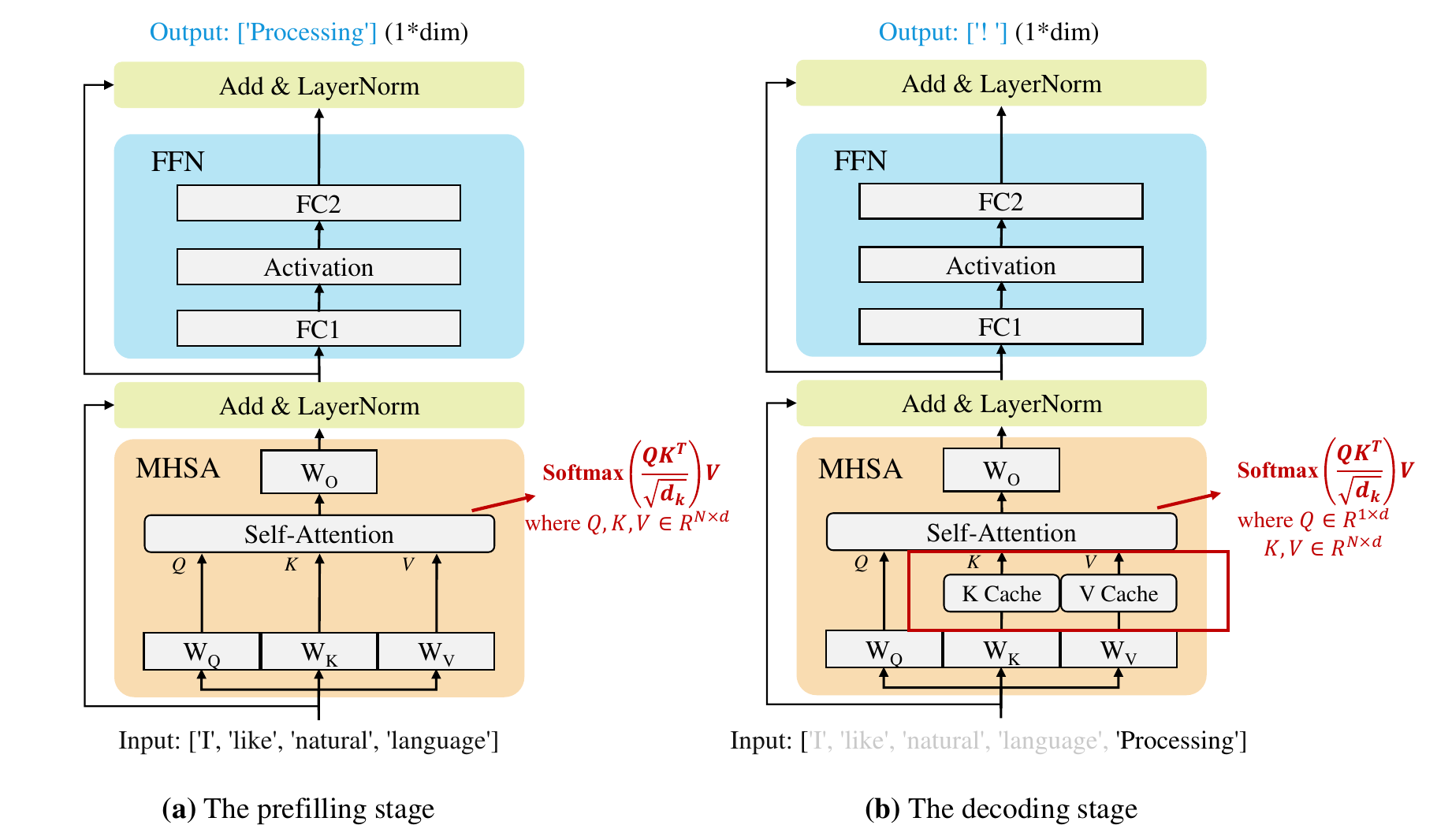}
    \caption{Demonstration of the prefilling stage (a) and decoding stage (b).}
    \label{fig:transformer_block}
\end{figure}

\subsection{Inference Process of LLMs}
\label{sec:inference}

The most popular LLMs, i.e., decoder-only LLMs, often adopt the auto-regressive method to generate the output sentence. Specifically, the auto-regressive method generates the tokens one by one. In each generation step, the LLM takes as input the whole token sequences, including the input tokens and previously generated tokens, and generates the next token. With the increase in sequence length, the time cost of the generation process grows rapidly. To address this challenge, a crucial technique, namely key-value (KV) cache, has been introduced to expedite the generation process. The KV cache technique, as its name suggests, involves storing and reusing previous key (K) and value (V) pairs within the Multi-Head Self-Attention (MHSA) block. This technique has been widely adopted in LLM inference engines and systems due to its substantial optimization of generation latency. 
Based on the above methods and techniques, the inference process of LLMs can be divided into two stages:
\begin{itemize}
    \item \textbf{Prefilling Stage}: The LLM calculates and stores the KV cache of the initial input tokens, and generates the first output token, as shown in Fig.~\ref{fig:transformer_block}(a).
    \item \textbf{Decoding Stage}: The LLM generates the output tokens one by one with the KV cache, and then updates it with the key (K) and value (V) pairs of the newly generated token, as shown in Fig.~\ref{fig:transformer_block}(b).
\end{itemize}

\begin{figure}
    \centering
    \includegraphics[width=0.98\linewidth]{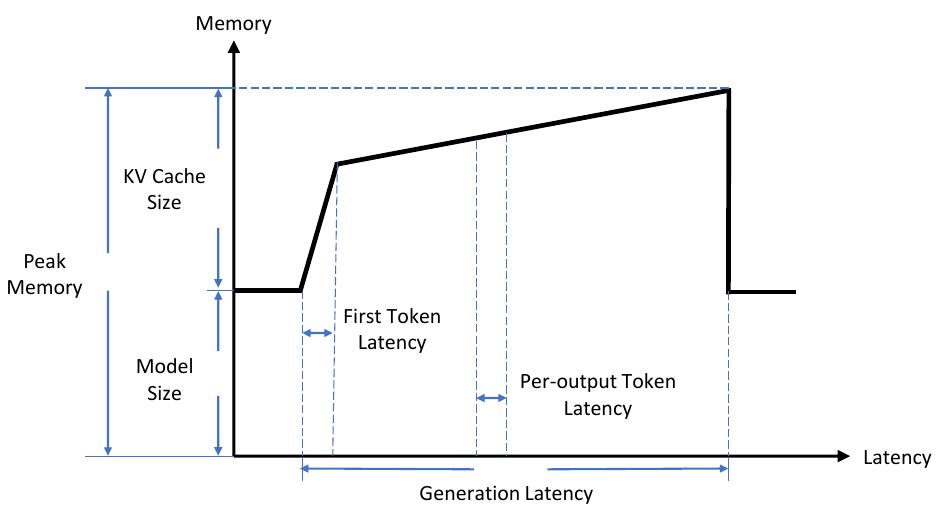}
    \caption{Illustration of the memory variation through time (latency) during one generation process. Note that we ignore the activation size in this figure for a simplification.}
    \label{fig:indicators}
\end{figure}

\noindent As shown in Fig.~\ref{fig:indicators}, we illustrate some critical efficiency indicators. As for the latency, we denote \textbf{first token latency} as the latency to generate the first output token in the prefilling stage, while we denote \textbf{per-output token latency} as the average latency to generate one output token in the decoding stage. Besides, we use \textbf{generation latency} to denote the latency to generate the whole output token sequences. As for the memory, we use \textbf{model size} to denote the memory to store the model weights, and use \textbf{KV cache size} to denote the memory to store the KV cache. Additionally, \textbf{peak memory} denotes the maximum memory usage during the generation process, which is approximately equal to the memory sum of model weights and KV cache. Apart from the latency and memory, throughput is also a widely-used indicator in the LLM serving system. We use \textbf{token throughput} to denote the number of generated tokens per second, and use \textbf{request throughput} to denote the number of completed requests per second.

% Fig.~\ref{fig:transformer_block} demonstrates 

\subsection{Efficiency Analysis}
\label{sec:efficiency}

%\todo{画一个大的阶段图，标出first token latency、per-output token latency、generation latency，throughput分output token/s、request/s，storage分model size、peak memory，缩写、全称、示意}

%[done] \todo{@zzx 忽略activation}

Deploying LLMs on resource-constrained scenarios while preserving their powerful capabilities poses a significant challenge for both practitioners and researchers. For instance, let's consider to deploy a LLaMA-2-70B model, which contains 70 billion parameters. Storing its weights in FP16 format necessitates 140 GB of VRAM, requiring at least 6 RTX 3090Ti GPUs (each with 24 GB VRAM) or 2 NVIDIA A100 GPUs (each with 80 GB VRAM) for inference. As for latency, generating one token on 2 NVIDIA A100 GPUs requires approximately 100 milliseconds. Consequently, generating a sequence with hundreds of tokens requires more than 10 seconds. In addition to storage and latency, the efficiency indicators, such as throughput, energy and power consumption, also need to be considered. During the LLM inference process, three important factors would largely affect these indicators, i.e., the \textbf{computational cost}, the \textbf{memory access cost} and the \textbf{memory usage}. Yuan et al.~\cite{yuan2024llm} provide a more systematic analysis to demonstrate how these factors affect the inference inefficiency with a roofline model. 
In the following, we further analyze three root causes of inefficiency in the LLM inference process, focusing on the above three key factors: 
\begin{itemize}
    \item \textbf{Model Size}: Mainstream LLMs typically incorporate billions or even trillions of parameters. For instance, the LLaMA-70B model comprises 70 billion parameters, while the GPT-3 model scales up to 175 billion parameters. This considerable model size contributes significantly to the elevated computational cost, memory access cost, and memory usage during the LLM inference process.
    
    \item \textbf{Attention Operation}: As illustrated in Sec.~\ref{sec:transformer} and Sec.~\ref{sec:inference}, in the prefilling stage, the self-attention operation exhibits quadratic computational complexity in the input length. Consequently, as the input length increases, the computational cost, memory access cost, and memory usage of the attention operation escalate rapidly.
    
    \item \textbf{Decoding Approach}: The auto-regressive decoding approach generates the tokens one by one. In each decoding step, all the model weights are loaded from the off-chip HBM to the GPU chip, leading to a large memory access cost. In addition, the size of KV cache increases with the growth in the input length, potentially leading to fragmented memory and irregular memory access patterns.
\end{itemize}

\begin{figure*}[h]
\centering
\tikzset{
    basic/.style  = {draw, text width=2cm, align=center, font=\sffamily, rectangle},
    root/.style   = {basic, rounded corners=2pt, thin, align=center, fill=white,text width=8cm, rotate=90, font=\footnotesize},
    dnode/.style = {basic, thin, rounded corners=2pt, align=center, fill=ngreen,text width=3.5cm, font=\footnotesize},
    dnode_1/.style = {basic, thin, rounded corners=2pt, align=center, fill=ngreen,text width=2.5cm, font=\footnotesize},
    mnode/.style = {basic, thin, rounded corners=2pt, align=center, fill=nblue, text width=3.5cm, font=\footnotesize},
    mnode_1/.style = {basic, thin, rounded corners=2pt, align=center, fill=nblue, text width=2.5cm, font=\footnotesize}, 
    snode/.style = {basic, thin, rounded corners=2pt, align=center, fill=npurple,text width=3.5cm, font=\footnotesize},
    snode_1/.style = {basic, thin, rounded corners=2pt, align=center, fill=npurple,text width=2.5cm, font=\footnotesize},
    tnode/.style = {basic, thin, align=left, fill=pink!60, text width=15em, align=center},
    xnode/.style = {basic, thin, rounded corners=2pt, align=center, fill=blue!20,text width=5cm,},
    wnode/.style = {basic, thin, align=left, fill=pink!10!blue!80!red!10, text width=6.5em},
    %edge from parent/.style = {draw=black, edge from parent fork right}
    %edge from parent/.style = {draw=black, edge from parent fork down}
}
\begin{forest} 
for tree={
    if level=0{
        grow=east,
        growth parent anchor=east,
        parent anchor=south,
        child anchor=west,
        edge path={\noexpand\path[\forestoption{edge},->, >={latex}] 
             (!u.parent anchor) -- +(5pt,0pt) |- (.child anchor)
             \forestoption{edge label};},
    }
    {
        grow=east,
        growth parent anchor=east,
        parent anchor=east,
        child anchor=west,
        edge path={\noexpand\path[\forestoption{edge},->, >={latex}] 
             (!u.parent anchor) -- +(5pt,0pt) |- (.child anchor)
             \forestoption{edge label};},
    }
}
% l sep is used for arrow distance
[Efficient Inference for Large Language Models, root
    [System-level Optimization (Sec.~\ref{sec:system-level-opt}), snode_1
        [Serving System \\ (Sec.~\ref{sec:serving_system}), snode
            [Distributed Systems, snode]
            [Scheduling, snode]
            [Batching, snode]
            [Memory Management, snode]
        ]
        [Inference Engine \\ (Sec.~\ref{sec:inference_engine}), snode
            %[Model Parallelism, snode]
            [Speculative Decoding, snode]
            [Offloading, snode]
            [Graph and Operator Optimization, snode]
        ]
    ]
    [Model-level Optimization (Sec.~\ref{sec:model-level-opt}), mnode_1
        [Model Compression \\ (Sec.~\ref{sec:model_compression}), mnode
            [Dynamic Inference, mnode]
            [Knowledge Distillation, mnode
                [Black-box KD, mnode]
                [White-box KD, mnode]
            ]
            [Structure Optimization, mnode
                [Neural Architecture Search, mnode]
                [Structure Factorization, mnode]
            ]
            [Sparsification, mnode
                [Sparse Attention, mnode]
                [Weight Pruning, mnode]
            ]
            [Quantization, mnode
                [Quantization-aware Training, mnode]
                [Post-Training Quantization, mnode]
            ]
        ]
        [Efficient Structure Design \\ (Sec.~\ref{sec:efficient_structure}), mnode
            [Transformer Alternate, mnode]
            [Efficient Attention Design, mnode
                [Multi/Group-Query Attention, mnode]
                [Low-Complexity Attention, mnode]
            ]
            [Efficient FFN Design, mnode]
        ]
    ]
    [Data-level Optimization (Sec.~\ref{sec:data-level-opt}), dnode_1
        [Output Organization \\ (Sec.~\ref{sec:output_compress}), dnode]
        [Input Compression \\ (Sec.~\ref{sec:input_compress}), dnode
            [Retrieval-Augmented Generation, dnode]
            [Soft Prompt-based Compression, dnode]
            [Prompt Summary, dnode]
            [Prompt Pruning, dnode]
        ]
    ]
]
\end{forest}

\caption{Taxonomy of efficient inference methods for Large Language Models.}
\label{fig:framework}
\end{figure*}
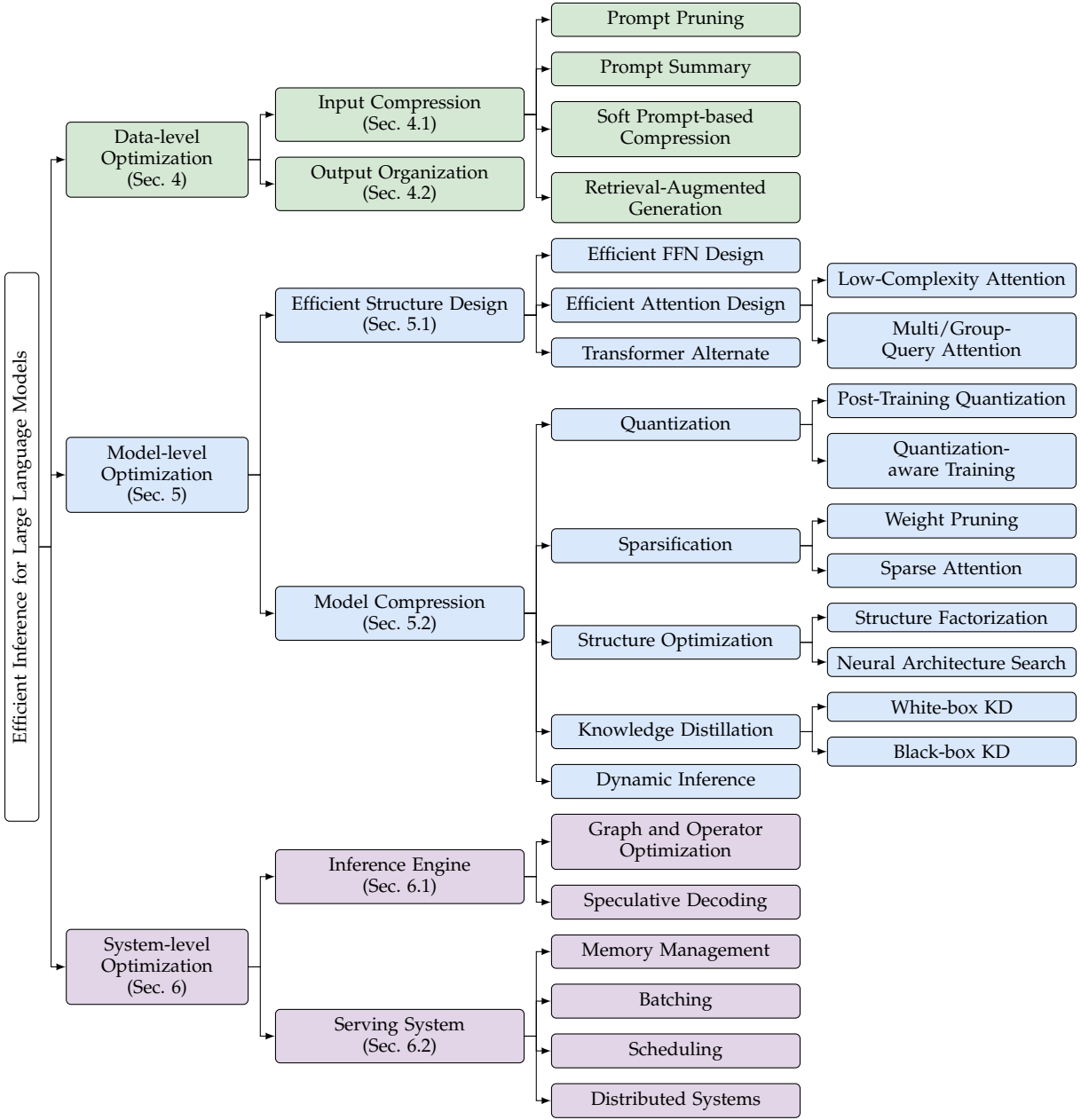

% \begin{figure*}[h]
%     \centering
%     \includegraphics[width=0.90\linewidth]{figures/framework.pdf}
%     \caption{Taxonomy of efficient inference methods for Large Language Models.}
%     \label{fig:framework}
% \end{figure*}

% \todo{combination of different level; overview analysis; high-level and practical guidance; MoE; data-level with application, system-level with hardware}

\section{Taxonomy}
\label{sec:taxonomy}

In the aforementioned discussion, we identify key factors (i.e., computational cost, memory access cost and memory usage) that significantly impact the efficiency during the LLM inference process, and further analyze three root causes (i.e., model size, attention operation and decoding approach). Many efforts have been made to optimize the inference efficiency from different perspectives. By carefully reviewing and summarizing these studies, we classify them into three levels, i.e., data-level optimization, model-level optimization and system-level optimization (as shown in Fig.~\ref{fig:framework}):  
\begin{itemize}
    \item \textbf{Data-level Optimization} refers to improving the efficiency via optimizing the input prompts (i.e., input compression) or better organizing the output content (i.e., output organization). This line of optimization typically does not change the original model, thus is free of costly model training cost (note that a small amount of training for auxiliary models might be required, but this cost can be ignored compared with the training cost for original LLMs). %, and (2) cannot guarantee the equivalence of the model outputs unlike system-level optimization. %\todo{计算结果不等价} is not equvi... unlike system-level optimization. 
    
    \item \textbf{Model-level Optimization} refers to designing an efficient model structure (i.e., efficient structure design) or compressing the pre-trained models (i.e., model compression) in the inference process to improve its efficiency. This line of optimization (1) often requires costly pre-training or a smaller amount of fine-tuning cost to retain or recover the model ability, and (2) is typically lossy in the model performance. 
    
    \item \textbf{System-level Optimization} refers to optimizing the inference engine or the serving system. This line of optimization (1) does not involve the costly model training, and (2) is typically lossless in model performance\footnote{A recent study~\cite{golden2024flash} shows that FlashAttention, a common-used system-level optimization technique, might cause the numeric deviation.}. In addition, we provide a brief introduction for hardware accelerator design in Sec.~\ref{sec:hardware}. 
    %For hardware accelerator design, we provide a brief introduction in Sec.~\ref{sec:discussion}. 
\end{itemize}

\section{Data-level Optimization}
\label{sec:data-level-opt}

In the data level, prior studies can be divided into two categories, i.e., input compression and output organization. Input compression techniques directly shorten the model input to reduce the inference cost. While output organization techniques enable batch (parallel) inference via organizing the structure of output content, which can improve the hardware utilization and reduce the generation latency.

\begin{figure*}[h]
\centering
\tikzset{
    basic/.style  = {draw, text width=2cm, align=center, font=\sffamily, rectangle},
    root/.style   = {basic, rounded corners=2pt, thin, align=center, fill=white,text width=8cm, rotate=90, font=\footnotesize},
    dnode/.style = {basic, thin, rounded corners=2pt, align=center, fill=ngreen, text width=3.5cm, font=\footnotesize},
    dnode_1/.style = {basic, thin, rounded corners=2pt, align=center, fill=ngreen,text width=2cm, font=\footnotesize},
    mnode/.style = {basic, thin, rounded corners=2pt, align=center, fill=blue!10,text width=3.5cm, font=\footnotesize},
    mnode_1/.style = {basic, thin, rounded corners=2pt, align=center, fill=blue!10,text width=2.5cm, font=\footnotesize}, 
    snode/.style = {basic, thin, rounded corners=2pt, align=center, fill=green!30,text width=3.5cm, font=\footnotesize},
    snode_1/.style = {basic, thin, rounded corners=2pt, align=center, fill=green!30,text width=2.5cm, font=\footnotesize},
    tnode/.style = {basic, thin, align=left, fill=pink!60, text width=15em, align=center},
    xnode/.style = {basic, thin, rounded corners=2pt, align=center, fill=blue!20,text width=5cm,},
    wnode/.style = {basic, thin, rounded corners=2pt, align=left, fill=white,text width=6cm, font=\footnotesize},
    %edge from parent/.style = {draw=black, edge from parent fork right}
    %edge from parent/.style = {draw=black, edge from parent fork down}
}
\begin{forest} 
for tree={
    grow=east,
    growth parent anchor=east,
    parent anchor=east,
    child anchor=west,
    edge path={\noexpand\path[\forestoption{edge},->, >={latex}] 
         (!u.parent anchor) -- +(5pt,0pt) |- (.child anchor)
         \forestoption{edge label};}
}
% l sep is used for arrow distance
[Input Compression, dnode_1
    [Retrieval-Augmented Generation, dnode
        [{RAG~\cite{lewis2020retrieval}, FLARE~\cite{chevalier2023adapting}, REPLUG~\cite{shi2023replug}, Self-RAG~\cite{asai2023selfrag}}, wnode]
    ]
    [Soft Prompt-based Compression, dnode
        [{PromptCompression~\cite{wingate2022prompt}, Gisting~\cite{mu2023learning}, AutoCompressors~\cite{chevalier2023adapting}, ICAE~\cite{ge2023context}}, wnode]
    ]
    [Prompt Summary, dnode
        [{RECOMP~\cite{xu2023recomp}, SemanticCompression~\cite{fei2023extending}}, wnode]
    ]
    [Prompt Pruning, dnode
        [{DYNAICL~\cite{zhou2023efficient}, Selective Context~\cite{li2023compressing}, STDC~\cite{yin2023did}, PCRL~\cite{jung2023discrete}, RECOMP~\cite{xu2023recomp}, LLMLingua~\cite{jiang2023llmlingua}, LongLLMLingua~\cite{jiang2023longllmlingua}, CoT-Influx~\cite{huang2023boosting}}, wnode]
    ]
]
\end{forest}

\caption{Taxonomy of the input compression methods for Large Language Models.}
\label{fig:taxonomy_input_compression}
\end{figure*}
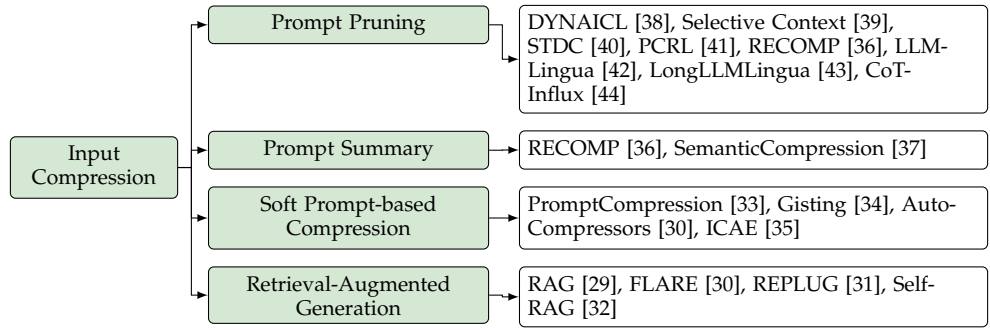

\subsection{Input Compression}
\label{sec:input_compress}

In the practical application of LLMs, prompts are crucial. Numerous studies suggest new ways to design prompts effectively and show in practice that well-designed prompts can unleash the capabilities of LLMs.
For instance, In-Context Learning (ICL)~\cite{dong2022survey} suggests to include multiple relevant examples within the prompt. This approach encourages LLMs to learn through analogy.
%Retrieval-Augmented Generation (RAG)~\cite{lewis2020retrieval} proposes to simply prepend retrieved documents to the prompt, which enables LLMs to retrieve answers from the documents. 
Chain-of-Thought (CoT)~\cite{wei2022chain} proposes to incorporate a sequence of intermediate reasoning steps within the in-context examples, which help LLMs to conduct complex reasoning. However, these prompting techniques inevitably lead to longer prompts, which poses a challenge because the computational cost and memory usage increase quadratically during the prefilling stage (as illustrated in Sec.~\ref{sec:efficiency}).

To address this challenge, input prompt compression~\cite{wingate2022prompt} has been proposed to shorten prompts without significantly impacting the quality of answers from LLMs. Within this field, relevant studies are categorized into four groups, as depicted in Figure \ref{fig:taxonomy_input_compression}: prompt pruning, prompt summary, soft prompt-based compression, and retrieval-augmented generation.

% \todo{每个subsection第一句写online还是offline，然后把对应的工作全部列在这}

\subsubsection{Prompt Pruning}

The core idea behind the prompt pruning is to remove unimportant tokens, sentences, or documents online from each input prompt based on predefined or learnable importance indicators. 
%This line of techniques prunes the input prompts online, which means that the pruning algorithm would be executed for each new coming prompt. 
DYNAICL~\cite{zhou2023efficient} proposes to dynamically decide the optimal number of in-context examples for a given input based on the computational budget via a well-trained LLM-based meta controller. %It trains and adopts a LLM-based meta controller to output the suitable number for the input to achieve the best performance-efficiency trade-off. Finally, DYNAICL saves up to 46\% token budget across 8 NLP tasks with a acceptable performance loss. 
Selective Context~\cite{li2023compressing} proposes to merge tokens into units, and then applies a unit-level prompt pruning based on the self-information indicator (i.e., negative log likelihood). %adopts the self-information value (i.e., negative log likelihood) of a token as its importance indicator, and prunes the tokens that are less important based on this indicator. To avoid the disjoint context after token-level pruning, it further proposes to first merge the tokens into the so-called lexical units, and then applies the unit-level prompt pruning. As a result, Selective Context achieves a 50\% context cost on four downstream tasks with a small performance drop. 
STDC~\cite{yin2023did} prunes the prompts based on the parse tree, which iteratively removes phrase nodes that cause the smallest performance drop after pruning it. %STDC prunes 60\% tokens with small performance loss on NIV2 benchmark~\cite{wang2022super}. 
PCRL~\cite{jung2023discrete} introduces a token-level pruning scheme based on reinforcement learning. The main idea behind PCRL is to train a policy LLM by combining faithfulness and compression ratio into the reward function. Faithfulness is measured as the output similarity between the compressed prompt and the original prompt. %, which employs a frozen LLM with several learnable MLP layers as the policy network, and uses the compression ratio as the reward function. %PCRL prunes 24.6\% tokens on average on the Alpaca++ dataset~\cite{mu2023learning} while preserving performance. 
RECOMP~\cite{xu2023recomp} implements a sentence-level pruning strategy to compress prompts for Retrieval-Augmented Language Models (RALMs). The approach involves encoding the input question and documents into latent embeddings using a pre-trained encoder. Then, it decides which documents to remove based on the similarity of their embeddings with the question's embedding. %designs two compressors to compress the prompts for the Retrieval-Augmented Language Model (RALM). One of them, named Extractive Compressor, adopts a sentence-level pruning strategy. It trains a encoder to embed the sentences and input into embeddings, and uses the similarity of the embeddings to evalate the helpfulness of the sentences. Sentences with lower similarity are pruned. Extractive Compressor achieves 25\% compression ratio at minimum performance drop on language modeling task. 
LLMLingua~\cite{jiang2023llmlingua} introduces a coarse-to-fine pruning scheme for prompt compression. Initially, it performs a demonstration-level pruning followed by token-level pruning based on perplexity. To enhance performance, LLMLingua proposes a budget controller that dynamically allocates the pruning budget across different parts of prompts. Additionally, it utilizes an iterative token-level compression algorithm to address inaccuracies introduced by conditional independence assumptions. Furthermore, LLMLingua incorporates a distribution alignment strategy to align the output distribution of the target LLM with a smaller LLM used for perplexity calculation. %LLMLingua achieves up to 20×\times compression with little performance loss across four datasets. 
LongLLMLingua~\cite{jiang2023longllmlingua} builds upon LLMLingua with several enhancements: (1) It utilizes perplexity conditioned on the input question as the indicator for prompt pruning. (2) It allocates varying pruning ratios to different demonstrations and reorders the demonstrations within the final prompt based on their indicator values. (3) It restores the original content based on the response. %LongLLMLingua achieves higher task performance with much shorter prompts across three NLP benchmarks. 
CoT-Influx~\cite{huang2023boosting} introduces a coarse-to-grained pruning method for Chain-of-Thought (CoT) prompts using reinforcement learning. Specifically, it prunes unimportant examples, followed by pruning unimportant tokens within the remaining examples. %CoT-Influx employs a small BERT model as the policy network, and designs the reward function based on the prompt length and the model performance after pruning.

\subsubsection{Prompt Summary}

The core idea of prompt summary is to condense the original prompt into a shorter summary while preserving similar semantic information. These techniques also serve as online compression methods for prompts. 
In contrast to the aforementioned prompt pruning techniques that preserve the unpruned tokens unchanged, this line of methods converts the entire prompt into its summation. 
RECOMP~\cite{xu2023recomp} introduces an Abstractive Compressor that takes an input question and retrieved documents as input, and produces a concise summary. Specifically, it distills a lightweight compressor from the extreme-scale LLMs to perform the summary. %Abstractive Compressor achieves high compression rate with small performance drop on language modeling task. 
SemanticCompression~\cite{fei2023extending} proposes a semantic compression method. It starts by breaking down the text into sentences. Next, it groups sentences together by topic and then summarizes the sentences within each group. %SemanticCompression achieves 6∼\sim8×\times compression ratio.

\subsubsection{Soft Prompt-based Compression}

The core idea of this kind of compression techniques is to design a soft prompt, significantly shorter than the original prompt, for use as input to LLMs. The soft prompt is defined as a sequence of learnable continuous tokens. 
Some techniques adopt offline compression for the fixed prefix prompt (e.g., system prompt, task-specific prompt). 
For example, PromptCompression~\cite{wingate2022prompt} trains a soft prompt to emulate a predetermined system prompt. The approach involves adding several soft tokens before the input tokens and enabling these soft tokens to be adjusted during back-propagation. Following fine-tuning on the prompt dataset, the sequence of soft tokens serves as the soft prompt. 
Gisting~\cite{mu2023learning} introduces a method to condense task-specific prompts into a concise set of gist tokens using prefix-tuning~\cite{li2021prefix}. Given that task-specific prompts differ across tasks, prefix-tuning is applied individually for each task. To enhance efficiency, Gisting further introduces a meta-learning approach that predicts gist tokens for new unseen tasks based on the gist tokens of previous tasks. %Gisting achieves up to 26×\times prompt compression on a proposed dataset Alpaca++. 

Other techniques adopt online compression for every new input prompts. 
For instance, AutoCompressors~\cite{chevalier2023adapting} train a pre-trained LM to compress the prompts into summary vectors via unsupervised learning. %AutoCompressors extends the LLM's context window size, enabling it to handle longer prompts. 
ICAE~\cite{ge2023context} trains an autoencoder to compress the original context into short memory slots. Specifically, ICAE employs a LoRA-adapted LLM as the encoder, and uses the target LLM as the decoder. A set of memory tokens is added before the input tokens and encoded into memory slots. %ICAE achieves 4×\times prompt compression with acceptable performance loss. 

\subsubsection{Retrieval-Augmented Generation}
Retrieval-Augmented Generation (RAG)~\cite{lewis2020retrieval} aims to improve the quality of LLMs' responses by incorporating external knowledge sources. RAG can be also viewed as a technique to improve the inference efficiency when handling a large amount of data. Instead of merging all information into an excessively long prompt, RAG only adds relevant retrieved information to the original prompt, ensuring that the model receives necessary information while reducing prompt length significantly. 
FLARE~\cite{chevalier2023adapting} uses predictions of upcoming sentences to proactively decide when and what information to retrieve. 
REPLUG~\cite{shi2023replug} treats the LLM as a black box and augments it with a tuneable retrieval model. It prepends retrieved documents to the input for the frozen black-box LLM, and further utilizes the LLM to supervise the retrieval model. 
Self-RAG~\cite{asai2023selfrag} enhances LLM's quality and factuality through retrieval and self-reflection. It introduces reflection tokens to make the LLM controllable during the inference phase.

\subsection{Output Organization}
\label{sec:output_compress}
%The core idea of ... is that \todo{generation过程不一定是完全串行的，可以通过输出结构来并行}
The traditional generation process of LLMs is entirely sequential, leading to significant time consumption. Output organization techniques aim to (partially) parallelize generation via organizing the structure of output content. 

Skeleton-of-Thought (SoT)~\cite{ning2023skeleton} is pioneering in this direction. The core idea behind SoT is to leverage the emerging ability of LLMs to plan the output content's structure. 
%\todo{Core idea: SoT探究利用大语言模型本身的能力做输出的规划}
% Skeleton-of-Thought (SoT)~\cite{ning2023skeleton} makes the first attempt to accelerate the LLM inference in the data-centric manner. 
%SoT is to first guide the LLM to plan its response's structure, and then apply batch decoding process to reduce the overall generation latency. 
%\todo{stage->phase @zzx}
Specifically, SoT consists of two main phases. In the first phase (i.e., skeleton phase), SoT instructs the LLM to generate a concise skeleton of the answer using a predefined "skeleton prompt." For instance, given a question like "What are the typical types of Chinese dishes?", the output at this stage would be a list of dishes (e.g., noodles, hot pot, rice) without elaborate descriptions. Then, in the second phase (i.e., point-expanding phase), SoT instructs the LLM to expand each point in the skeleton simultaneously using a "point-expanding prompt," and then concatenates these expansions to form the final answer. When applied to open-source models, point-expanding can be performed through batch inference, which optimizes hardware utilization and reduces overall generation latency using the same computational resources. 
%\todo{参考下revision版本文章rephrase下 @zzx}
%For example, as shown in Fig.~\ref{fig:sot_demo}(b), LLM writes more detailed descriptions of the dishes listed in the first stage. 
\begin{figure}[h]
    \centering
    \includegraphics[width=0.98\linewidth]{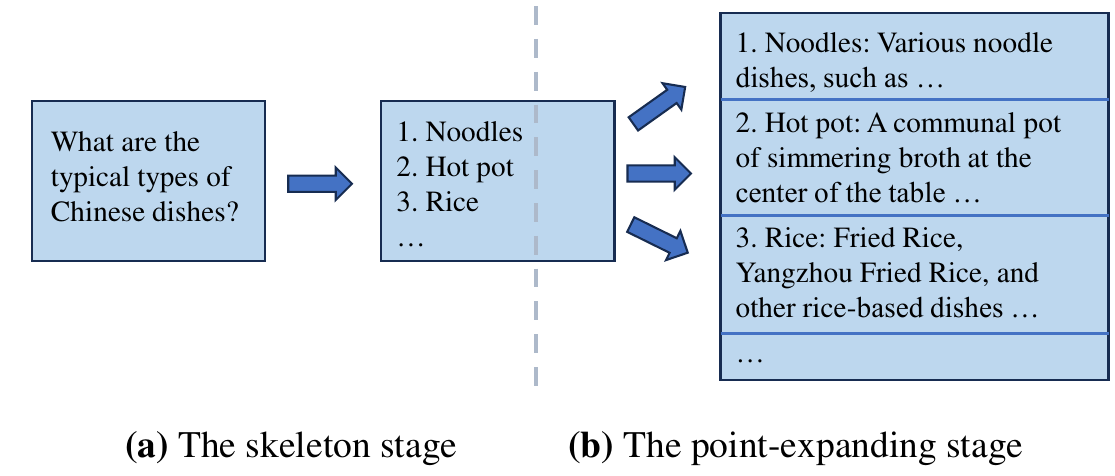}
    \caption{Demonstration of the inference process of SoT.}
    \label{fig:sot_demo}
\end{figure}
To mitigate the additional computation overhead brought by the extra prompt (i.e., \textit{skeleton prompt} and \textit{point-expanding prompt}), SoT discusses the possibility of sharing the KV cache of the common prompt prefix across multiple points in the point expansion phase. Additionally, SoT uses a router model to decide whether applying SoT is appropriate for specific questions, aiming to limit its use to suitable cases. 
As a result, SoT achieves up to a 2.39$\times$ speed-up on 12 recently released LLMs, and improves the answer quality for many questions by improving the diversity and relevance of their answer. 
%Following the above paradigm, SoT achieves 1.13$\sim$2.39$\times$ speed-ups on 9 open-source LLMs and 3 API-based LLMs. In addition, SoT also improves the answer quality in some cases, since it can improve diversity and relevance of the answer. Note that SoT does not require any change to the model, system and hardware, and achieves such the outstanding efficiency improvement via prompt engineering. In summary, SoT demonstrates the potential of optimizing the inference efficiency of LLMs in the data-level. Nevertheless, SoT still lacks of generalization ability in practical scenarios, and causes unignorable token overheads, which need to be improved in the future work. 
% \todo{除了batch inference，SoT提出共享kvcache来降低额外的prompt导致的computation overhead}
% \todo{To only trigger SoT
%  for suitable questions, SoT proposes ... router ...}

SGD~\cite{jin2024adaptive} further extends the idea of SoT by organizing sub-problem points into a Directed Acyclic Graph (DAG) and answering the logic-independent sub-problems in parallel in one turn. Similar to SoT, SGD also leverages the emerging ability of LLMs to generate the output structure by providing manually-crafted prompts along with several examples. 
%\todo{通过人工设计的examples guide大语言模型...}
SGD relaxes the strict independence assumption among different points to enhance the quality of answers, especially for math and coding problems. Compared with SoT, SGD prioritizes answer quality over speed. Additionally, SGD introduces an adaptive model selection approach, assigning an optimal model size to handle each sub-problem based on its estimated complexity, thus further improving efficiency. 

% APAR~\cite{liu2024apar} adopts a similar idea with SoT, leveraging the parallelism in the output content to enable parallel decoding. The core design of APAR leveraging the enhanced LLMs, which are instruct-tuned on carefully-designed data that formed in specific tree structure, to automatically organizing the parallelizable structure in the output content. APAR significantly benefits from LLM fine-tuning to not only reduce the end-to-end generation latency but also retain the answer quality. Furthermore, APAR combines their decoding approach with the speculative decoding technique (i.e., Medusa~\cite{cai2024medusa}) and serving system (i.e. vLLM~\cite{kwon2023vllm}) to further improve the inference latency and system throughput, respectively. 

APAR~\cite{liu2024apar} adopts a similar idea with SoT, leveraging LLMs to output special control tokens (i.e., ${\rm [fork]}$) for automatically and dynamically triggering the parallel decoding. To effectively exploit the inherent parallelizable structure within the output content and accurately generate control tokens, APAR fine-tunes the LLMs on carefully-designed data that formed in specific tree structure. As a result, APAR achieves an average 1.4$\sim$2.0$\times$ speed-up on benchmarks and cases a negligible impact on the answer quality. Furthermore, APAR combines their decoding approach with the speculative decoding technique (i.e., Medusa~\cite{cai2024medusa}) and serving system (i.e. vLLM~\cite{kwon2023vllm}) to further improve the inference latency and system throughput, respectively. 

% SGLang~\cite{zheng2023efficiently} proposes an efficient LLM programming system by co-organizing the front-end language (data) and the back-end runtime. 
SGLang~\cite{zheng2023efficiently} introduces a domain-specific language (DSL) in Python featuring primitives that flexibly facilitate LLM programming. The core idea behind SGLang is to analyze dependencies among various generation calls automatically, and perform batch inference and KV cache sharing based on this analysis.
%\todo{自动分析依赖关系，根据依赖关系做batch inference和kv cache的共享}
%Then, SGLang develops an interpreter to execute SGLang programs automatic parallelism, batching, caching, and sharing across multiple calls and multiple programs.
With this language, users can implement various prompting strategies easily and benefit from the automatic efficiency optimization of SGLang (e.g., SoT~\cite{ning2023skeleton}, ToT~\cite{yao2024tree}). 
Furthermore, SGLang introduces and combines several system-level compilation techniques, such as code movement and prefetching annotations. 
%\todo{SGLang这样一个interface，open up more possibility for furture 后端优化，比如code movement...}
% Furthermore, SGLang also develops an effective compiler to optimize the program execution via several enhanced strategies, such as code movement and prefetching annotations. 

%Note that SGD obtains lower speed-ups than SoT since it yields a lower degree of parallelism. Thus, SGD actually explores a better trade-off between generation efficiency and quality. 
%\todo{SGD 牺牲了一些speed-up for higher quality}

\subsection{Knowledge, Suggestions and Future Direction}
\label{sec:data_summary}

%Among the data-level optimization field, prompting the LLM to organize its output content (which we denote as \textit{Output Organization}) is a quite promising and recommended way to accelerate the LLM inference process. 

% \todo{随着大语言模型能力需要处理更多的数据、输出更多的数据，data-level（input、output）的压缩更重要。随着大语言模型能力的提高，利用大语言模型本身做压缩input和organize output是一种promsing的新可能性}

%\todo{展望的部分不用事实，might/we believe @zzx}

The growing demand for LLMs to handle longer inputs and generate longer outputs highlights the importance of the data-level optimization techniques. Within these techniques, input compression methods primarily target enhancing the prefilling stage by diminishing the computational and memory cost resulting from the attention operation. Additionally, for API-based LLMs, these methods can reduce the API cost associated with input tokens. In contrast, output organization methods concentrate on optimizing the decoding stage by alleviating the substantial memory access cost associated with auto-regressive decoding approach. 

As LLMs become more and more capable, there is potential to utilize them to compress the input prompts or structure the output content. Recent advancements in output organization methods~\cite{ning2023skeleton,jin2024adaptive,liu2024apar} demonstrate the effectiveness of leveraging LLMs to organize the output content into independent points or a dependency graph, facilitating batch inference for improving generation latency. These methods capitalize on the inherent parallelizable structure within output content, enabling LLMs to perform parallel decoding to enhance hardware utilization and thereby reduce end-to-end generation latency.

% Despite their efficiency, these methods still face challenges in generalizing across different types of questions. For example, both SoT~\cite{ning2023skeleton} and SGD~\cite{jin2024adaptive} struggle with math and coding questions, whose answers are not suitable to be organized in parallel. Therefore, how to enhance the general applicability to different types of questions without compromising their answer quality is a problem worth exploring. One potential direction is designing dynamic output structure. Rather than ahead-of-time planning, dynamically deciding or adjusting output structure during the generation process could leverage both global thinking abilities (e.g., SoT~\cite{ning2023skeleton}, SGD~\cite{jin2024adaptive}) for efficiency enhancement and logical thinking abilities (e.g., CoT~\cite{wei2022chain}) for answer quality improvement. Another potential way is enabling LLMs to trigger the parallel decoding mode for only the suitable questions via fine-tuning. APAR~\cite{liu2024apar} makes the first attempt in this direction by instruct-tuning LLMs on general domain data that contains hierarchical tree structure. 

Recently, diverse prompting pipelines (e.g., ToT~\cite{yao2024tree}, GoT~\cite{besta2024graph}) and agent frameworks~\cite{xi2023rise,sun2023corex,guo2024large} are emerging. 
While these innovations enhance LLMs' capabilities, they also extend the length of inputs, leading to increased computational cost. To address this challenge, adopting input compression techniques to reduce input length shows promise as a solution. 
Simultaneously, these pipelines and frameworks naturally introduce more parallelism into output structures, offering increased potential for parallel decoding and key-value (KV) cache sharing across different decoding threads. SGLang~\cite{zheng2023efficiently} supports flexible LLM programming and offers opportunities for front-end and back-end co-optimization, laying the groundwork for further extensions and improvements in this area. 
%\todo{一方面，增加了LLM的输入长度}
%\todo{另一方面，输出的结构性更强，(1)可以并行 (2)可以share KV cache}
In summary, data-level optimization, including input compression and output organization techniques, would become increasingly necessary to enhance efficiency in the foreseeable future. 
In addition to optimizing the efficiency of existing frameworks, certain studies focus on designing more efficient agent frameworks directly. For example, FrugalGPT~\cite{chen2023frugalgpt} proposes a model cascade comprising LLMs of varying sizes, with the inference process being halted early if the model reaches a sufficient level of certainty regarding the answer. This approach aims to achieve efficiency by leveraging a tiered model architecture and intelligent inference termination based on model confidence estimation. Compared with model-level dynamic inference techniques (Sec.~\ref{sec:dynamic_inference}), FrugalGPT performs dynamic inference at the pipeline level.

\section{Model-level Optimization}
\label{sec:model-level-opt}

%In this section, we discuss model-level optimization, which mainly concentrates on optimizing the model structure or parameters of the inference-time architecture. One line of research focuses on designing a more efficient model structure or architecture. The LLMs normally require costly re-training or pre-training. Another line focuses on directly compressing the pre-trained LLMs. Typically, after compression, the models require a smaller amount of fine-tuning to recover their performance. 

The model-level optimization for LLM efficient inference mainly concentrates on optimizing the model structure or data representation. Model structure optimization involves directly designing efficient model structure, modifying the original model and adjusting the inference-time architecture. In terms of data representation optimization, the model quantization technique is commonly employed. 
%a commonly employed technique is model quantization. 

In this section, we categorize model-level optimization techniques based on the additional training overhead they require. The first category involves designing more efficient model structures (referred to as efficient structure design). Models developed using this approach typically require training from scratch. The second category focuses on compressing pre-trained models (referred to as model compression). Compressed models in this category generally require only minimal fine-tuning to restore their performance. 

%\todo{修改模型结构或数据表征。修改模型结构包含三种，直接设计、修改结构、修改inference-time结构，修改数据表征是量化。这一章是是按照retraining和基于pretrain两种分开讲}

\begin{figure*}[h]
\centering
\tikzset{
    basic/.style  = {draw, text width=2cm, align=center, font=\sffamily, rectangle},
    root/.style   = {basic, rounded corners=2pt, thin, align=center, fill=white,text width=8cm, rotate=90, font=\footnotesize},
    dnode/.style = {basic, thin, rounded corners=2pt, align=center, fill=yellow!30,text width=3.5cm, font=\footnotesize},
    dnode_1/.style = {basic, thin, rounded corners=2pt, align=center, fill=yellow!30,text width=2cm, font=\footnotesize},
    mnode/.style = {basic, thin, rounded corners=2pt, align=center, fill=nblue,text width=3.5cm, font=\footnotesize},
    mnode_1/.style = {basic, thin, rounded corners=2pt, align=center, fill=nblue,text width=2cm, font=\footnotesize}, 
    snode/.style = {basic, thin, rounded corners=2pt, align=center, fill=green!30,text width=3.5cm, font=\footnotesize},
    snode_1/.style = {basic, thin, rounded corners=2pt, align=center, fill=green!30,text width=2cm, font=\footnotesize},
    tnode/.style = {basic, thin, align=left, fill=pink!60, text width=15em, align=center},
    xnode/.style = {basic, thin, rounded corners=2pt, align=center, fill=blue!20,text width=5cm,},
    wnode/.style = {basic, thin, rounded corners=2pt, align=left, fill=white,text width=6cm, font=\footnotesize},
    wnode_1/.style = {basic, thin, rounded corners=2pt, align=left, fill=white,text width=3.3cm, font=\footnotesize},
    wnode_2/.style = {basic, thin, rounded corners=2pt, align=left, fill=white,text width=10cm, font=\footnotesize},
    %edge from parent/.style = {draw=black, edge from parent fork right}
    %edge from parent/.style = {draw=black, edge from parent fork down}
}
\begin{forest} 
for tree={
    grow=east,
    growth parent anchor=east,
    parent anchor=east,
    child anchor=west,
    edge path={\noexpand\path[\forestoption{edge},->, >={latex}] 
         (!u.parent anchor) -- +(5pt,0pt) |- (.child anchor)
         \forestoption{edge label};}
}
% l sep is used for arrow distance
[Efficient Structure Design, mnode_1
    [Transformer Alternates, mnode
        [Others, mnode
            [{SGConv~\cite{li2022makes}, CKConv~\cite{romero2021ckconv}, Hyena~\cite{poli2023hyena}, RWKV~\cite{peng2023rwkv}, RetNet~\cite{sun2023retentive}}, wnode]
        ]
        [SSM, mnode
            [{HiPPO~\cite{gu2020hippo}, LSSL~\cite{gu2021combining}, S4~\cite{gu2021efficiently}, DSS~\cite{gupta2022diagonal}, S4D~\cite{gu2022parameterization}, GSS~\cite{mehta2023long}, H3~\cite{fu2022hungry}, Liquid S4~\cite{hasani2022liquid}, S5~\cite{smith2022simplified}, BST~\cite{pilault2024block}, BiGS~\cite{wang2022pretraining}, Mamba~\cite{gu2023mamba}, MambaFormer~\cite{park2024can}}, wnode]
        ]
    ]
    [Efficient Attention Design, mnode
        [Multi/Group-Query Attention, mnode
            [{MQA~\cite{shazeer2019fast}, GQA~\cite{ainslie2023gqa}}, wnode]
        ]
        [Low-Complexity Attention, mnode
            [Low-Rank Attention, mnode_1
                [{Linformer~\cite{wang2020linformer}, LRT~\cite{winata2020lightweight}, FLuRKA~\cite{gupta2023flurka},Luna~\cite{ma2021luna}, Set Transformer~\cite{lee2019set}}, wnode_1]
            ]
            [Kernel-based Attention, mnode_1
                [{Linear Transformer~\cite{katharopoulos2020transformers}, Performers~\cite{choromanski2020rethinking}, RFA~\cite{peng2022random}, PolySketchFormer~\cite{kacham2023polysketchformer}}, wnode_1]
            ]
        ]
    ]
    [Efficient FFN Design, mnode
        [{Switch Transformers~\cite{fedus2022switch}, MoEfication~\cite{zhang2022moefication}, MPOE~\cite{gao2022parameter}, Sparse Upcycling~\cite{komatsuzaki2022sparse}, BASE~\cite{lewis2021base}, Expert Choice~\cite{zhou2022mixture}, SE-MoE~\cite{zoph2022st}, StableMoE~\cite{dai2022stablemoe}, SMoE-Dropout~\cite{chen2022sparse}, GLaM~\cite{du2022glam}, Mixtral 8x7B~\cite{jiang2024mixtral}}, wnode_2]
    ]
]
\end{forest}

\caption{Taxonomy of the efficient structure design for Large Language Models.}
\label{fig:taxonomy_efficient_structure}
\end{figure*}
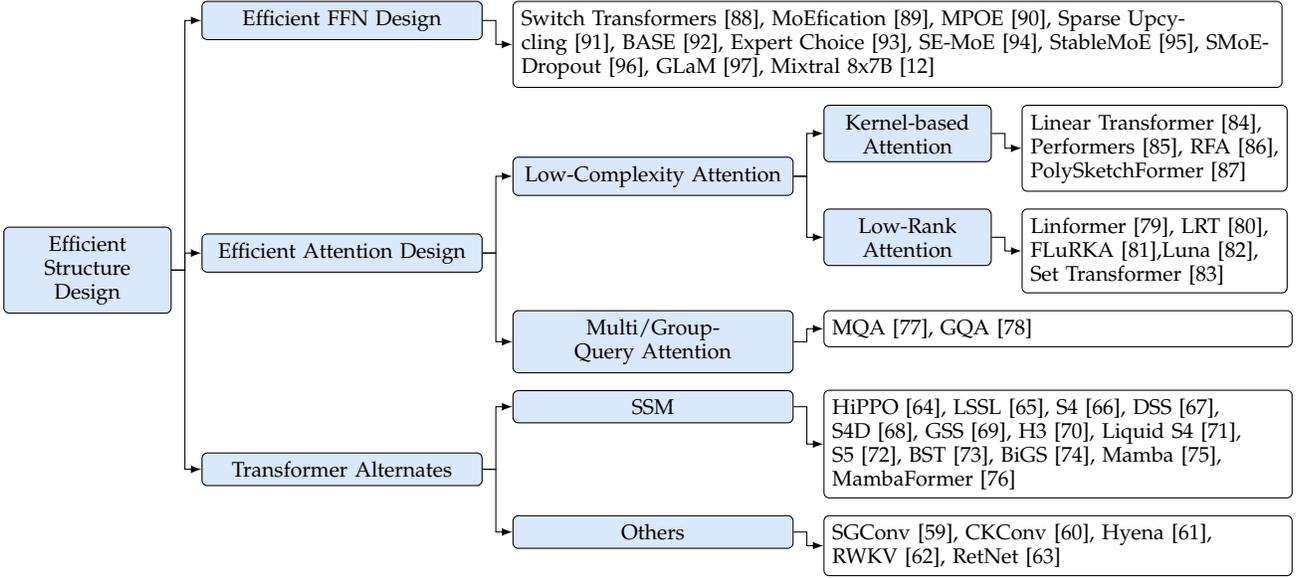

\subsection{Efficient Structure Design}
\label{sec:efficient_structure}

Currently, state-of-the-art LLMs commonly employ the Transformer architecture, as discussed in Section~\ref{sec:transformer}. However, the key components of Transformer-based LLMs, including the Feed Forward Network (FFN) and attention operation, present efficiency challenges during inference. We identify the causes as follows: 
\begin{itemize}
    \item The FFN contributes a substantial portion of the model parameters in Transformer-based LLMs, resulting in significant memory access cost and memory usage, particularly during the decoding stage. For instance, the FFN module accounts for 63.01\% of the parameters in the LLaMA-7B model and 71.69\% in the LLaMA-70B model. 
    \item The attention operation demonstrates quadratic complexity in the input length, leading to substantial computational cost and memory usage, especially when dealing with longer input contexts.
\end{itemize}
To tackle these efficiency challenges, several studies have concentrated on developing more efficient model structures. We categorize these studies into three groups (as depicted in Fig.~\ref{fig:taxonomy_efficient_structure}): efficient FFN design, efficient attention design, and Transformer alternates. %that adopt non-Transformer architectures to perform more efficient language modeling. 

%Specifically, in the prefilling stage, the attention operation causes the primary computational cost. As shown in Sec.~\ref{sec:efficiency}, the attention operation has quadratic computation 

%Specifically, FFN takes up a large portion of the model parameters, which causes large memory access cost and memory cost in the decoding stage. Meanwhile, attention operation has quadratic complexity in the input length, which causes large computational cost in the prefilling stage. 

%\todo{结构化下这段话，分点写FFN和attention的efficiency；写attention计算量大强调特别是当inputlen很长的时候}

\subsubsection{Efficient FFN Design}
\label{sec:moe}

In this field, many studies concentrate on integrating the Mixture-of-Experts (MoE) technique~\cite{shazeer2016outrageously} into LLMs to enhance their performance while maintaining the computational cost. The core idea of MoE is to dynamically allocate varying computational budgets to different input tokens. In MoE-based Transformers, multiple parallel Feed Forward Networks (FFNs), namely experts, are utilized alongside a trainable routing module. During inference, the model selectively activates specific experts for each token controlled by the routing module. 

%\todo{GPT-J https://arankomatsuzaki.wordpress.com/2021/06/04/gpt-j/}

% Some studies focus on how to construct the experts and the routing module. 
%\todo{expert目的：如何高效继承已有知识得到expert的权重，或让expert更轻量化}
%\todo{routing目的是什么：效果上防止capacity塌缩，效率上balance有什么好处}
Some researches concentrate on the construction of FFN expert, which mainly focus on optimizing the process of acquiring expert weights or making these experts more lightweight for efficiency. 
For instance, MoEfication~\cite{zhang2022moefication} devises a method to transform a non-MoE LLM into the MoE version using its pre-trained weights. This approach eliminates the need for expensive pre-training of the MoE model. To accomplish this, MoEfication first divides FFN neurons of the pre-trained LLM into multiple groups. Within each group, the neurons are commonly activated simultaneously by the activation function. Then, it restructures each group of neurons as an expert. 
Sparse Upcycling~\cite{komatsuzaki2022sparse} introduces a method to initialize the weights of MoE-based LLM directly from a dense model's checkpoint. In this approach, the experts within the MoE-based LLM are exact replicas of the FFN from the dense model. By employing this straightforward initialization, Sparse Upcycling can efficiently train the MoE model to achieve high performance. 
MPOE~\cite{gao2022parameter} proposes to reduce the parameters of MoE-based LLMs through Matrix Product Operators (MPO) decomposition. This method involves decomposing each weight matrix of the FFN into a global shared tensor containing common information and a set of local auxiliary tensors that capture specialized features. 

Another line of researches focuses on improving the design of the routing module (or strategy) within MoE models. In previous MoE models, the routing module often causes the load imbalance problem, which denotes that some experts are assigned a large number of tokens while the others handle only a few. This imbalance not only wastes the capacities of the under-utilized experts, which degrades model performance, but also degrades the inference efficiency. Current MoE implementations~\cite{lepikhin2020gshard,fedus2022switch,hwang2023tutel} often use batched matrix multiplication to compute all FFN experts simultaneously. This requires that the input matrices of each expert must have the same shape. However, since the load imbalance problem exists, input token sets for these under-utilized experts are needed to be padded to meet the shape constraint, resulting in a waste of computation. Therefore, the major aim of routing module design is achieving better balance in token assignment for MoE experts. 
Switch Transformers~\cite{fedus2022switch} introduces an additional loss, namely the load balancing loss, into the final loss function to penalize imbalanced assignments by the routing module. This loss is formulated as the scaled dot-product between the token assignment fraction vector and a uniform distribution vector. As a result, the loss is minimized only when the token assignment is balanced across all experts. This approach encourages the routing module to distribute tokens evenly among experts, promoting load balance and ultimately improving model performance and efficiency. 
BASE~\cite{lewis2021base} learns an embedding for each expert in an end-to-end manner and then assigns experts to tokens based on the similarity of their embeddings. To ensure load balance, BASE formulates a linear assignment problem and utilizes the auction algorithm~\cite{bertsekas1992auction} to solve this problem efficiently. 
Expert Choice~\cite{zhou2022mixture} introduces a simple yet effective strategy to ensure perfect load balance within MoE-based models. Unlike previous methods that assign experts to tokens, Expert Choice allows each expert to independently select the top-$k$ tokens based on their embedding similarities. This approach ensures that each expert handles a fixed number of tokens, even though each token might be assigned to a different number of experts. 

In addition to the aforementioned researches focusing on the model architecture itself, there are also studies that concentrate on improving the training methods for MoE-based models. 
SE-MoE~\cite{zoph2022st} introduces a new auxiliary loss called the router z-loss, which aims to enhance the stability of model training without compromising performance. SE-MoE identifies that the exponential functions introduced by softmax operations in the routing module can exacerbate roundoff errors, leading to training instability. To address this issue, the router z-loss penalizes large logits that are input into exponential functions, thereby minimizing roundoff errors during training. 
StableMoE~\cite{dai2022stablemoe} points out the routing fluctuation problem existing in the MoE-based LLMs, which denotes the inconsistency of the expert assignment in the training and inference stage. For the same input token, it is assigned to different experts along with training, but only activates one expert at inference time. To address this issue, StableMoE suggests a more consistent training approach. It first learns a routing strategy and then keeps it fixed during both the model backbone training and the inference stage.
SMoE-Dropout~\cite{chen2022sparse} designs a novel training method for MoE-based LLMs, which proposes to gradually increase the number of activated experts during the training process. This approach enhances the scalability of MoE-based models for inference and downstream fine-tuning. 
GLaM~\cite{du2022glam} pre-trains and releases a series of models with various parameter sizes, demonstrating their comparable performance to dense LLMs on few-shot tasks. The largest model in this family has a parameter size of up to 1.2 trillion. 
Mixtral 8x7B~\cite{jiang2024mixtral} is a remarkable recently released open-source model. During inference, it utilizes only 13 billion active parameters and achieves superior performance compared to the LLaMA-2-70B model across different benchmarks. Mixtral 8x7B consists of 8 Feed-Forward Network (FFN) experts in each layer, with each token assigned to two experts during inference.

\subsubsection{Efficient Attention Design}

The attention operation is a critical component in the Transformer architecture. However, its quadratic complexity in relation to input length leads to substantial computational cost, memory access cost, and memory usage, especially when dealing with long contexts. To address this issue, researchers are exploring more efficient approaches to approximate the functionality of the original attention operation. These studies can be broadly categorized into two main branches: multi-query attention and low-complexity attention. 

\noindent \textbf{Multi-Query Attention.} Multi-query attention (MQA)~\cite{shazeer2019fast} optimizes the attention operation by sharing the key (K) and value (V) cache across different attention heads. This strategy effectively reduces both memory access cost and memory usage during inference, contributing to improved efficiency in Transformer models. 
As introduced in Sec.~\ref{sec:inference}, the Transformer-style LLMs typically adopts multi-head attention (MHA) operation. This operation requires storing and retrieving K and V pairs for each attention head during the decoding stage, leading to substantial increases in memory access cost and memory usage. 
MQA tackles this challenge by using the same K and V pairs across different heads while maintaining distinct query (Q) values. Through extensive testing, it has been demonstrated that MQA significantly reduces memory requirements with only a minimal impact on model performance, making it a crucial strategy for enhancing inference efficiency. 
The concept of MQA is further extended by Grouped-query attention (GQA)~\cite{ainslie2023gqa}, which can be seen as a blend of MHA and MQA. Specifically, GQA segments the attention heads into groups, storing a single set of K and V values for each group. This method not only sustains the benefits of MQA in reducing memory overhead but also offers an enhanced balance between inference speed and output quality. 

\noindent \textbf{Low-Complexity Attention.} 
Low-complexity attention methods aim to design new mechanisms that reduce the computational complexity of each attention head. 
To simplify the discussion, we assume that the dimensions of the Q (query), K (key), and V (value) matrices are identical, with $Q, K, V \in \mathbb{R}^{n\times d}$. Since the following work does not involve altering the number of attention heads like MQA, our discussions focus on the attention mechanism within each head. As introduced in Section~\ref{sec:inference}, the computational complexity of the conventional attention mechanism scales as $\mathcal{O}(n^2)$, exhibiting quadratic growth with respect to the input length $n$. 
To address the inefficiency issue, kernel-based attention and low-rank attention methods are proposed to reduce the complexity to $\mathcal{O}(n)$.
\begin{packed_itemize}

\item \textbf{Kernel-based Attention}. Kernel-based attention designs kernel $\phi$ to approximate the non-linear softmax operation of ${\rm Softmax}(QK^T)$ with a linear dot product between kernel-transformed feature maps, i.e., $\phi(Q)\phi(K)^T$. It avoids the conventional quadratic computation associated with $QK^T \in \mathbb{R}^{n\times n}$ by prioritizing the computation of $\phi(K)^T V \in \mathbb{R}^{d\times d}$, followed by its multiplication with $\phi(Q) \in \mathbb{R}^{n\times d}$. 
Specifically, the input Q and K matrices are first mapped into kernel space using a kernel function $\phi$, while maintaining their original dimensions. Leveraging the associative property of matrix multiplication allows for the multiplication of K and V prior to their interaction with Q. The attention mechanism is reformulated as:
\begin{equation}
{\rm Softmax}(QK^T)V \approx \phi(Q)(\phi(K)^TV),
\end{equation}
where $\phi(Q), \phi(K) \in \mathbb{R}^{n\times d}$. This strategy effectively reduces the computational complexity to $\mathcal{O}(nd^2)$, rendering it linear with respect to the input length.
Linear Transformer~\cite{katharopoulos2020transformers} is the first work to propose the kernel-based attention. It adopts $\phi(x)={\rm elu}(x) + 1$ as the kernel function, where ${\rm elu}(\cdot)$ denotes the exponential linear unit activation function. 
Performers~\cite{choromanski2020rethinking} and RFA~\cite{peng2022random} proposes to use random feature projection to better approximate the softmax function.
PolySketchFormer~\cite{kacham2023polysketchformer} employs polynomial functions and sketching techniques to approximate the softmax function. 

% \todo{the same, which name do you want to use?}
% \item \textbf{Feature-Compressed Attention}. Feature-compressed attention refers to the methods that compress the long sequence of token features into a short one. Thus, the computational cost can be reduced.  
\item \textbf{Low-Rank Attention}. 
Low-Rank Attention technique employs compression on the token dimensions (i.e., $n$) of the K and V matrices to a smaller, fixed length (i.e., $k$) before performing the attention computation.
The approach is based on the insight that the $n \times n$ attention matrix often exhibits a low-rank property, making it feasible to compress it in the token dimension. 
The main focus of this line of researches is to design effective methods for the compression, where $X$ can be context matrix or K and V matrices:

\begin{equation}
    X \in \mathbb{R}^{n\times d} \rightarrow X' \in \mathbb{R}^{k\times d}.
\end{equation}

\ \ \quad One line of work uses linear projection to compress the token dimension. It is done by multiplying K and V matrices with projection matrices $P_k, P_v \in \mathbb{R}^{k\times n}$.
% \begin{equation}
%     {\rm Softmax}(QK^T)V \approx {\rm Softmax}(Q(EK)^T)(FV), 
% \end{equation}
In this way, the computational complexity of the attention operation is reduced to $\mathcal{O}(nkd)$, which is linear to the input length. 
Linformer~\cite{wang2020linformer} first observes and analyses the low-rank property of the attention map, and proposes the low-rank attention framework. 
LRT~\cite{winata2020lightweight} proposes to simultaneously apply low-rank transformation to both attention block and FFN to further improve the computational efficiency. 
FLuRKA~\cite{gupta2023flurka} combines the low-rank transformation and kernalization to the attention matrices to further improve the efficiency. Specifically, it first reduces the token dimension of K and V matrices, and then applies kernel function to the Q and low-rank K matrices. 

\ \ \quad Aside from linear projection, other token-dimension compression methods are also proposed.
Luna~\cite{ma2021luna} and Set Transformer~\cite{lee2019set} leverage additional attention computations alongside smaller queries to effectively compress the K and V matrices.
Luna~\cite{ma2021luna} involves an extra query matrix of fixed length $k$. The small query performs attention with the original context matrix, termed as pack attention, to compress the context matrix to size $\mathbb{R}^{k\times d}$. Subsequently, the regular attention, termed unpack attention, applies attention to the original Q matrices and the compressed K and V matrices. The extra query matrix can be learnable parameters or acquired from previous layers.
Set Transformer~\cite{lee2019set} designs the similar technique by introducing an \textit{inducing points} vector with fixed length. 
Unlike previous works that compress K and V, Funnel-Transformer~\cite{dai2020funnel} uses pooling operation to gradually compress the sequence length of the Q matrix. 
\end{packed_itemize}

\begin{table*}[tb]
\caption{Efficiency comparison of some novel non-Transformer models. Note that we denote $n$ as the input length and $d$ as the input dimension.}
\label{tab:transformer_alternate}
\begin{center}
\resizebox{0.98\textwidth}{!}
{
\begin{tabular}{c|ccc|ccc}%cp{1.6cm}<{\centering}p{1.6cm}<{\centering}p{1.6cm}<{\centering}p{1.6cm}<{\centering}p{1.6cm}<{\centering}}
\toprule

\multirow{2}{*}[-1ex]{Model} & \multirow{2}{*}[-1ex]{Training Form} & \multirow{2}{*}[-0.5ex]{\makecell[c]{Training Computational \\ Complexity}} & \multirow{2}{*}[-0.5ex]{\makecell[c]{Training Memory \\ Complexity}} & \multirow{2}{*}[-1ex]{Inference Form} & \multicolumn{2}{c}{Inference Computational Complexity} \\
\cmidrule(lr){6-7}
 & & & & & Prefilling & \makecell[c]{Decoding (per token)} 
 \\
\midrule
Transformer~\cite{vaswani2017attention} & Transformer-like & $\mathcal{O}(n^2d)$ & $\mathcal{O}(n^2+nd)$ & Transformer-like & $\mathcal{O}(n^2d)$ & $\mathcal{O}(nd)$ \\
S4~\cite{gu2021efficiently} & Convolution & $\mathcal{O}(nd^2\log n)$ & $\mathcal{O}(nd)$ & Recurrence & $\mathcal{O}(nd^2)$ & $\mathcal{O}(d^2)$ \\
Mamba~\cite{gu2023mamba} & Recurrence & $\mathcal{O}(nd^2\log n)$ & $\mathcal{O}(nd)$ & Recurrence & $\mathcal{O}(nd^2)$ & $\mathcal{O}(d^2)$ \\
Hyena~\cite{poli2023hyena} & Convolution & $\mathcal{O}(nd\log n)$ & $\mathcal{O}(nd)$ & Convolution & $\mathcal{O}(nd\log n)$ & $\mathcal{O}(nd\log n)$ \\
RetNet~\cite{sun2023retentive} & Transformer-like & $\mathcal{O}(n^2d)$ & $\mathcal{O}(n^2 + nd)$ & Recurrence & $\mathcal{O}(nd^2)$ & $\mathcal{O}(d^2)$ \\
RWKV~\cite{peng2023rwkv} & Recurrence & $\mathcal{O}(nd^2)$ & $\mathcal{O}(nd)$ & Recurrence & $\mathcal{O}(nd^2)$ & $\mathcal{O}(d^2)$ \\

\bottomrule
\end{tabular}
}
\end{center}
\end{table*}

\subsubsection{Transformer Alternates}
\label{sec:alternate}
%\todo{整体改成计算复杂度 computational complexity}

In addition to applying efficient techniques to the attention operation, recent studies have also innovated to design sequence modeling architectures that are efficient yet effective. Table~\ref{tab:transformer_alternate} compares the efficiency of some representative non-Transformer models. These architectures exhibit sub-quadratic computational complexity with respect to sequence length during both training and inference, enabling LLMs to significantly increase their context length.

Within this research field, two prominent lines of study have garnered significant attention. One line of studies concentrates on the State Space Model (SSM), which formulates sequence modeling as a recurrence transformation based on the HiPPO theory~\cite{gu2020hippo}. Additionally, other studies primarily focus on employing long convolutions or designing attention-like formulations to model sequences.

\noindent \textbf{State Space Model.} 
The State Space Model (SSM) has demonstrated competitive modeling capabilities in certain Natural Language Processing (NLP)~\cite{gu2023mamba} and and Computer Vision (CV)~\cite{zhu2024vision} tasks. Compared to attention-based Transformers, SSM exhibits linear computational and memory complexity with respect to the input sequence length, which enhances its efficiency in handling long-context sequences. In this survey, SSM refers to a series of model architectures that satisfy the following two properties: (1) They model sequence based on the following formulation proposed by HiPPO~\cite{gu2020hippo} and LSSL~\cite{gu2021combining}:
\begin{equation}
\begin{split}
    x_k & = \overline{\bm{A}}x_{k-1} + \overline{\bm{B}}u_{k}, \\
    y_k & = \overline{\bm{C}}x_{k}, 
\end{split}
\label{eq:lssl}
\end{equation}
where $\overline{\bm{A}}$, $\overline{\bm{B}}$ and $\overline{\bm{C}}$ denote the transition matrices, $x$ denotes the intermediate state and $u$ denotes the input sequence. (2) They design the transition matrix $A$ based on the HiPPO theory~\cite{gu2020hippo}. Specifically, HiPPO proposes to compress the input sequence into a sequence of coefficients (namely \textit{state}) by projecting it onto a set of polynomial bases. 

Building upon the aforementioned framework, several studies concentrate on improving the parameterization or initialization of the transition matrix $A$. This involves refining how the matrix is formulated or initialized within the SSM to enhance its effectiveness and performance in sequence modeling tasks. 
LSSL~\cite{gu2021combining} firstly proposes to initialize $A$ with the optimal transition matrix \textit{HiPPO-LegS} designed by HiPPO. In addition, LSSL also trains the SSM in a convolution manner by unrolling the Eq.~\ref{eq:lssl}. Specifically, through a convolution kernel defined as $\mathcal{K}_L(\bm{A}, \bm{B}, \bm{C})=(\bm{C}\bm{A}^i\bm{B})_{i\in [L]}=(\bm{C}\bm{B}, \bm{C}\bm{A}\bm{B}, ..., \bm{C}\bm{A}^{L-1}\bm{B})$, the Eq.~\ref{eq:lssl} can be rewritten as $y=\mathcal{K}_L(\overline{\bm{A}}, \overline{\bm{B}}, \overline{\bm{C}})*u$ and also can be computed efficiently via Fast Fourier Transform (FFT). 
However, computing this convolution kernel is expensive, since it requires multiple times of multiplication by $A$. To this end, S4~\cite{gu2021efficiently}, DSS~\cite{gupta2022diagonal} and S4D~\cite{gu2022parameterization} propose to diagonalize the matrix $A$, which can accelerate the computing. This can be seen as a parameterization technique to the transition matrix $A$. 
Previous SSMs processed each input dimension independently, resulting in a large number of trainable parameters. To enhance efficiency, S5~\cite{smith2022simplified} proposes to simultaneously process all input dimensions using a single set of parameters. Building upon this structure, S5 introduces a parameterization and initialization method for $A$ based on the standard HiPPO matrix. 
Liquid S4~\cite{hasani2022liquid} and Mamba~\cite{gu2023mamba} parameterize the transition matrices in a input-dependent manner, which further enhances the modeling capability of SSM. Additionally, both S5~\cite{smith2022simplified} and Mamba~\cite{gu2023mamba} adopt a parallel scan technique for efficient model training without the need for convolution operations. This technique offers advantages in implementation and deployment on modern GPU hardware. 

Another line of research aim to design better model architecture based on SSMs. 
GSS~\cite{mehta2023long} and BiGS~\cite{wang2022pretraining} combines the Gated Attention Unit (GAU)~\cite{hua2022transformer} with SSM. Specifically, they replace the attention operation in GAU with SSM operation. 
BST~\cite{pilault2024block} combines the SSM model with the proposed Block Transformer which introduces a strong local inductive bias. 
H3~\cite{fu2022hungry} observes that SSM is weak in recalling the earlier tokens and comparing a token across the sequence. To this end, it proposes to add a shift SSM operation before the standard SSM operation, which is used to directly shift the input tokens into the state. 
MambaFormer~\cite{park2024can} combines the standard Transformer and SSM model by substituting the FFN layer in the Transformer with an SSM layer. 
Jamba~\cite{ai212024jamba} introduces another approach to combining the Transformer and SSM models by adding four Transformer layers into an SSM model. 
DenseMamba~\cite{he2024densemamba} explores the issue of hidden state degradation in traditional SSMs and introduces dense connections within the SSM architecture to preserve fine-grained information across deeper layers of the model. 
BlackMamba~\cite{anthony2024blackmamba} and MoE-Mamba~\cite{pioro2024moe} propose to enhance SSM models with the Mixture-of-Experts (MoE) technique to optimize the training and inference efficiency while maintaining the model performance. 

\noindent \textbf{Other Alternates.} In addition to SSMs, several other efficient alternates have also garnered significant attention, including long convolution and attention-like recurrence operation. 

% Long convolution has been adopted in long sequence modeling in some studies~\cite{li2022makes,romero2021ckconv,poli2023hyena}. These efforts mainly focus on the parameterization of the convolution parameters. For example, Hyena~\cite{poli2023hyena} adopts a implicit parametrization for the long convolution via a shallow feed-forward neural
% network (FFN). Besides, it also introduces a element-wise multiplicative gating to control the parametrization by the input data. 
Several recent studies have applied long convolution in the context of modeling long sequences~\cite{li2022makes,romero2021ckconv,poli2023hyena}. These investigations primarily concentrate on refining the parameterization of the convolution kernel. For instance, Hyena~\cite{poli2023hyena} employs an data-dependent parameterization method for long convolutions using a shallow feed-forward neural network (FFN). 

Other studies~\cite{peng2023rwkv,sun2023retentive} aim to design the operation that has a similar form as the attention operation but can be enrolled to the recurrent manner, enabling both efficient training and efficient inference. 
For instance, RWKV~\cite{peng2023rwkv} builds upon AFT~\cite{zhai2021attention}, which proposes to substitute the attention operation in the Transformer model with the following equation: 
\begin{equation}
    Y_t = \sigma_q(Q_t)\odot \frac{\sum_{t^{'}=1}^{T} {\rm exp}(K_{t^{'}}+w_{t,t^{'}})\odot V_{t^{'}}}{\sum_{t^{'}=1}^{T} {\rm exp}(K_{t^{'}}+w_{t,t^{'}})}, 
    \label{eq:aft}
\end{equation}
where $Q$, $K$, and $V$ are the query, key, and value matrices as in Transformer, $w\in R^{T\times T}$ denotes a learnable pair-wise position bias and $\sigma_q(\cdot)$ denotes a non-linear function. Specifically, it further reparameterizes the position bias as $w_{t,t^{'}}=-(t-t^{'})w$, and thus can rewrite Eq.~\ref{eq:aft} in a recursive form. In this way, RWKV can combine the effective parallelizable training feature of Transformer and the efficient inference ability of RNN.

\noindent \textbf{Efficiency Analysis.} We analyze and compare the computational and memory complexity of several innovative and representative non-transformer architectures in Table \ref{tab:transformer_alternate}. In terms of training time, many studies (e.g., S4, Hyena, RetNet) aim to preserve training parallelism by adopting training forms such as the convolution or attention. Notably, Mamba utilizes parallel scan techniques for processing input sequences, thereby leveraging training parallelism as well. 

On the other hand, during inference, most studies opt for recurrent architectures to maintain linear computational complexity in the prefilling stage and to remain context length-agnostic in the decoding stage. Furthermore, in the decoding phase, these novel architectures eliminate the need to cache and load features of previous tokens (similar to the key-value cache in Transformer-based language models), resulting in significant memory access cost savings.

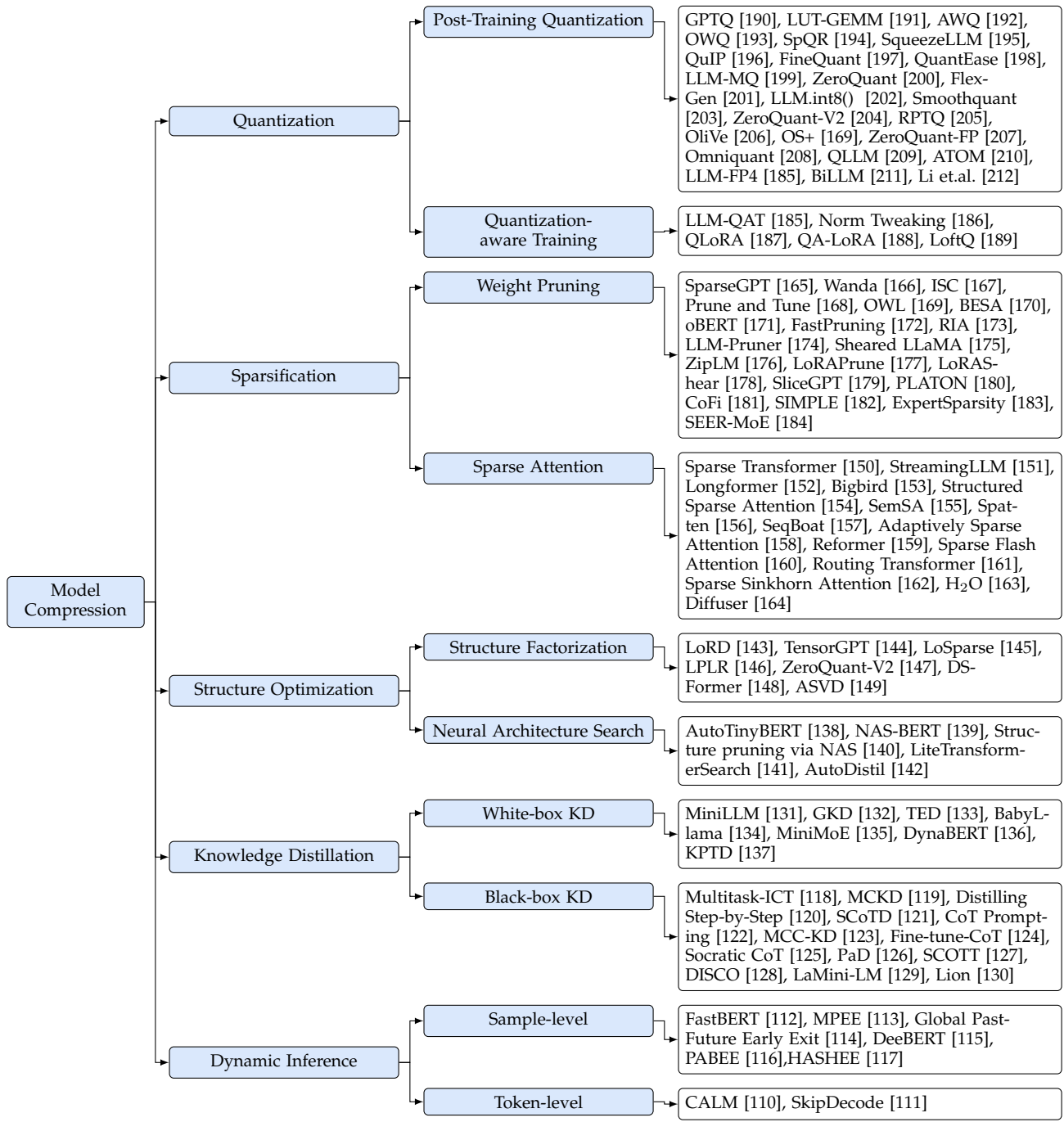
\begin{figure*}[h!]
\centering
\tikzset{
    basic/.style  = {draw, text width=2cm, align=center, font=\sffamily, rectangle},
    root/.style   = {basic, rounded corners=2pt, thin, align=center, fill=white,text width=8cm, rotate=90, font=\footnotesize},
    dnode/.style = {basic, thin, rounded corners=2pt, align=center, fill=yellow!30,text width=3.5cm, font=\footnotesize},
    dnode_1/.style = {basic, thin, rounded corners=2pt, align=center, fill=yellow!30,text width=2cm, font=\footnotesize},
    mnode/.style = {basic, thin, rounded corners=2pt, align=center, fill=nblue,text width=3.5cm, font=\footnotesize},
    mnode_1/.style = {basic, thin, rounded corners=2pt, align=center, fill=nblue,text width=2cm, font=\footnotesize}, 
    snode/.style = {basic, thin, rounded corners=2pt, align=center, fill=green!30,text width=3.5cm, font=\footnotesize},
    snode_1/.style = {basic, thin, rounded corners=2pt, align=center, fill=green!30,text width=2cm, font=\footnotesize},
    tnode/.style = {basic, thin, align=left, fill=pink!60, text width=15em, align=center},
    xnode/.style = {basic, thin, rounded corners=2pt, align=center, fill=blue!20,text width=5cm,},
    wnode/.style = {basic, thin, rounded corners=2pt, align=left, fill=white,text width=6cm, font=\footnotesize},
    wnode_1/.style = {basic, thin, rounded corners=2pt, align=left, fill=white,text width=3.3cm, font=\footnotesize},
    wnode_2/.style = {basic, thin, rounded corners=2pt, align=left, fill=white,text width=10cm, font=\footnotesize},
    %edge from parent/.style = {draw=black, edge from parent fork right}
    %edge from parent/.style = {draw=black, edge from parent fork down}
}
\begin{forest} 
for tree={
    grow=east,
    growth parent anchor=east,
    parent anchor=east,
    child anchor=west,
    edge path={\noexpand\path[\forestoption{edge},->, >={latex}] 
         (!u.parent anchor) -- +(5pt,0pt) |- (.child anchor)
         \forestoption{edge label};}
}
% l sep is used for arrow distance
[Model Compression, mnode_1
    [Dynamic Inference, mnode
        [Token-level, mnode
            [{CALM~\cite{schuster2022confident}, SkipDecode~\cite{del2023skipdecode}}, wnode]
        ]
        [Sample-level, mnode
            [{FastBERT~\cite{liu2020fastbert}, MPEE~\cite{kong2022accelerating}, Global Past-Future Early Exit~\cite{liao2021global}, DeeBERT~\cite{xin2020deebert}, PABEE~\cite{zhou2020bert},HASHEE~\cite{sun2022simple}}, wnode]
        ]
    ]
    [Knowledge Distillation, mnode
        [Black-box KD, mnode
            [{Multitask-ICT~\cite{huang2022context}, MCKD~\cite{zhao2023multistage}, Distilling Step-by-Step~\cite{hsieh2023distilling}, SCoTD~\cite{li2023symbolic}, CoT Prompting~\cite{magister2022teaching}, MCC-KD~\cite{chen2023mcc}, Fine-tune-CoT~\cite{ho2022large}, Socratic CoT~\cite{shridhar2023distilling}, PaD\cite{zhu2023pad}, SCOTT~\cite{wang2023scott}, DISCO~\cite{chen2023disco}, LaMini-LM~\cite{wu2023lamini}, Lion~\cite{jiang2023lion}}, wnode]
        ]
        [White-box KD, mnode
            [{MiniLLM~\cite{gu2023knowledge}, GKD~\cite{agarwal2023gkd}, TED~\cite{liang2023less}, BabyLlama~\cite{timiryasov2023baby}, MiniMoE~\cite{zhang2023lifting}, DynaBERT~\cite{hou2020dynabert}, KPTD~\cite{padmanabhan2023propagating}}, wnode]
        ]
    ]
    [Structure Optimization, mnode
        [Neural Architecture Search, mnode
            [{AutoTinyBERT~\cite{yin2021autotinybert}, NAS-BERT~\cite{xu2021bert}, Structure pruning via NAS~\cite{klein2023structural}, LiteTransformerSearch~\cite{javaheripi2022litetransformersearch}, AutoDistil~\cite{xu2022few}}, wnode]
        ]
        [Structure Factorization, mnode
            [{LoRD~\cite{kaushal2023lord}, TensorGPT~\cite{xu2023tensorgpt}, LoSparse~\cite{li2023losparse}, LPLR~\cite{saha2023matrix}, ZeroQuant-V2~\cite{yao2023zeroquant}, DSFormer~\cite{chand2023dsformer}, ASVD~\cite{yuan2023asvd}}, wnode]
        ]
    ]
    [Sparsification, mnode
        [Sparse Attention, mnode
            [{Sparse Transformer~\cite{spTrans}, StreamingLLM~\cite{xiao2023efficient}, Longformer~\cite{longformer}, Bigbird~\cite{bigbird}, Structured Sparse     Attention~\cite{structuredSparse}, SemSA~\cite{anonymous2023semsa}, Spatten~\cite{spatten}, SeqBoat~\cite{seqBoat}, Adaptively Sparse Attention~\cite{adaptiveSparse}, Reformer~\cite{kitaev2020reformer}, Sparse Flash Attention~\cite{pagliardini2023faster}, Routing Transformer \cite{roy2021efficient}, Sparse Sinkhorn Attention~\cite{tay2020sparse}, H$_2$O~\cite{zhang2024h2o}, Diffuser~\cite{diffuser}}, wnode]
        ]
        [Weight Pruning, mnode
            [{SparseGPT~\cite{frantar2023sparsegpt}, Wanda~\cite{sun2023simple}, ISC~\cite{shao2023one}, Prune and Tune~\cite{syed2023prune}, OWL~\cite{wei2023outlier}, BESA~\cite{xu2023besa}, oBERT~\cite{kurtic2022optimal}, FastPruning~\cite{kwon2022fast}, RIA~\cite{zhang2023efficient}, LLM-Pruner~\cite{ma2024llm}, Sheared LLaMA~\cite{xia2023sheared}, ZipLM~\cite{kurtic2024ziplm}, LoRAPrune~\cite{zhang2023loraprune}, LoRAShear~\cite{chen2023lorashear}, SliceGPT~\cite{ashkboos2024slicegpt}, PLATON~\cite{zhang2022platon}, CoFi~\cite{xia2022structured}, SIMPLE~\cite{tao2023structured}, ExpertSparsity~\cite{lu2024not}, SEER-MoE~\cite{muzio2024seer}, Pruner-Zero~\cite{dong2024pruner}, DS$\O$T~\cite{zhang2024dynamic}}, wnode]
        ]
    ]
    [Quantization, mnode
        [Quantization-aware Training, mnode
            [{LLM-QAT~\cite{liu2023llm}, Norm Tweaking~\cite{li2023norm}, QLoRA~\cite{dettmers2024qlora}, QA-LoRA~\cite{xu2023qa}, LoftQ~\cite{li2023loftq}}, wnode]
        ]
        [Post-Training Quantization, mnode
            [{GPTQ~\cite{frantar2022gptq}, LUT-GEMM~\cite{park2023lut}, AWQ~\cite{lin2023awq}, OWQ~\cite{lee2023owq}, SpQR~\cite{dettmers2023spqr}, SqueezeLLM~\cite{kim2023squeezellm}, QuIP~\cite{chee2023quip}, FineQuant~\cite{kim2023finequant}, QuantEase~\cite{behdin2023quantease}, LLM-MQ~\cite{li2023llm}, ZeroQuant~\cite{yao2022zeroquant}, FlexGen~\cite{sheng2023flexgen}, LLM.int8() ~\cite{dettmers2022llm}, Smoothquant ~\cite{xiao2023smoothquant}, ZeroQuant-V2~\cite{yao2023zeroquantv2}, RPTQ~\cite{yuan2023rptq}, OliVe~\cite{guo2023olive}, OS+~\cite{wei2023outlier}, ZeroQuant-FP~\cite{wu2023zeroquant}, Omniquant~\cite{shao2023omniquant}, QLLM~\cite{liu2023qllm}, ATOM~\cite{zhao2023atom}, LLM-FP4~\cite{liu2023llm}, BiLLM~\cite{huang2024billm}, Li et.al.~\cite{li2024evaluating}, AffineQuant~\cite{ma2024affinequant}, QuIP~\cite{chee2023quip}, QuIP\#~\cite{tseng2024quip}, QuaRot~\cite{ashkboos2024quarot}, SpinQuant~\cite{liu2024spinquant}}, wnode]
        ]
    ]
]
\end{forest}

\caption{Taxonomy of model compression methods for Large Language Models.}
\label{fig:taxonomy_model_compression}
\end{figure*}

\subsection{Model Compression}
\label{sec:model_compression}

Model compression encompasses a range of techniques designed to enhance the inference efficiency of a pre-trained model by modifying its data representation (e.g., quantization) or altering its architecture (e.g., sparsification, structural optimization, and dynamic inference), as depicted in Fig.~\ref{fig:taxonomy_model_compression}.

\begin{table*}[tb]
\caption{Summary of the representative studies on Post-Training Quantization. \textit{Quantized Tensor Type} denotes which parts of tensors are quantized. \textit{Data Format} denotes whether to adopt uniform or non-uniform quantization. \textit{Quantization Parameter Determination Scheme} denotes the how to decide the parameters (e.g., scaling factor, zero-point). \textit{Quantized Value Update} denotes whether to change the model weight (e.g., compensation, re-parameterization) during the quantization process.}
\label{tab:ptq}
\begin{center}
\resizebox{0.8\textwidth}{!}
{
\begin{tabular}{c|ccc|c|c|c}%cp{1.6cm}<{\centering}p{1.6cm}<{\centering}p{1.6cm}<{\centering}p{1.6cm}<{\centering}p{1.6cm}<{\centering}}
\toprule

\multirow{2}{*}[-1ex]{Model} & \multicolumn{3}{c|}{Quantized Tensor Type} & \multirow{2}{*}[-0.5ex]{\makecell[c]{Data \\ Format}} & \multirow{2}{*}[-0.5ex]{\makecell[c]{Quantization Parameter \\ Determination Scheme}} & \multirow{2}{*}[-0.5ex]{\makecell[c]{Quantized \\ Value Update}} \\
\cmidrule(lr){2-4}
 & Weight & Activation & KV Cache & & & \\
\midrule

GPTQ~\cite{frantar2022gptq} & \checkmark & & & Uniform & Statistic-based & \checkmark \\
LUT-GEMM~\cite{park2023lut} & \checkmark & & & Non-uniform & Statistic-based &  \\
AWQ~\cite{lin2023awq} & \checkmark & & & Uniform & Search-based & \checkmark \\
SqueezeLLM~\cite{kim2023squeezellm} & \checkmark & & & Non-uniform & Statistic-based &  \\
LLM.int8()~\cite{dettmers2022llm} & \checkmark & \checkmark & & Uniform & Statistic-based &  \\
SmoothQuant~\cite{xiao2023smoothquant} & \checkmark & \checkmark & & Uniform & Statistic-based & \checkmark \\
RPTQ~\cite{yuan2023rptq} & \checkmark & \checkmark & & Uniform & Statistic-based &  \\
OmniQuant~\cite{shao2023omniquant} & \checkmark & \checkmark & & Uniform & Search-based &  \\
FlexGen~\cite{sheng2023flexgen} & \checkmark & & \checkmark & Uniform & Statistic-based &  \\
Atom~\cite{zhao2023atom} & \checkmark & \checkmark & \checkmark & Uniform & Statistic-based &  \\
KVQuant~\cite{hooper2024kvquant} & & & \checkmark & Non-uniform & Statistic-based &  \\
KIVI~\cite{liu2024kivi} & & & \checkmark & Uniform & Statistic-based &  \\ 

\bottomrule
\end{tabular}
}
\end{center}
\end{table*}

\subsubsection{Quantization}
Quantization is a widely employed technique that reduces the computational and memory cost of LLMs by converting the models' weights and activations from high bit-width to low bit-width representations. Specifically, many methods involve quantizing FP16 tensors into low-bit integer tensors, which can be represented as follows: 
\begin{equation}
\label{eq:integer_quant_main}
    \textbf{X}_{\mathrm{INT}} = \left[ \frac{\textbf{X}_{\mathrm{FP16}}-Z}{S} \right],
\end{equation}
\begin{equation}
\label{eq:integer_quant_scale}
    S = \frac{\mathrm{max}(\textbf{X}_{\mathrm{FP16}})-\mathrm{min}(\textbf{X}_{\mathrm{FP16}})}{2^{N-1}-1},
\end{equation}
where $X_{\mathrm{FP16}}$ denotes the 16-bit floating-point (FP16) value, $X_{\mathrm{INT}}$ denotes the low-precision integer value, $N$ denotes the number of bits, and $S$ and $Z$ denote the scaling factor and zero-point.

In the following, we start with an efficiency analysis to illustrate how quantization techniques reduce the end-to-end inference latency of LLMs. Subsequently, we offer a detailed introduction to two distinct quantization workflows: Post-Training Quantization (PTQ) and Quantization-Aware Training (QAT), respectively.

\begin{figure}[h]
    \centering
    \includegraphics[width=0.90\linewidth]{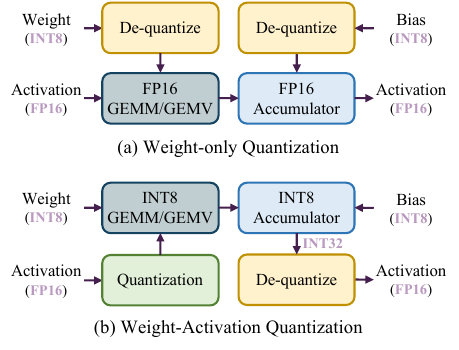}
    \caption{(a) The inference workflow of Weight-only Quantization. (b) The inference workflow of Weight-Activation Quantization.}
    \label{fig:quant_procedure}
\end{figure}

\noindent \textbf{Efficiency Analysis.}
As discussed in Section~\ref{sec:inference}, the inference process of LLMs involves two stages: the prefilling stage and the decoding stage. During the prefilling stage, LLMs typically handle long token sequences, and the primary operation is general matrix multiplication (GEMM). The latency of the prefilling stage is primarily constrained by the computation performed by high-precision CUDA Cores. To address this challenge, existing methods quantize both weights and activations to accelerate computation using low-precision Tensor Cores. As illustrated in Figure~\ref{fig:quant_procedure} (b), activation quantization is performed online before each GEMM operation, allowing computation with low-precision Tensor Cores (e.g., INT8). This quantization approach is referred to as \textit{Weight-Activation Quantization}.

In contrast, during the decoding stage, LLMs process only one token at each generation step using general matrix-vector multiplication (GEMV) as the core operation. The latency of the decoding stage is mainly influenced by the loading of large weight tensors. To tackle this challenge, existing methods focus on quantizing only the weights to accelerate memory access. This method, known as \textit{Weight-only Quantization}, involves offline quantization of weights, followed by de-quantization of the low-precision weights into FP16 format for computation, as shown in Figure~\ref{fig:quant_procedure} (a).

\noindent \textbf{Post-Training Quantization.} Post-training quantization (PTQ) involves quantizing pre-trained models without the need for retraining, which can be a costly process. While PTQ methods have been well-explored for smaller models, applying existing quantization techniques directly to LLMs presents challenges. This is primarily because the weights and activations of LLMs often exhibit more outliers and have a wider distribution range compared to smaller models, making their quantization more challenging. In summary, the complex nature of LLMs, characterized by their size and complexity, requires specialized approaches to effectively handle the quantization process. The presence of outliers and wider distribution ranges in LLMs necessitates the development of tailored quantization techniques that can account for these unique characteristics without compromising model performance or efficiency. 

Numerous studies have concentrated on developing effective quantization algorithms to compress LLMs. We provide a synthesis of representative algorithms categorized across four dimensions in Tab.~\ref{tab:ptq}. Regarding the types of quantized tensors, certain studies~\cite{frantar2022gptq,park2023lut,lin2023awq,kim2023squeezellm} concentrate on weight-only quantization, whereas many others~\cite{dettmers2022llm,xiao2023smoothquant,yuan2023rptq} focus on quantizing both weights and activations. Notably, in LLMs, the KV cache represents a distinctive component that impacts memory and memory access. Consequently, some investigations~\cite{sheng2023flexgen,zhao2023atom,hooper2024kvquant} propose KV cache quantization. 
Regarding data formats, the majority of algorithms adopt a uniform format for straightforward hardware implementation. Concerning the determination of quantized parameters (e.g., scale, zero-point), most studies rely on statistics derived from weight or activation values. Nevertheless, some research efforts~\cite{lin2023awq,shao2023omniquant} advocate for searching optimal parameters based on reconstruction loss. Furthermore, certain studies~\cite{frantar2022gptq,lin2023awq,xiao2023smoothquant} suggest updating unquantized weights (referred to as \textit{Quantized Value Update}) before or during the quantization process to enhance performance.

In weight-only quantization, GPTQ \cite{frantar2022gptq} represents an early advancement in LLM quantization, building upon the traditional algorithm OBQ \cite{frantar2022optimal}. OBQ utilizes an optimal quantization order per row of the weight matrix, guided by the reconstruction error relative to the Hessian matrix of unquantized weights. After each quantization step, OBQ iteratively adjusts the unquantized weights to mitigate reconstruction errors. However, the frequent updating of the Hessian matrix during quantization escalates computational complexity. GPTQ streamlines this process by adopting a uniform left-to-right order for quantizing each row, thus circumventing the need for extensive Hessian matrix updates. This strategy substantially reduces computational demands by computing the Hessian matrix solely during the quantization of one row, then leveraging the computing results for subsequent rows, expediting the overall quantization procedure. 
LUT-GEMM~\cite{park2023lut} presents a novel dequantization method utilizing a Look-Up Table (LUT), aiming to accelerate the inference process of quantized LLMs by reducing the dequantization overhead. Additionally, it adopts a non-uniform quantization approach known as Binary-Coding Quantization (BCQ), which incorporates learnable quantization intervals. 
AWQ~\cite{lin2023awq} observes that weight channels vary in importance for performance, particularly emphasizing those aligned with input channels exhibiting outliers in activations. To enhance the preservation of critical weight channels, AWQ utilizes a reparameterization method. This technique selects reparameterization coefficients via grid search to minimize reconstruction errors effectively. 
OWQ~\cite{lee2023owq} observes the difficulty of quantizing weights associated with activation outliers. To address this challenge, OWQ employs a mixed-precision quantization strategy. This method identifies weak columns in the weight matrix and allocates higher precision to these specific weights, while quantizing the rest of the weights at a lower precision level. 
SpQR~\cite{dettmers2023spqr} introduces a methodology where weight outliers are identified and allocated higher precision during quantization, while the rest of the weights are quantized to 3 bits. 
SqueezeLLM~\cite{kim2023squeezellm} proposes to store the outliers in a full-precision sparse matrix, and apply non-uniform quantization to the remaining weights. The values for non-uniform quantization are determined based on quantization sensitivity, which contributes to improved performance of the quantized model. 
QuIP~\cite{chee2023quip} introduces LDLQ, an optimal adaptive method for a quadratic proxy objective. The study reveals that ensuring incoherence between weight and Hessian matrices can enhance the effectiveness of LDLQ. QuIP utilizes LDLQ and achieves incoherence by employing random orthogonal matrix multiplication. 
FineQuant~\cite{kim2023finequant} utilizes a heuristic approach to determine the granularity of quantization per column, combining empirical insights gained from experiments to design a quantization scheme. 
QuantEase~\cite{behdin2023quantease} builds upon GPTQ. When quantizing each layer, it proposes a method based on Coordinate Descent to compensate for the unquantized weights more precisely. Additionally, QuantEase can leverage quantized weights from GPTQ as an initialization and further refine the compensation process. 
LLM-MQ~\cite{li2023llm} protects the weight outliers with FP16 format, and stores them in Compressed Sparse Row (CSR) format for efficient computation. Besides, LLM-MQ models the bit-width assignment to each layer as an integer programming problem, and employs an efficient solver to solve it within a few seconds. Moveover, LLM-MQ designs a efficient CUDA kernel to integrate dequantization operators, thereby reducing memory access cost during computation. 
Inspired by the equivalent transformations used in the previous PTQ methods, AffineQuant~\cite{ma2024affinequant} firstly introduces equivalent affine transformations in quantization, which extends the optimization scope and further reduces the quantization errors. 
Recently, many studies~\cite{chee2023quip,tseng2024quip,liu2024spinquant,ashkboos2024quarot} follows the computational invariance idea, by multiplying rotation matrices to the weight matrices and activation matrices. In this way, they can effectively eliminate the outliers in the weights and activations, thus help to quantize the LLMs. These studies use different rotation matrices. For example, QuaRot~\cite{ashkboos2024quarot} applies randomize Hadamard transformations to the weights and activations. SpinQuant~\cite{liu2024spinquant} finds the optimal rotation matrices by training on a small validation dataset. 
%\todo{spinquant, qserve, quarot}

\begin{table*}[htb]
\caption{Comparison of speed-ups in different scenarios (e.g., model size, batch size, input context length, inference framework) with W4A16 quantization based on TensorRT-LLM~\cite{tensorrt-llm} and LMDeploy~\cite{lmdeploy} framework, respectively. We test the speed-ups of prefilling/decoding/end-to-end latency on a single NVIDIA A100 GPU. OOM denotes ``Out Of Memory''.}
\label{tab:quant_comparison}
\begin{center}
\resizebox{0.75\textwidth}{!}
{
\begin{tabular}{ccccccc}%cp{1.6cm}<{\centering}p{1.6cm}<{\centering}p{1.6cm}<{\centering}p{1.6cm}<{\centering}p{1.6cm}<{\centering}}
\toprule
 & & \multicolumn{5}{c}{TensorRT-LLM} \\
 \cmidrule(lr){3-7} 
& B & 128 & 256 & 512 & 1024 & 2048 \\
\midrule
\multirow{5}{*}{LLaMA-2-7B} 
& 1 & 1.06/2.40/2.37 & 0.90/2.38/2.34 & 0.92/2.30/2.28 & 0.88/2.19/2.17 & 0.91/2.00/1.98  \\
& 2 & 0.88/2.10/2.05 & 0.91/2.07/2.04 & 0.89/2.01/1.98 & 0.91/1.92/1.89 & 0.88/1.78/1.76  \\
& 4 & 0.92/1.72/1.67 & 0.89/1.67/1.64 & 0.90/1.61/1.58 & 0.87/1.53/1.51 & 0.84/1.42/1.40  \\
& 8 & 0.91/1.43/1.36 & 0.88/1.38/1.33 & 0.83/1.33/1.28 & 0.77/1.25/1.21 & 0.78/1.16/1.14 \\
& 16 & 0.91/1.43/1.36 & 0.88/1.38/1.33 & 0.83/1.33/1.28 & 0.77/1.25/1.21 & 0.78/1.16/1.14 \\
\midrule

& B & 128 & 256 & 512 & 1024 & 2048 \\
\midrule
\multirow{5}{*}{LLaMA-2-13B} 
& 1 & 1.24/2.51/2.50 & 0.89/2.45/2.47 & 0.94/2.34/2.42 & 0.90/2.18/2.32 & 0.83/1.94/2.16  \\
& 2 & 0.90/2.51/2.50 & 0.95/2.45/2.47 & 0.90/2.34/2.42 & 0.83/2.18/2.32 & 0.80/1.94/2.16  \\
& 4 & 0.96/1.80/1.76 & 0.91/1.78/1.74 & 0.83/1.73/1.69 & 0.80/1.65/1.62 & 0.83/1.54/1.52  \\
& 8 & 0.91/1.86/1.77 & 0.83/1.81/1.73 & 0.80/1.73/1.66 & 0.82/1.62/1.56 & 0.75/1.46/1.41  \\
& 16 & 0.84/1.84/1.69 & 0.81/1.77/1.63 & 0.82/1.63/1.53 & 0.78/1.46/1.39 & OOM \\

\midrule

 & & \multicolumn{5}{c}{LMDeploy} \\
 \cmidrule(lr){3-7}  
& B & 128 & 256 & 512 & 1024 & 2048  \\
\midrule
\multirow{5}{*}{LLaMA-2-7B} 
& 1 & 1.30/2.11/2.09 & 0.94/2.07/2.05 & 0.90/2.03/2.02 & 0.88/1.97/1.96 & 0.94/1.92/1.91 \\
& 2 & 1.03/2.24/2.20 & 0.90/2.19/2.15 & 0.88/2.11/2.08 & 0.93/1.97/1.95 & 0.85/1.78/1.76 \\
& 4 & 0.90/2.18/2.10 & 0.87/2.12/2.05 & 0.93/2.01/1.96 & 0.92/1.86/1.83 & 0.92/1.64/1.62 \\
& 8 & 0.92/1.92/1.77 & 0.91/1.82/1.71 & 0.92/1.65/1.57 & 0.93/1.45/1.41 & 0.94/1.28/1.26 \\
& 16 & 0.92/1.92/1.77 & 0.91/1.82/1.71 & 0.92/1.65/1.57 & 0.93/1.45/1.41 & 0.94/1.28/1.26 \\
\midrule

& B & 128 & 256 & 512 & 1024 & 2048 \\
\midrule
\multirow{5}{*}{LLaMA-2-13B} 
& 1 & 1.32/2.34/2.32 & 0.94/2.31/2.28 & 0.92/2.22/2.20 & 0.94/2.15/2.13 & 0.94/2.01/1.99 \\
& 2 & 1.06/2.42/2.36 & 0.92/2.37/2.32 & 0.94/2.29/2.25 & 0.94/2.15/2.12 & 0.95/1.95/1.93 \\
& 4 & 0.93/2.36/2.26 & 0.94/2.29/2.21 & 0.94/2.18/2.12 & 0.95/2.01/1.97 & 0.96/1.78/1.75 \\
& 8 & 0.92/2.24/2.10 & 0.93/1.93/2.02 & 0.94/1.81/1.89 & 0.94/1.65/1.71 & 0.95/1.45/1.49 \\
& 16 & 0.93/2.02/1.85 & 0.94/1.90/1.76 & 0.94/1.73/1.63 & 0.95/1.50/1.45 & OOM \\

\bottomrule
\end{tabular}
}
\end{center}
\end{table*}

For weight-activation quantization, ZeroQuant~\cite{yao2022zeroquant} employs finer-grained quantization for weights and activations, leveraging kernel fusion to minimize the memory access cost during quantization and conducting layer-by-layer knowledge distillation to recover the performance. 
FlexGen~\cite{sheng2023flexgen} quantizes weights and KV cache directly into INT4 to reduce the memory footprint during inference with large batch sizes. 
LLM.int8()~\cite{dettmers2022llm} identifies that outliers in activations are concentrated within a small subset of channels. Leveraging this insight, LLM.int8() splits activations and weights into two distinct parts based on the outlier distribution within input channels to minimize quantization errors in activations. Channels containing outlier data in both activations and weights are stored in FP16 format, while other channels are stored in INT8 format. 
SmoothQuant~\cite{xiao2023smoothquant} employs a reparameterization technique to address the challenges of quantizing activation values. This method introduces a scaling factor that expands the data range of weight channels while shrinking the data range of corresponding activation channels. 
ZeroQuant~\cite{yao2022zeroquant} introduces a group-wise quantization strategy for weights and a token-wise quantization approach for activations. Building upon this methodology, ZeroQuant-V2~\cite{yao2023zeroquantv2} presents the LoRC (Low-Rank Compensation) technique, employing low-rank matrices to mitigate quantization inaccuracies. 
RPTQ~\cite{yuan2023rptq} identifies substantial variations in the distribution of different activation channels, which present challenges for quantization. To mitigate this issue, RPTQ reorganizes channels with similar activation distributions into clusters and independently applies quantization within each cluster. 
OliVe~\cite{guo2023olive} observes that the normal values neighboring to the outliers are less critical. Therefore, it pairs each outlier with a normal value, sacrificing the latter to achieve a broader representation range for outliers. 
OS+~\cite{wei2023outlier} observes that the distribution of outliers is concentrated and asymmetrical, posing a challenge to LLM quantization. To address this, OS+ introduces a channel-wise shifting and scaling technique aimed at alleviating these challenges. The shifting and scaling parameters are determined through a search process to effectively handle the concentrated and asymmetrical outlier distribution. 
ZeroQuant-FP~\cite{wu2023zeroquant} investigates the feasibility of quantizing weight and activation values into FP4 and FP8 formats. The study reveals that quantizing activations into floating-point types (FP4 and FP8) produces superior results compared to integer types. 
Omniquant~\cite{shao2023omniquant} diverges from prior approaches that rely on empirical design of quantization parameters. Instead, it optimizes the boundaries for weight clipping and the scaling factor for equivalent transformation to minimize quantization errors.  
QLLM~\cite{liu2023qllm} addresses the impact of outliers on quantization by implementing channel reassembly. Additionally, it introduces learnable low-rank parameters to minimize quantization errors in the post-quantized model. 
Atom~\cite{zhao2023atom} employs a strategy involving mixed-precision and dynamic quantization for activations. Notably, it extends this approach to quantize the KV cache into INT4 to enhance throughput performance. 
LLM-FP4~\cite{liu2023llm} endeavors to quantize the entire model into FP4 format and introduces a pre-shifted exponent bias technique. This approach combines the scaling factor of activation values with weights to address quantization challenges posed by outliers. 
BiLLM~\cite{huang2024billm} represents one of the lowest-bit PTQ efforts to date. BiLLM identified the bell-shaped distribution of weights and the exceptionally long-tail distribution of weights' Hessian matrix. Based on this, it proposes to categorize weights into salient and non-salient values structurally based on the Hessian matrix and binarizes them separately. As a result, BiLLM can extensively quantize LLMs to 1.08 bits without significant degradation in perplexity. 
KVQuant~\cite{hooper2024kvquant} proposes a non-uniform quantization scheme for KV cache quantization, by deriving the optimal datatypes offline on a calibration set. 
KIVI~\cite{liu2024kivi} proposes a tuning-free 2bit KV cache quantization algorithm, which utilizes per-channel quantization for key cache and per-token quantization for value cache in a group-wise manner. 
Li et al.~\cite{li2024evaluating} conducted a thorough evaluation to assess the impact of quantization on different tensor types (including KV Cache), various tasks, 11 LLM families, and SOTA quantization methods.

\noindent \textbf{Quantization-Aware Training.}
Quantization-aware training (QAT) incorporates the influence of quantization within the model training procedure. By integrating layers that replicate quantization effects, this approach facilitates weight adaptation to quantization-induced errors, leading to enhanced task performance. Nevertheless, training LLMs typically demands substantial training data and considerable computational resources, posing potential bottlenecks for QAT implementation. Consequently, current research endeavors focus on strategies to reduce the training data requirements or alleviate the computational burden associated with QAT implementation. 

To reduce the data requirements, LLM-QAT~\cite{liu2023llm} introduces a data-free method to generate the training data by using the original FP16 LLMs. Specifically, LLM-QAT uses every token in the tokenization vocabulary as a starting token to generate sentences. Based on the generated training data, LLM-QAT applies a distillation-based workflow to train the quantized LLM to match the output distribution of the original FP16 LLM. 
Norm Tweaking~\cite{li2023norm} limits the selection of the starting token to only those language categories listed among the top languages with the highest proportion. This strategy can effectively improve the generalization of the quantized model on different tasks.

To reduce the computation cost, many methods apply parameter-efficient tuning (PEFT) strategies to accelerate QAT. 
QLoRA~\cite{dettmers2024qlora} quantizes the weights of LLMs into 4-bit and subsequently employs LoRA~\cite{hu2021lora} in BF16 for each 4-bit weight matrix to fine-tune the quantized model. QLoRA allows for the efficient fine-tuning of a 65B parameter LLM on one GPU with only 30GB of memory.
QA-LoRA~\cite{xu2023qa} proposes to incorporate group-wise quantization into QLoRA. The authors observe that the number of quantization parameters in QLoRA is significantly smaller than the number of LoRA parameters, leading to an imbalance between quantization and low-rank adaptation. They suggest that group-wise operations can address this issue by increasing the number of parameters dedicated to quantization. In addition, QA-LoRA can merge the LoRA terms into the corresponding quantized weight matrices.
LoftQ~\cite{li2023loftq} identifies that initializing LoRA matrices with zeros in QLoRA is inefficient for downstream tasks. As an alternative, LoftQ suggests initializing the LoRA matrices using the Singular Value Decomposition (SVD) of the difference between the original FP16 weights and quantized weights. LoftQ iteratively applies quantization and SVD to achieve a more accurate approximation of the original weights.
Norm Tweaking~\cite{li2023norm} proposes to train the LayerNorm layer after quantization and use knowledge distillation to match the output distribution of the quantized model with that of the FP16 model, achieving effects similar to LLM-QAT while avoiding high training costs. 

\noindent \textbf{Comparative Experiments and Analysis.} In this section, we conduct experiments to evaluate the speed-ups achieved by employing the weight-only quantization technique in various scenarios. Specifically, we focus on two widely-used large language models (LLMs), LLaMA-2-7B and LLaMA-2-13B, and quantize their weights to 4-bit using the AWQ~\cite{lin2023awq} algorithm. Subsequently, we deploy these quantized models on a single NVIDIA A100 GPU using two different inference frameworks: TensorRT-LLM~\cite{tensorrt-llm} and LMDeploy~\cite{lmdeploy}. We then evaluate the speed-ups achieved by these frameworks across different input sequences characterized by varying batch sizes and context lengths. 

We present the speed-ups of prefilling latency, decoding latency, and end-to-end latency, as summarized in Tab.~\ref{tab:quant_comparison}. From the results, several key observations can be made: 
(1) Weight-only quantization can substantially accelerate the decoding stage, leading to improvements in end-to-end latency. This enhancement primarily stems from the capability of loading the quantized model with low-precision weight tensors much more swiftly from the High Bandwidth Memory (HBM), as illustrated in the preceding ``Efficient Analysis'' part. Consequently, this approach markedly diminishes the memory access overhead. 
(2) Regarding the prefilling stage, weight-only quantization may actually increase the latency. This is due to the fact that the bottleneck in the prefilling stage is the computational cost rather than the memory access cost. Therefore, quantizing only the weights without the activations has minimal impact on latency. Additionally, as illustrated in Fig.~\ref{fig:quant_procedure}, weight-only quantization necessitates the de-quantization of low-precision weights to FP16, leading to additional computational overhead and consequently slowing down the prefilling stage. 
(3) As the batch size and input length increase, the extent of speed-up achieved by weight-only quantization gradually diminishes. This is primarily because, with larger batch sizes and input lengths, the computational cost constitutes a larger proportion of latency. While weight-only quantization predominantly reduces memory access cost, its impact on latency becomes less significant as the computational demands become more prominent with larger batch sizes and input lengths. 
(4) Weight-only quantization offers greater benefits for larger models due to the significant memory access overhead associated with larger model sizes. As models grow in complexity and size, the amount of memory required to store and access weights increases proportionally. By quantizing the model weights, weight-only quantization effectively reduces this memory footprint and memory access overhead.

\subsubsection{Sparsification}
\label{sec:sparse}

% overview of sparse
% [refer to chap 3.2.4]
Sparsification is a compression technique that increases the proportion of zero-valued elements in data structures such as model parameters or activations. This method aims to decrease computational complexity and memory usage by efficiently ignoring zero elements during computation. 
In the context of LLMs, sparsification is commonly applied to weight parameters and attention activations. It leads to the development of weight pruning strategies and sparse attention mechanisms.

\begin{figure}[h]
    \centering
    \includegraphics[width=0.90\linewidth]{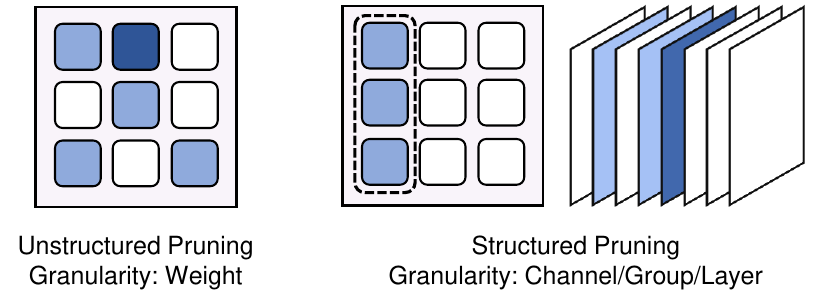}
    \caption{Illustration of Unstructured Pruning (left) and Structured Pruning (right).}
    \label{fig:pruning}
\end{figure}

\noindent \textbf{Weight Pruning.} Weight pruning systematically removes less critical weights and structures from models, aiming to reduce computational and memory cost during both prefilling stages and decoding stages without significantly compromising performance. 
This sparsification approach is categorized into two main types: unstructured pruning and structured pruning. The categorization is based on the granularity of the pruning process, as illustrated in Figure~\ref{fig:pruning}. 

Unstructured pruning prunes individual weight values with fine granularity. Compared with structured pruning, it typically achieves a greater level of sparsity with minimal impact on model prediction.
However, the sparse pattern achieved through unstructured pruning lacks high-level regularity, leading to irregular memory access and computation patterns. 
This irregularity can significantly hinder the potential for hardware acceleration, as modern computing architectures are optimized for dense, regular data patterns. Consequently, despite achieving higher sparsity levels, the practical benefits of unstructured pruning in terms of hardware efficiency and computational speedup may be limited.

The common focus of this line of work is the pruning criterion, including the weight importance and pruning ratio. Considering the huge parameter size of LLMs, improving the pruning efficiency is also crucial.
One pruning criterion is to minimize the reconstruction loss of the model.
SparseGPT~\cite{frantar2023sparsegpt} is a representative approach in this field. It follows the idea of OBS~\cite{hassibi1993optimal}, which considers the impact of removing each weight on the network's reconstruction loss. OBS iteratively decides a pruning mask to prune the weights and reconstructs the unpruned weights to compensate for the pruning loss. SparseGPT overcomes the efficiency bottleneck of OBS via the Optimal Partial Updates technique, and designs an adaptive mask selection technique based on the OBS reconstruction error. 
Prune and Tune~\cite{syed2023prune} improves upon SparseGPT by fine-tuning the LLMs with minimal training steps during pruning. 
ISC~\cite{shao2023one} designs a novel pruning criterion by combining the saliency criteria in OBS~\cite{hassibi1993optimal} and OBD~\cite{lecun1989optimal}. It further assigns non-uniform pruning ratios to each layer based on Hessian information. 
oBERT~\cite{kurtic2022optimal} and FastPruning~\cite{kwon2022fast} utilizes the second-order information of the loss function to decide the pruned weights. 
BESA~\cite{xu2023besa} learns a differentiable binary mask via gradient descent of the reconstruction loss. 
The pruning ratio for each layer is sequentially decided by minimizing the reconstruction error.
The other popular pruning criterion is magnitude-based.
Wanda~\cite{sun2023simple} proposes to use the element-wise product between the weight magnitude and the norm of input activation as the pruning criterion. 
RIA~\cite{zhang2023efficient} jointly considers the weights and activations by using the metric of Relative Importance and Activations, which evaluates the importance of each weight element based on all its connected weights.
In addition, RIA converts the unstructured sparsity pattern to a structured N:M sparsity pattern, which can enjoy the actual speed-up on NVIDIA GPUs. 
The recent study, Pruner-Zero~\cite{dong2024pruner}, proposes to automatically identify the optimal pruning metric for LLMs, going beyond the hand-designed matrics. As a result, the optimal metric tailored for LLaMA and LLaMA-2 is $\pmb{W}\odot\pmb{W}\odot\sigma(\pmb{G})$, where $\pmb{W}$ and $\pmb{G}$ represent the weights and gradients, and $\sigma(\cdot)$ scales a tensor to [0,1] using its mininum and maximum value.
Additionally, OWL~\cite{wei2023outlier} focuses on deciding the pruning ratio of each layer. It assigns the pruning ratios to each layer based on its activation outlier ratios. 
DS$\O$T~\cite{zhang2024dynamic} proposes a training-free appoarch to fine-tune the pruned LLMs. It builds upon the ``pruning-and-growing'' workflow adopted in Dynamic Sparse Training~\cite{mocanu2018scalable}, which first prunes the model and then iteratively adjusts the network topology without training or weight update. DS$\O$T further designs the pruning and growing metrics tailored for LLMs.

Structured pruning prunes larger structural units of the model, such as entire channels or layers, operating at a coarser granularity compared to unstructured pruning. 
These methods directly facilitate inference speed-up on conventional hardware platforms due to their alignment with the dense, regular data patterns these systems are optimized to process. However, the coarse granularity of structured pruning often results in a more pronounced impact on model performance. 
The pruning criterion of this line of work additionally enforces the structured pruning pattern.
LLM-Pruner~\cite{ma2024llm} proposes a task-agnostic structured pruning algorithm. Specifically, it first identifies the couple structures in the LLM, based on the connection dependencies between neurons. Then, it decides which structure groups to remove based on a well-designed group-wise pruning metric. After pruning, it further proposes to recover the model performance by a parameter-efficient training technique, i.e., LoRA~\cite{hu2021lora}. 
Sheared LLaMA~\cite{xia2023sheared} proposes to prune the original LLM to a specific target architecture of existing pre-trained LLMs. In addition, it designs dynamic batch-loading techniques to improve post-training performance. 
ZipLM~\cite{kurtic2024ziplm} iteratively identifies and prunes the structural components with the worst trade-off between loss and runtime. 
LoRAPrune~\cite{zhang2023loraprune} proposes a structured pruning framework for the pre-trained LLMs with LoRA modules to enable fast inference of LoRA-based models. It designs a LoRA-guided pruning criterion that uses the weights and gradients of LoRA, and an iterative pruning scheme to remove the unimportant weights based on the criterion. 
LoRAShear~\cite{chen2023lorashear} also designs a pruning method for LoRA-based LLMs with (1) a graph algorithm to identify the minimal removal structures, (2) a progressive structured pruning algorithm LHSPG, and (3) a dynamic knowledge recovery mechanism to recover the model performance. 
SliceGPT~\cite{ashkboos2024slicegpt} builds on the idea of computational invariance of RMSNorm operation. It proposes to structurally arrange the sparsity in each weight matrix, and to slice out the entire rows or columns. 
PLATON~\cite{zhang2022platon} proposes to prune the weights by considering both their importance and uncertainty. It uses the exponential moving average (EMA) of the importance scores to estimate the importance, and adopts the upper confidence bound (UCB) for the uncertainty. 
CoFi~\cite{xia2022structured} and SIMPLE~\cite{tao2023structured} propose to prune the attention head, FFN neurons and hidden dimension via learning the corresponding sparsity masks. After pruning, they further adopt knowledge distillation to fine-tune the pruned models for performance recovery. 
MoE techniques (Sec.~\ref{sec:moe}) have attracted much attention in the field of efficient LLMs. Recent studies tend to explore the expert pruning methods for MoE-based LLMs. For example, ExpertSparsity~\cite{lu2024not} proposes to prune some less important FFN experts in each model layer. Specifically, it utilizes the Frobenius norm of the difference between the original output and the output of the pruned layer to quantify the loss of pruned experts. In constrast, SEER-MoE~\cite{muzio2024seer} uses the total number of times that one expert gets activated on a calibration dataset, to quantify this expert's importance.

% \todo{knowledge: 1. qk insensitive, vo sensitive}
\begin{figure}[h]
    \centering
    \includegraphics[width=0.90\linewidth]{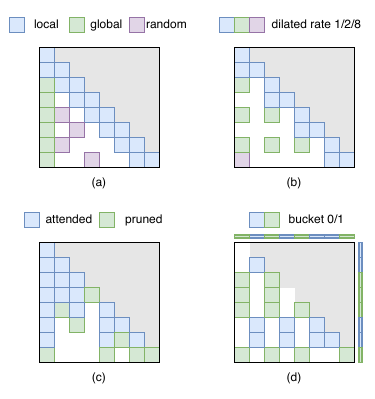}
    \caption{Examples of different sparse attention masks. (a) Static mask with local, global, and random attention pattern. (b) Static mask with dilated attention pattern of different dilated rate. (c) Dynamic token pruning. (d) Dynamic attention pruning.}
    \label{fig:sa}
\end{figure}
\noindent \textbf{Sparse Attention.} 
Sparse attention techniques in Multi-Head Self-Attention (MHSA) components of transformer models strategically omit certain attention calculations to enhance computational efficiency of the attention operation mainly in the prefilling stage. These mechanisms diverge into static and dynamic categories based on their reliance on specific input data. 

Static sparse attention removes activation values independently of specific inputs~\cite{longformer, spTrans, bigbird, structuredSparse}.
These methods pre-determine the sparse attention mask and enforce it on the attention matrix during inference. 
Previous studies combine different sparse patterns to preserve the most essential elements within each attention matrix.
As shown in Figure~\ref{fig:sa}(a), the most common sparse attention patterns are the local and global attention patterns. The local attention pattern captures the local context of each token with a fixed-size window attention surrounding each token. The global attention pattern captures the correlation of specific tokens to all other tokens by computing and attending to all tokens across the sequence. Note that leveraging global patterns can eliminate the need to store key-value (KV) pairs for unused tokens, thereby reducing memory access cost and memory usage during the decoding stage. 
Sparse Transformer~\cite{spTrans} combines these patterns to capture the local context with a local pattern, and then aggregates the information with the global pattern for every few words.
StreamingLLM~\cite{xiao2023efficient} applies the local pattern, along with the global pattern only for the first few tokens. It shows that such a global pattern serves as the attention sink to keep the strong attention scores toward initial tokens. It helps the LLMs to generalize to infinite input sequence length.
Bigbird~\cite{bigbird} also uses the random pattern, where all tokens attend to a set of random tokens. The combination of local, global and random patterns is proven to encapsulate all continuous sequence-to-sequence functions, affirming its Turing completeness.
As shown in Figure~\ref{fig:sa}(b), Longformer~\cite{longformer} additionally introduces the dilated sliding window pattern. It is analogous to dilated CNNs and makes the sliding window “dilated” to increase the receptive field.
To adapt the model to the sparse setting, Structured Sparse Attention~\cite{structuredSparse} advocates an entropy-aware training method that congregates
high-probability attention values into denser regions.
Unlike previous studies that manually design sparse patterns, SemSA~\cite{anonymous2023semsa} uses gradient-based profiling to identify important attention patterns and automatically optimizes the attention density distribution to further improve model efficiency.

In contrast, Dynamic sparse attention adaptively eliminates activation values based on varying inputs, employing real-time monitoring of neuronal activation values to bypass computations for neurons with negligible impact, thereby achieving pruning. 
Most dynamic sparse attention methods employ the dynamic token-pruning methods, as Figure~\ref{fig:sa}(c) shows.
Spatten~\cite{spatten}, SeqBoat~\cite{seqBoat} and Adaptively Sparse Attention~\cite{adaptiveSparse} 
leverage the inherent redundancy in linguistic constructs to propose dynamic token-level pruning strategies. 
Spatten~\cite{spatten} assesses the cumulative importance of each word by aggregating the attention matrix columns, subsequently pruning tokens with minimal cumulative significance from the input in subsequent layers. 
SeqBoat~\cite{seqBoat} trains a linear State Space Model (SSM) with a sparse sigmoid function to determine which token to prune for each attention head. 
Both Spatten and SeqBoat prune the uninformative tokens for the whole input.
Adaptively Sparse Attention~\cite{adaptiveSparse} gradually prunes the tokens during the generation process. It drops parts of the context that are no longer required for future generation.

In addition to dynamic token pruning, dynamic attention pruning strategies are also employed~\cite{zhang2024h2o, pagliardini2023faster, roy2021efficient, tay2020sparse, kitaev2020reformer}. As Figure~\ref{fig:sa}(d) shows, instead of pruning all the attention values of certain tokens, these methods dynamically prune the selective part of the attention based on the input.
A prominent approach within this domain is dynamically segmenting input tokens into groups, known as buckets, and strategically omitting the attention calculations for tokens that reside in separate buckets. The challenge and focus of these methods lie in the way to cluster related tokens together, thereby facilitating attention computations solely among them to enhance efficiency.
Reformer~\cite{kitaev2020reformer} leverages locality-sensitive hashing to cluster keys and queries that share identical hash codes into the same bucket.
Following this, Sparse Flash Attention~\cite{pagliardini2023faster} introduces specialized GPU kernels optimized for this hash-based sparse attention mechanism, further improving computational efficiency.
Meanwhile, the Routing Transformer \cite{roy2021efficient} employs a spherical k-means clustering algorithm to aggregate tokens into buckets, optimizing the selection process for attention computations.
Sparse Sinkhorn Attention~\cite{tay2020sparse} adopts a learned sorting network to align keys with their relevant query buckets, ensuring that attention is computed only between the corresponding query-key pairs.
Diverging from the bucket-level operation, H$_2$O~\cite{zhang2024h2o} introduces the token-level dynamic attention pruning mechanism. It combines static local attention with dynamic computations between the current query and a set of dynamically identified key tokens, termed heavy-hitters (H$_2$). These heavy-hitters are dynamically adjusted with an eviction policy aimed at removing the least significant keys at each generation step, effectively managing the size and relevance of the heavy-hitter set.

Moreover, viewing each token as a graph node and attention between tokens as edges offers an extended perspective on static sparse attention~\cite{bigbird, diffuser}. The original, full attention mechanism equates to a complete graph with a uniform shortest path distance of 1. Sparse attention, with its random mask, introduces random edges, effectively reducing the shortest path distance between any two nodes to \(O(\log n)\), thus maintaining efficient information flow akin to full attention.
Diffuser~\cite{diffuser} utilizes the perspective of graph theory to expand the receptive field of sparse attention with multi-hop token correlations. It also takes inspiration from the expander graph properties to design better sparse patterns that approximate the information flow of full attention.

Beyond the attention-level and token-level sparsity, the scope of attention pruning extends to various granularities. 
Spatten~\cite{spatten} also extends pruning beyond token granularity to attention head granularity, eliminating computations for inessential attention heads to further reduce computational and memory demands. %DynamicBERT~\cite{dynabert} implements an early stopping mechanism at the layer level, enabling the immediate return of results for simpler tasks and bypassing subsequent computations.

\subsubsection{Structure Optimization}

The objective of structure optimization is to refine model architecture or structure with the goal of enhancing the balance between model efficiency and performance. Within this field of research, two prominent techniques stand out: Neural Architecture Search (NAS) and Low Rank Factorization (LRF). 

\noindent \textbf{Neural Architecture Search.} 
Neural Architecture Search (NAS)~\cite{zoph2016neural} aims to automatically search the optimal neural architectures that strike an optimized balance between efficiency and performance. 
AutoTinyBERT~\cite{yin2021autotinybert} utilizes one-shot Neural Architecture Search (NAS) to discover the hyper-parameters of the Transformer architecture. Notably, it introduces a compelling batch-wise training approach to train a Super Pre-trained Language Model (SuperPLM) and subsequently employs an evolutionary algorithm to identify the optimal sub-models. 
NAS-BERT~\cite{xu2021bert} trains a large super-net on conventional self-supervised pre-training tasks using several innovative techniques, such as block-wise search, search space pruning, and performance approximation. This approach allows NAS-BERT to be applied efficiently across various downstream tasks without requiring extensive re-training. 
Structure pruning via NAS~\cite{klein2023structural} treats structural pruning as a multi-objective NAS problem, and solves it via the one-shot NAS method. 
LiteTransformerSearch~\cite{javaheripi2022litetransformersearch} proposes to use a training-free indicator, i.e., the number of parameters, as a proxy indicator to guide the search. This method enables efficient exploration and selection of the optimal architectures without the need for actual training during the search phase. 
AutoDistil~\cite{xu2022few} presents a fully task-agnostic few-shot NAS algorithm featuring three primary techniques: search space partitioning, task-agnostic SuperLM training, and task-agnostic search. This approach aims to facilitate efficient architecture discovery across various tasks with minimal task-specific adaptations. 
Typically, NAS algorithms necessitate evaluating the performance of each sampled architecture, which can incur significant training cost. Consequently, these techniques are challenging to apply to LLMs.

\noindent \textbf{Low Rank Factorization.} Low Rank Factorization (LRF), or Low Rank Decomposition, aims to approximate a matrix $A^{m\times n}$ with two low-rank matrices $B^{m\times r}$ and $C^{r\times n}$ by: 
\begin{equation}
    A^{m\times n} \approx B^{m\times r}\times C^{r\times n}, 
\end{equation}
where $r$ is much smaller than $m$ and $n$. In this way, LRF can diminish memory usage and enhance computational efficiency. Furthermore, during the decoding stage of LLM inference, memory access cost presents a bottleneck to the decoding speed. Therefore, LRF can reduce the number of parameters that need to be loaded, thereby accelerating the decoding speed. 
LoRD~\cite{kaushal2023lord} shows the potential of compressing the LLMs without largely degrading the performance via LRF. Specifically, it adopts Singular Value Decomposition (SVD) to factorize the weight matrices, and successfully compresses a LLM with 16B parameters to 12.3B with minimal performance drop. 
TensorGPT~\cite{xu2023tensorgpt} introduces a method to compress the embedding layer using Tensor-Train Decomposition. Each token embedding is treated as a Matrix Product State (MPS) and efficiently computed in a distributed manner. 
LoSparse~\cite{li2023losparse} combines the benefits of LRF and weight pruning for LLM compression. By leveraging low-rank approximation, LoSparse mitigates the risk of losing too many expressive neurons that typically occurs with direct model pruning. 
LPLR~\cite{saha2023matrix} and ZeroQuant-V2~\cite{yao2023zeroquant} both propose to compress the weight matrix by simultaneously applying LRF and quantization to it. 
DSFormer~\cite{chand2023dsformer} proposes to factorize the weight matrix into the product of a semi-structured sparse matrix and a small dense matrix. 
ASVD~\cite{yuan2023asvd} designs an activation-aware SVD method. This approach involves scaling the weight matrix based on activation distribution prior to applying SVD for matrix decomposition. ASVD also involves determining an appropriate truncation rank for each layer through a search process. 
SVD-LLM~\cite{wang2024svd} analyses the relationship between the singular values of the transformed weight matrices and the compression loss. Then, it designs a truncation-aware data whitening technique to identify the singular value that causes the minimal loss after removing it. Additionally, SVD-LLM develops a layer-wise closed-form update strategy to recover the task performance after the factorization.

\subsubsection{Knowledge Distillation}

\begin{figure}[h]
    \centering
    \includegraphics[width=0.90\linewidth]{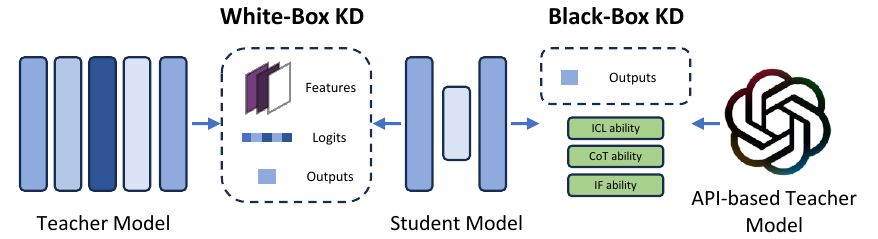}
    \caption{Illustration of White-Box KD (left) and Black-Box KD (right).}
    \label{fig:kd}
\end{figure}

Knowledge Distillation (KD) is a well-established technique for model compression, wherein knowledge from large models (referred to as teacher models) is transferred to smaller models (referred to as student models). In the context of LLMs, KD involves using the original LLMs as teacher models to distill smaller LMs. Numerous studies have focused on effectively transferring various abilities of LLMs to smaller models. In this domain, methods can be categorized into two main types: white-box KD and black-box KD (as illustrated in Fig.~\ref{fig:kd}).

%\todo{改一下图的右侧}

\noindent \textbf{White-box KD.} White-box KD refers to distillation methods that leverage access to the structure and parameters of the teacher models. This approach enables KD to effectively utilize the intermediate features and output logits of the teacher models for enhanced performance of the student models. 
MiniLLM~\cite{gu2023knowledge} proposes to adopt the standard white-box KD approach but replace the forward Kullback-Leibler divergence (KLD) with the reverse KLD. 
GKD~\cite{agarwal2023gkd} introduces the use of on-policy data, which includes output sequences generated by the student model itself, to further distill the student model. This method focuses on aligning the output logits between the teacher and student models using these on-policy data. 
TED~\cite{liang2023less} presents a task-aware layer-wise KD method. This approach involves adding filters after each layer in both the teacher and student models, training these task-specific filters, and subsequently freezing the teacher model's filters while training the student filters to align their output features with the corresponding teacher filters. 
MiniMoE~\cite{zhang2023lifting} mitigates the capacity gap by utilizing a Mixture-of-Experts (MoE) model as the student model. 
DynaBERT~\cite{hou2020dynabert} proposes to progressively decrease the models' width and depth, and uses knowledge distillation to train the smaller models. 
For newly emerging entities, pre-trained language models (LLMs) may lack up-to-date information. To address this, one solution involves incorporating additional retrieved texts into prompts, albeit at an increased inference cost. Alternatively, KPTD~\cite{padmanabhan2023propagating} suggests transferring knowledge from entity definitions into LLM parameters via knowledge distillation. This method generates a transfer set based on entity definitions and distills the student model to match output distributions with the teacher model based on these definitions. 

\noindent \textbf{Black-box KD.} Black-box KD refers to the knowledge distillation methods in which the structure and parameters of teacher models are not available. Typically, black-box KD only uses the final results obtained by the teacher models to distill the student models. In the field of LLMs, black-box KD mainly guides the student models to learn LLMs' generalization ability and emergent ability, including In-Context Learning (ICL) ability~\cite{dong2022survey}, Chain-of-Thought (CoT) reasoning ability~\cite{wei2022chain} and Instruction Following (IF) ability~\cite{ouyang2022training}. %\todo{rephrase}

Regarding the ICL ability, Multitask-ICT~\cite{huang2022context} introduces in-context learning distillation to transfer the multitask few-shot ability of Large Language Models (LLMs), leveraging both in-context learning and language modeling proficiency. 
MCKD~\cite{zhao2023multistage} observes that student models distilled from in-context learned teacher models often exhibit superior performance on unseen input prompts. Building on this observation, MCKD devises a multi-stage distillation paradigm where the student model from previous stages is employed to generate distillation data for subsequent stages, enhancing the effectiveness of the distillation method. 

To distill the Chain-of-Thought (CoT) reasoning ability, several techniques such as Distilling Step-by-Step~\cite{hsieh2023distilling}, SCoTD~\cite{li2023symbolic}, CoT Prompting~\cite{magister2022teaching}, MCC-KD~\cite{chen2023mcc}, and Fine-tune-CoT~\cite{ho2022large} propose distillation methods that incorporate responses and rationales extracted from LLMs to train student models. 
Socratic CoT~\cite{shridhar2023distilling} also targets reasoning ability transfer to smaller models. Specifically, it fine-tunes a pair of student models, namely a Question Generation (QG) model and a Question Answering (QA) model. The QG model is trained to generate intermediate questions based on input questions, guiding the QA model in producing the final response. 
PaD~\cite{zhu2023pad} observes that faulty reasoning (i.e., correct final answer but incorrect reasoning steps) can be detrimental to student models. To address this, PaD proposes generating synthetic programs for reasoning problems, which can then be automatically checked by an additional interpreter. This approach helps in removing distillation data with faulty reasoning, enhancing the quality of the training data for student models. 

For the IF ability, several methods have been proposed to transfer this capability to smaller models. DISCO~\cite{chen2023disco} introduces a technique where phrasal perturbations are generated using a LLM. These perturbations are then filtered by a task-specific teacher model to distill high-quality counterfactual data. 
LaMini-LM~\cite{wu2023lamini} aims to transfer instruction following ability by designing a diverse instruction set for distilling student models. 
Lion~\cite{jiang2023lion} utilizes the teacher model to identify difficult instructions, and generates new and complex instructions to distill the small model. 

\subsubsection{Dynamic Inference}
\label{sec:dynamic_inference}

\begin{figure}[h]
    \centering
    \includegraphics[width=0.750\linewidth]{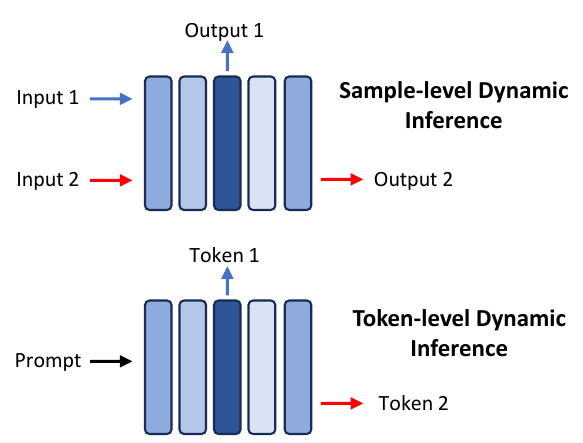}
    \caption{Illustration of Token-level (up) and Sample-level (down) dynamic inference.}
    \label{fig:dynamic}
\end{figure}

Dynamic inference involves the adaptive selection of model sub-structures during the inference process, conditioned on input data. This section focuses on early exiting techniques, which enable a LLM to halt its inference at different model layers depending on specific samples or tokens. Notably, while MoE techniques (discussed in Sec.~\ref{sec:moe}) also adjust model structure during inference, they typically involve expensive pre-training cost. In contrast, early exiting techniques only require training a small module to determine when to conclude the inference. Some previous surveys~\cite{han2021dynamic,xu2023survey} have reviewed dynamic inference techniques for traditional language models (e.g., RNN, LSTM). In this paper, we categorize studies on early exiting techniques for LLMs into two main types: sample-level early exiting and token-level early exiting (illustrated in Fig.~\ref{fig:dynamic}). 

\noindent \textbf{Sample-level.} Sample-level early exiting techniques focus on determining the optimal size and structure of Language Models (LLMs) for individual input samples. A common approach is to augment LLMs with additional modules after each layer, leveraging these modules to decide whether to terminate inference at a specific layer. FastBERT~\cite{liu2020fastbert}, DeeBERT~\cite{xin2020deebert}, MP~\cite{he2023magic}, and MPEE~\cite{kong2022accelerating} train these modules directly to make decisions (e.g., outputting 0 to continue or 1 to stop) based on features from the current layer. 
Global Past-Future Early Exit~\cite{liao2021global} proposes a method that enriches the input to these modules with linguistic information from both preceding and subsequent layers. Given that future layer features are not directly accessible during inference, a simple feed-forward layer is trained to estimate these future features. 
PABEE~\cite{zhou2020bert} trains the modules as output heads for direct prediction, suggesting inference termination when predictions remain consistent. 
HASHEE~\cite{sun2022simple} employs a non-parametric decision-making approach based on the hypothesis that similar samples should exit inference at the same layer.

\noindent \textbf{Token-level.} In the decoding stage of LLM inference, where tokens are generated sequentially, token-level early exiting techniques aim to optimize the size and structure of LLMs for each output token. 
CALM~\cite{schuster2022confident} introduces early exit classifiers after each Transformer layer, training them to output confidence scores that determine whether to halt inference at a specific layer. Notably, in the self-attention block, computing the current token's feature at each layer relies on all previous tokens' features (i.e., KV cache) in the same layer. To address the issue of missing KV cache due to early exiting of previous tokens, CALM proposes directly copying the feature from the exiting layer to subsequent layers, with experimental results showing only minor performance degradation. 
SkipDecode~\cite{del2023skipdecode} addresses limitations of previous early exiting methods that hinder their applicability to batch inference and KV caching, thereby limiting actual speed-up gains. For batch inference, SkipDecode proposes a unified exit point for all tokens within a batch. Regarding KV caching, SkipDecode ensures a monotonic decrease in exit points to prevent recomputation of KV cache, facilitating efficiency gains during inference.

\subsection{Knowledge, Suggestions and Future Direction}
\label{sec:model_summary}

In the field of efficient structure design, the pursuit of alternative architectures to Transformers is a burgeoning area of research. Examples such as Mamba~\cite{gu2023mamba}, RWKV~\cite{peng2023rwkv}, and their respective variants~\cite{zhu2024vision,he2024densemamba} have demonstrated competitive performance across various tasks, garnering increasing attention in recent times. 
Nevertheless, it remains pertinent to investigate whether these non-Transformer models may exhibit certain shortcomings compared to Transformer models. Concurrently, exploring the integration of non-Transformer architectures with the attention operation~\cite{togertherai2023stripedhyena,park2024can,ai212024jamba} represents another promising avenue for future research.

In the realm of model compression, quantization stands out as the predominant method employed in Large Language Model (LLM) deployment, primarily due to two key factors. Firstly, quantization presents a convenient means of compressing LLMs. For instance, employing Post-Training Quantization (PTQ) methods can reduce the parameter count of an LLM with seven billion parameters to a compressed form within a matter of minutes. Secondly, quantization holds the potential to achieve substantial reductions in memory consumption and inference speed, while introducing only minor performance trade-offs. This compromise is generally deemed acceptable for numerous real-world applications. However, it's worth noting that quantization may still compromise certain emergent abilities of LLMs, such as self-calibration or multi-step reasoning. Additionally, in specific scenarios like dealing with long contexts, quantization could lead to significant performance degradation~\cite{li2024evaluating}. Consequently, it is required to carefully select appropriate quantization methods to mitigate the risk of such degradation in these specialized cases.

Extensive literature has devoted into studying sparse attention techniques for efficient long-context processing. For example, a recent representative work, StreamingLLM~\cite{xiao2023efficient}, can process 4 million tokens by only restoring several attention sink tokens. Nonetheless, these approaches often sacrifice critical information, resulting in performance degradation. Therefore, the challenge of preserving essential information while efficiently managing long contexts remains an important area for future exploration. As for the weight pruning techniques, LLM-KICK~\cite{jaiswal2023compressing} notes that current state-of-the-art (SOTA) methods experience considerable performance degradation even at relatively low sparsity ratios. Consequently, developing effective weight pruning methods to maintain LLM performance remains an emerging and critical research direction. 

The optimization of model structures often involves the use of Neural Architecture Search (NAS), which typically demands extensive computational resources, posing a potential barrier to its practical application in compressing LLMs. Therefore, investigating the feasibility of employing automatic structure optimization for LLM compression warrants further exploration. Additionally, the challenge remains for techniques like low-rank factorization (LRF) to achieve an optimal balance between compression ratio and task performance. For instance, ASVD~\cite{yuan2023asvd} achieves only a modest 10\% to 20\% compression ratio without compromising the reasoning capabilities of LLMs.

In addition to employing individual model compression techniques, several studies explore the combination of different methods to compress LLMs, leveraging their respective advantages for improved efficiency. For instance, MPOE~\cite{gao2022parameter} applies weight matrix factorization specifically to the expert Feed-Forward Networks (FFNs) in MoE-based LLMs, with the goal of further reducing memory requirements. LLM-MQ~\cite{li2023llm} utilizes weight sparsity techniques to protect weight outliers during model quantization, thereby minimizing quantization errors. LPLR~\cite{saha2023matrix} focuses on quantizing low-rank factorized weight matrices to further decrease memory footprint and memory access cost during LLM inference. Furthermore, LoSparse~\cite{li2023losparse} combines low-rank factorization with weight pruning, leveraging pruning to enhance the diversity of low-rank approximation while using low-rank factorization to retain important weights and prevent loss of critical information. These approaches highlight the potential of integrating multiple compression techniques to achieve better optimization of LLMs.

\section{System-level Optimization}
\label{sec:system-level-opt}
% [03.31] hongke: 还差attention operator和graph-level optimization没改
% first we should discuss the difference between inference engines and serving systems. @hongke
% [done] \todo{分类有点怪，inference engine里的技巧也会在serving system。描述下场景，inference engine统一描述，online场景有新的需求：动态query、多用户，serving专门针对这种场景有新的优化技巧 @hk}
% We discuss the LLM system designs in two scenarios: off-line inference and on-line serving. Off-line inference involves only one user, and all the requests are launched at the beginning. The goal of off-line inference is to reduce the inference latency. Therefore, the system-level designs for inference engines are dedicated to enhance the model forward process. Unlike off-line inference, on-line LLM serving receives requests from multiple users, and hence faces the challenge of handling asynchronous requests. In serving systems, the memory management is optimized to hold more requests, and efficient batching and scheduling strategies are integrated to improve the system throughput. 

The system-level optimization for LLM inference primarily involves enhancing the model forward pass. Considering the computational graph of an LLM, there exist multiple operators, with attention and linear operators dominating most of the runtime. As mentioned in Sec.~\ref{sec:efficiency}, system-level optimization primarily considers the distinctive characteristics of the attention operator and the decoding approach within LLM. In particular, to address the specific issues related to the decoding approach of LLMs, the linear operator requires special tiling designs, and speculative decoding methods are proposed to improve the utilization. The substantial memory demand of LLMs leads to the offloading of parameters or KV cache to the CPU. Furthermore, in the context of online serving, requests come from multiple users. Therefore, beyond the optimizations discussed earlier, online serving faces challenges related to memory, batching, and scheduling arising from asynchronous requests. %\revise{[add offloading.]}

\subsection{Inference Engine}
\label{sec:inference_engine}

The optimizations for inference engines are dedicated to accelerating the model forward process. The main operators and the computational graph in LLM inference are highly optimized. Besides, the speculative decoding technique is proposed to accelerate the inference speed without performance degradation, and the offloading technique is introduced to mitigate the memory pressure. %\revise{[add offloading.]}

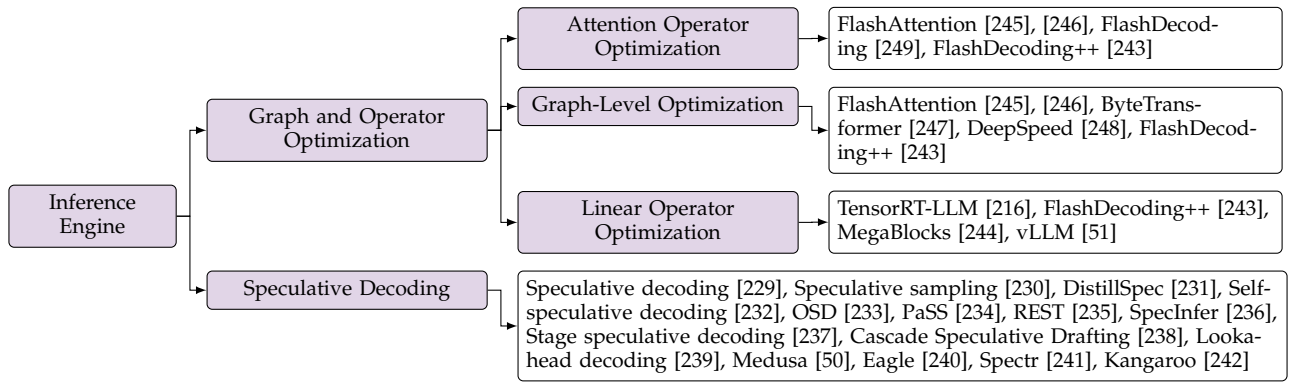
\begin{figure*}[h]
\centering
\tikzset{
    basic/.style  = {draw, text width=2cm, align=center, font=\sffamily, rectangle},
    root/.style   = {basic, rounded corners=2pt, thin, align=center, fill=white,text width=8cm, rotate=90, font=\footnotesize},
    dnode/.style = {basic, thin, rounded corners=2pt, align=center, fill=yellow!30,text width=3.5cm, font=\footnotesize},
    dnode_1/.style = {basic, thin, rounded corners=2pt, align=center, fill=yellow!30,text width=2cm, font=\footnotesize},
    mnode/.style = {basic, thin, rounded corners=2pt, align=center, fill=blue!10,text width=3.5cm, font=\footnotesize},
    mnode_1/.style = {basic, thin, rounded corners=2pt, align=center, fill=blue!10,text width=2cm, font=\footnotesize}, 
    snode/.style = {basic, thin, rounded corners=2pt, align=center, fill=npurple,text width=3.5cm, font=\footnotesize},
    snode_1/.style = {basic, thin, rounded corners=2pt, align=center, fill=npurple,text width=2cm, font=\footnotesize},
    tnode/.style = {basic, thin, align=left, fill=pink!60, text width=15em, align=center},
    xnode/.style = {basic, thin, rounded corners=2pt, align=center, fill=blue!20,text width=5cm,},
    wnode/.style = {basic, thin, rounded corners=2pt, align=left, fill=white,text width=5.8cm, font=\footnotesize},
    wnode_1/.style = {basic, thin, rounded corners=2pt, align=left, fill=white,text width=3.3cm, font=\footnotesize},
    wnode_2/.style = {basic, thin, rounded corners=2pt, align=left, fill=white,text width=10cm, font=\footnotesize},
    %edge from parent/.style = {draw=black, edge from parent fork right}
    %edge from parent/.style = {draw=black, edge from parent fork down}
}
\begin{forest} 
for tree={
    grow=east,
    growth parent anchor=east,
    parent anchor=east,
    child anchor=west,
    edge path={\noexpand\path[\forestoption{edge},->, >={latex}] 
         (!u.parent anchor) -- +(5pt,0pt) |- (.child anchor)
         \forestoption{edge label};}
}
% l sep is used for arrow distance
[Inference Engine, snode_1
    %[Model Parallelism, snode]
    [Speculative Decoding, snode
        [{Speculative decoding~\cite{leviathan2023fast}, Speculative sampling~\cite{chen2023accelerating}, DistillSpec~\cite{zhou2023distillspec}, Self-speculative decoding~\cite{zhang2023draft}, OSD~\cite{liu2023online}, PaSS~\cite{monea2023pass}, REST~\cite{he2023rest}, SpecInfer~\cite{miao2023specinfer}, Stage speculative decoding~\cite{spector2023accelerating}, Cascade Speculative Drafting~\cite{chen2023cascade}, Lookahead decoding~\cite{fu2023lookahead}, Medusa~\cite{cai2024medusa}, Eagle~\cite{li2023eagle}, Spectr~\cite{sun2023spectr}, Kangaroo~\cite{liu2024kangaroo}}, wnode_2]
    ]
    [Offloading, snode
        [{FlexGen~\cite{sheng2023flexgen}, llama.cpp~\cite{ggerganov2024llamacpp}, powerinfer~\cite{song2023powerinfer}, FastDecode~\cite{he2024fastdecode}}, wnode_2]
    ]
    [Graph and Operator Optimization, snode
        [Linear Operator Optimization, snode
            [{TensorRT-LLM~\cite{tensorrt-llm},
            FlashDecoding++~\cite{hong2024flashdecoding},
            MegaBlocks~\cite{gale2023megablocks},
            vLLM~\cite{kwon2023vllm}}, wnode]
        ]
        [Graph-Level Optimization, snode
            [{FlashAttention~\cite{flashattention,flashattention2}, 
            ByteTransformer~\cite{bytetransformer}, 
            DeepSpeed~\cite{deepspeed}, 
            FlashDecoding++~\cite{hong2024flashdecoding}}, wnode]
        ]
        [Attention Operator Optimization, snode
            [{FlashAttention~\cite{flashattention,flashattention2}, FlashDecoding~\cite{flashdecoding}, FlashDecoding++~\cite{hong2024flashdecoding}}, wnode]
        ]
    ]
]
\end{forest}

\caption{Taxonomy of the optimization for LLM inference engine.}
\label{fig:engine_framework}
\end{figure*}

\subsubsection{Graph and Operator Optimization} 
\label{sec:graph_operator_optimize}

\begin{figure*}[htb]
    \centering
    \includegraphics[width=0.95\linewidth]{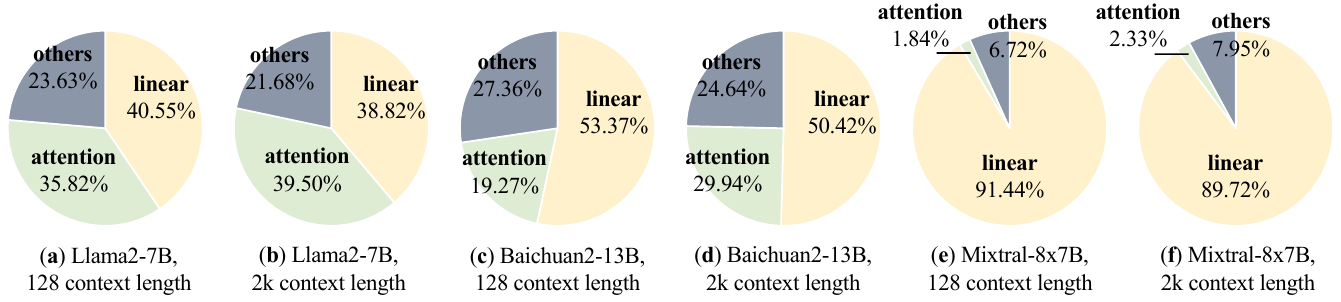}
    \caption{Inference runtime breakdown over multiple LLMs.}
    \label{fig:runtime}
\end{figure*}

\noindent \textbf{Runtime Profiling.} Using HuggingFace~\cite{huggingface2024transformers} implementation, we profile the inference runtime with different models and context lengths. The profiling results in Fig.~\ref{fig:runtime} demonstrate that attention operators and linear operators collectively dominate runtime, with their combined duration often exceeding 75\% of the inference duration. 
Consequently, a significant portion of optimization efforts at the operator level is dedicated to enhancing the performance of the two operators. Furthermore, there are multiple operators occupying a small proportion of runtime, which fragments the operator execution timeline and increases the cost of kernel launch on the CPU side. To address this issue, at the computational graph level, current optimized inference engines implement highly fused operators. 

\noindent \textbf{Attention Operator Optimization.} 
The standard attention computation (e.g., using Pytorch) involves the multiplication of the Query matrix ($\textbf{Q}$) with the Key matrix ($\textbf{K}$), resulting in quadratic time and space complexity in relation to the input sequence length. As shown in Fig.~\ref{fig:runtime}, the time proportion of the attention operator increases as the context length grows. This translates to high demands on memory size and computational capability, especially when dealing with long sequences. To address the computational and memory overhead of standard attention computation on GPUs, customized attention operators are essential. FlashAttention~\cite{flashattention, flashattention2} fuses the entire attention operation into a single, memory-efficient operator to alleviate memory access overhead. The input matrices (Q, K, V) and attention matrix are tiled into multiple blocks, which eliminates the need for complete data loading. Built upon Flash Attention, FlashDecoding~\cite{flashdecoding} aims to maximize computational parallelism for decoding. Due to the application of the decoding approach, the Q matrix degrades into a batch of vectors during decoding, which makes it challenging to fill the computational units if the parallelism is limited to the batch size dimension. FlashDecoding addresses this by introducing parallel computation along the sequence dimension. While this introduces some synchronization overhead to softmax computation, it leads to noticeable improvements in parallelism, particularly for small batch sizes and long sequences. The subsequent work, FlashDecoding++~\cite{hong2024flashdecoding}, observes that in previous works~\cite{flashattention, flashattention2, flashdecoding}, the maximum value within the softmax only serves as a scaling factor to prevent data overflow. However, the dynamical maximum value incurs significant synchronization overhead. Moreover, extensive experiments indicate that in typical LLM (e.g., Llama2~\cite{touvron2023llama2}, ChatGLM~\cite{chatglm}), over 99.99\% of the softmax inputs fall within a certain range. Thus, FlashDecoding++ proposes to determine the scaling factor based on statistics in advance. This eliminates the synchronization overhead in softmax computation, enabling parallel execution of subsequent operations alongside the softmax computation.
%\todo{figure for FA, FD and FD++}

\noindent \textbf{Linear Operator Optimization} 
The linear operator plays a pivotal role in LLM inference, performing in feature projection and Feedforward Neural Networks (FFNs). In traditional neural networks, linear operators can be abstracted into General Matrix-Matrix Multiplication (GEMM) operations. However, in the case of LLM, the application of the decoding approach results in a notably reduced dimension, diverging from the conventional GEMM workload. The low-level implementation of traditional GEMM has been highly optimized, and mainstream LLM frameworks (\textit{e.g.,} DeepSpeed~\cite{deepspeed}, vLLM~\cite{kwon2023vllm}, OpenPPL~\cite{OpenPPL} and etc.) primarily call the GEMM APIs offered by cuBLAS~\cite{cuBLAS} for linear operators. Without an explicitly tailored implementation for GEMMs with a reduced dimension, the linear operators during decoding suffer inefficiency. A notable trend to address the issue is observed in the latest release of TensorRT-LLM~\cite{tensorrt-llm}. It introduces a dedicated General Matrix-Vector Multiplication (GEMV) implementation, potentially improving efficiency for the decoding step. Recent research FlashDecoding++~\cite{hong2024flashdecoding} makes a further step, addressing the inefficiency of cuBLAS~\cite{cuBLAS} and CUTLASS~\cite{CUTLASS} libraries when dealing with small batch sizes during the decode step. The authors first introduce the concept of the FlatGEMM operation to represent the workload of GEMM with a highly reduced dimension (dimension size $<8$ in FlashDecoding++). As FlatGEMM poses new computational characteristics, the tiling strategy for traditional GEMMs necessitates modification to be applied. The authors observe that two challenges exist as the workload varies: low parallelism and memory access bottleneck. To tackle the challenges, FlashDecoding++ adopts a fine-grained tiling strategy to improve parallelism, and leverages the double buffering technique to hide memory access latency. Furthermore, recognizing that the linear operations in typical LLM (e.g., Llama2~\cite{touvron2023llama2}, ChatGLM~\cite{chatglm}) often have fixed shapes, FlashDecoding++ establishes a heuristic selection mechanism. This mechanism dynamically chooses between different linear operators based on the input size. The options include FastGEMV~\cite{fastgemv}, FlatGEMM, and GEMM provided by cuBLAS~\cite{cuBLAS,CUTLASS} libraries. This approach ensures the selection of the most efficient operator for the given linear workload, potentially leading to better end-to-end performance.

Recently, the application of the MoE FFN to enhance the model capability has become a trend in LLMs~\cite{jiang2024mixtral}. This model structure also puts forward new requirements for operator optimization. As shown in Fig.~\ref{fig:runtime}, in the Mixtral model with MoE FFN, the linear operator dominates the runtime due to the non-optimized FFN computation in the HuggingFace implementation. Besides, Mixtral's adoption of the GQA attention structure decreases the attention operator's runtime proportion, which further points out the urgent need to optimize the FFN layer. MegaBlocks~\cite{gale2023megablocks} is the first to optimize the computation for MoE FFN layers. The work formulates the MoE FFN computation into block-sparse operations and proposes tailored GPU kernels for acceleration. However, MegaBlocks concentrates on the efficient training of the MoE models and hence ignores the characteristics of inference (e.g., the decoding approach). Existing frameworks are working hard to optimize the computations of the MoE FFN inference stage. The official repository of vLLM~\cite{kwon2023vllm} integrates the fused kernels for MoE FFN in Triton~\cite{tillet2019triton}, seamlessly removing the index overhead. 

%[done] \todo{描述下切片策略；修改下整段文字 @hk}

\noindent \textbf{Graph-Level Optimization.} 
Kernel fusion stands out as a prevalent graph-level optimization because of its capability to reduce runtime. There are three main advantages of applying kernel fusion: (1) To reduce memory access. The fused kernel inherently removes the memory access of intermediate results, mitigating the memory bottleneck for operators. (2) To mitigate kernel launching overhead. For some lightweight operators (e.g., residual adding), the kernel launching time occupies most of the latency, and kernel fusion reduces individual kernel launchings. (3) To enhance parallelism. For those operators without data dependency, when one-by-one kernel execution fails to fill the hardware capacity, it is beneficial to parallel the kernels via fusion. 

The technique of kernel fusion proves effective with LLM inference, with all of the aforementioned benefits. FlashAttention~\cite{flashattention} formulates the attention operator into one single kernel, removing the overhead of accessing the attention results. Based on the fact that the attention operator is memory-bounded, the reduction of memory access effectively transfers to runtime speed-up. ByteTransformer~\cite{bytetransformer} and DeepSpeed~\cite{deepspeed} propose to fuse lightweight operators including residual adding, layernorm, and activation functions, into the former linear operators to reduce the kernel launching overhead. As a result, those lightweight operators disappear in the timeline with nearly no extra latency. Moreover, kernel fusion is also adopted to enhance the utilization of LLM inference. The projections of Query, Key, and Value matrices are originally three individual linear operations, and are fused into one linear operator to deploy on modern GPUs. Currently, the kernel fusion technique has been exploited in LLM inference practice, and highly optimized inference engines employ only a few fused kernels within the runtime. For example, in FlashDecoding++~\cite{hong2024flashdecoding} implementation, a transformer block integrates merely seven fused kernels. Leveraging the aforementioned operators and kernel fusion optimization, FlashDecoding++ achieves up to 4.86$\times$ speed-up over the HuggingFace implementation. 

\begin{table*}[htb]
\caption{Comparison of several open-source implementations of speculative decoding. In this table, we also show the additional overhead of constructing draft models. Note that for SpD~\cite{leviathan2023fast,chen2023accelerating}, LADE~\cite{fu2023lookahead}, Medusa~\cite{cai2024medusa} and Eagle~\cite{li2023eagle}, we report the training cost from their original papers. And for SSD~\cite{zhang2023draft} and REST~\cite{lewis2020retrieval}, we run the sub-LLM search and datastore construction with the code they provide, and report the time cost. Besides, for Medusa, we use Medusa-1~\cite{cai2024medusa} which does not fine-tune the original LLM backbone.}
\label{tab:spec_comparison}
\begin{center}
\resizebox{0.98\textwidth}{!}
{
\begin{tabular}{ccccccc}
\toprule
Method & Draft Model & Draft Construction & Draft Verifier & \makecell[c]{Additional Overhead \\ (GPU hours)} & Acceptance Rate & Speed-up
\\ 
\midrule
SpD~\cite{leviathan2023fast,chen2023accelerating} & small speculative model & one draft sequence & speculative sampling & 275 & 1.77$\sim$2.02$\times$ & 1.05$\sim$1.77$\times$ \\
LADE~\cite{fu2023lookahead} & LLM + N\_grams & one draft sequence & greedy sampling & 0 & 1.92$\sim$2.14$\times$ & 1.12$\sim$1.30$\times$ \\
SSD~\cite{zhang2023draft} & sub-LLM & one draft sequence & speculative sampling & 4  &1.64$\sim$1.74$\times$ & 1.01$\sim$1.23$\times$ \\
REST~\cite{lewis2020retrieval} & datastore & token tree & speculative sampling & 1.5 & 2.18$\sim$2.31$\times$ & 1.72$\sim$2.27$\times$ \\
Medusa-1~\cite{cai2024medusa} & four LLM heads & token tree & speculative sampling & $\sim$24 & 2.52$\sim$2.62$\times$ & 2.04$\sim$2.86$\times$ \\
Eagle~\cite{li2023eagle} & one Transformer Layer & token tree & speculative sampling & 96$\sim$192 & 3.47$\sim$3.72$\times$ & 2.77$\sim$3.74$\times$ \\
\bottomrule
\end{tabular}
}
\end{center}
\end{table*}

\subsubsection{Speculative Decoding}

\begin{figure}[h]
    \centering
    \includegraphics[width=0.98\linewidth]{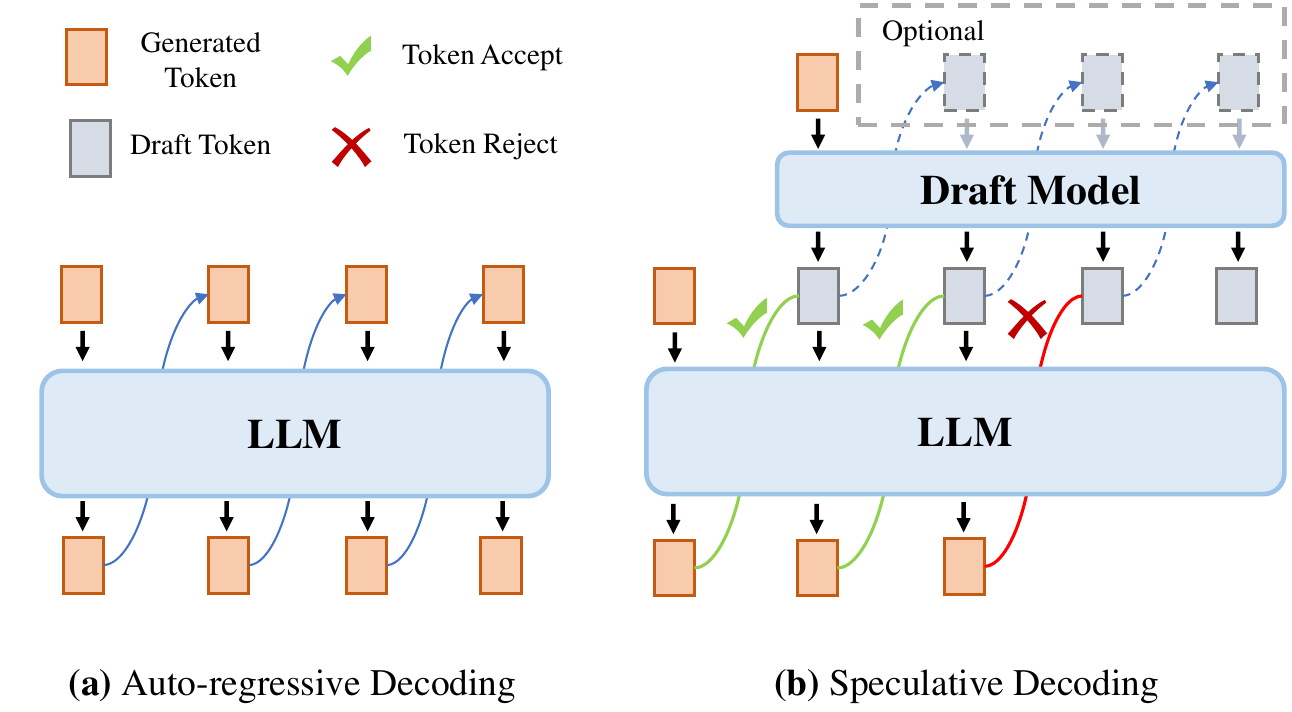}
    \caption{Comparison of auto-regressive decoding (a) and speculative decoding (b).}
    \label{fig:spec_framework}
\end{figure}

Speculative decoding~\cite{zhang2024beyond,xia2024unlocking} is an innovative decoding technique for auto-regressive LLMs designed to enhance decoding efficiency without compromising the fidelity of outputs. The core idea of this approach involves employing a smaller model, termed a draft model, to predict several subsequent tokens efficiently, followed by validation of these predictions using the target LLM in parallel. This methodology aims to enable the LLM to generate multiple tokens within the time frame typically required for a single inference. 
Fig.~\ref{fig:spec_framework} demonstrates the comparison of the traditional auto-regressive decoding method and the speculative decoding approach. 
Formally, speculative decoding approach consists of two steps: 
\begin{enumerate}
    \item \textbf{Draft Construction}: It employs the draft model to generate several subsequent tokens, namely draft tokens, in parallel or in the auto-regressive manner. 
    \item \textbf{Draft Verification}: It employs the target model to compute the conditional probabilities of all the draft tokens in a single LLM inference step, subsequently determining the acceptance of each draft token sequentially. The acceptance rate, representing the average number of accepted draft tokens per inference step, serves as a key metric for evaluating the performance of a speculative decoding algorithm.
\end{enumerate}

Speculative decoding ensures output equivalence with standard auto-regressive decoding methods. Traditional decoding techniques typically employ two primary sampling strategies: greedy sampling and nucleus sampling. Greedy sampling involves selecting the token with the highest probability at each decoding step to generate a specific output sequence. The initial attempt at speculative decoding, known as Blockwise Parallel Decoding~\cite{stern2018blockwise}, aims to ensure that the draft tokens precisely match the tokens sampled via greedy sampling, thus preserving output token equivalence. In contrast, nucleus sampling involves sampling tokens from a probability distribution, resulting in diverse token sequences with each run. This diversity makes nucleus sampling popular. To accommodate nucleus sampling within speculative decoding frameworks, speculative sampling techniques~\cite{leviathan2023fast,chen2023accelerating} have been proposed. Speculative sampling maintains output distribution equivalence, aligning with the probabilistic nature of nucleus sampling to generate varied token sequences. 
Formally, given a sequence of tokens $x_1, x_2, ..., x_n$ and a sequence of draft tokens $\hat{x}_{n+1}, \hat{x}_{n+2}, ..., \hat{x}_{n+k}$, the speculative sampling strategy accepts the $i$-th draft token with the following probabilities: 
\begin{equation}
    \min \left( 1, \frac{p(\hat{x}_i|x_1, x_2, ..., x_{i-1})}{q(\hat{x}_i|x_1, x_2, ..., x_{i-1})} \right), 
\end{equation}
where $p(\cdot|\cdot)$ and $q(\cdot|\cdot)$ denote the conditional probabilities from the target LLM and the draft model, respectively. If the $i$-th draft token is accepted, it sets $x_i\xleftarrow{}\hat{x}_i$. Otherwise, it quits the verification of the following draft tokens, and resamples $x_i$ from the following distribution:
\begin{equation}
    {\rm norm}(\max(0, p(\cdot|x_1, x_2, ..., x_{i-1})-q(\cdot|x_1, x_2, ..., x_{i-1}) )).
\end{equation}
Building upon speculative sampling, several variants~\cite{sun2023spectr,miao2023specinfer} have emerged, aimed at validating multiple draft token sequences. Notably, the token tree verifier~\cite{miao2023specinfer} has become a widely adopted verification strategy within this context. This approach utilizes a tree-structured representation of draft token sets and employs a tree attention mechanism to efficiently perform the verification process. 

In the speculative decoding approach, the acceptance rate of draft tokens is significantly influenced by the degree to which the output distributions of draft models align with those of original LLMs. As a result, considerable research efforts have been directed towards improving the design of draft models. 
DistillSpec~\cite{zhou2023distillspec} directly distills a smaller draft model from the target LLM. 
SSD~\cite{zhang2023draft} involves automatically identifying a sub-model (a subset of model layers) from the target LLM to serve as the draft model, eliminating the need for separate training of the draft model. 
OSD~\cite{liu2023online} dynamically adjusts the output distribution of the draft model to match the user query distribution in online LLM services. It achieves this by monitoring rejected draft tokens from the LLM and using this data to refine the draft model through distillation. 
PaSS~\cite{monea2023pass} proposes utilizing the target LLM itself as the draft model, incorporating trainable tokens (look-ahead tokens) into the input sequence to enable simultaneous generation of subsequent tokens. 
REST~\cite{he2023rest} introduces a retrieval-based speculative decoding approach, employing a non-parametric retrieval data store as the draft model. 
SpecInfer~\cite{miao2023specinfer} introduces a collective boost-tuning technique to align the output distribution of a group of draft models with that of the target LLM. 
Lookahead decoding~\cite{fu2023lookahead} involves generating n-grams of the target LLM in parallel to aid in generating draft tokens. 
Medusa~\cite{cai2024medusa} fine-tunes several heads of the LLM specifically for generating subsequent draft tokens. 
Eagle~\cite{li2023eagle} adopts a lightweight transformer layer called an auto-regression head to generate draft tokens in an auto-regressive manner, integrating rich contextual features from the target LLM into the draft model's input. 
Kangaroo~\cite{liu2024kangaroo} uses a fixed shallow sub-network as the draft model, and trains a lightweight adapter on the top of the sub-network. In this way, it does not need to train a separate draft model. 

Another line of studies focuses on designing more effective draft construction strategies. 
Conventional approaches often yield single draft token sequences, posing challenges for passing verification. In response, Spectr~\cite{sun2023spectr} advocates generating multiple draft token sequences and employs a $k$-sequential draft selection technique to concurrently verify $k$ sequences. This method leverages speculative sampling, ensuring equivalence in output distributions. 
Similarly, SpecInfer~\cite{miao2023specinfer} adopts a comparable approach. However, unlike Spectr, SpecInfer merges draft token sequences into a ``token tree'' and introduces a tree attention mechanism for validation. This strategy is called the "token tree verifier". Due to its efficacy, token tree verifier has been widely embraced in numerous speculative decoding algorithms~\cite{spector2023accelerating,he2023rest,cai2024medusa,li2023eagle}. 
In addition to these efforts, Stage Speculative Decoding~\cite{spector2023accelerating} and Cascade Speculative Drafting (CS Drafting)~\cite{chen2023cascade} propose accelerating draft construction by integrating speculative decoding directly into the token generation process.

\noindent \textbf{Comparative Experiments and Analysis.} 
We conduct an experiment to evaluate the speed-up performance of the speculative decoding methods. Specifically, we thoroughly review the studies of this field, and select six of them that have open-sourced their codes, i.e., Speculative Decoding (SpD)~\cite{leviathan2023fast,chen2023accelerating}, Lookahead Decoding (LADE)~\cite{fu2023lookahead}, REST~\cite{he2023rest}, Self-speculative Decoding (SSD)~\cite{zhang2023draft}, Medusa~\cite{cai2024medusa} and Eagle~\cite{li2023eagle}. 
As for the evaluation dataset, we use Vicuna-80~\cite{chiang2023vicuna} to evaluate the above methods, which contains 80 questions that classified into 10 categories. We report the average results on these 80 questions. 
As for target LLMs, we adopt five fashion open-source LLMs, i.e., Vicuna-7B-V1.3~\cite{chiang2023vicuna}, Vicuna-13B-V1.3~\cite{chiang2023vicuna}, Vicuna-33B-V1.3~\cite{chiang2023vicuna}, LLaMA-2-7B~\cite{touvron2023llama} and LLaMA-2-13B~\cite{touvron2023llama}. We report the range of evaluation metrics across these 5 LLMs. 
As for draft models, we adopt two well-trained draft models, i.e., LLaMA-68M and LLaMA-160M~\cite{miao2023specinfer} for SpD. For other speculative decoding methods, we follow their proposed draft construction approach and use the checkpoints they provide. 
As for the evaluation metrics, we adopt \textit{acceptance rate}, which denotes the ratio of the number of accepted tokens to the number of generation steps, and \textit{ speed-up}, which denotes the ratio of the latency of original auto-regressive decoding to the latency of speculative decoding when fixing the total length of output. 

Tab.~\ref{tab:spec_comparison} provides a comparison of various speculative decoding methods, highlighting several key observations: (1) Eagle demonstrates exceptional performance, achieving a notable 3.47$\sim$3.72$\times$ end-to-end speed-up across multiple LLMs. To understand its success, a deeper analysis of Eagle reveals two key factors. Firstly, Eagle employs an auto-regressive approach for decoding draft tokens, leveraging information from previously generated tokens directly. Secondly, Eagle integrates rich features from previous tokens of both original LLMs and draft models to enhance the accuracy of the next draft token generation. (2) The token tree verifier proves to be an effective technique in enhancing the performance of speculative decoding methods. (3) The end-to-end speed-up achieved by these methods is often lower than the acceptance rate. This difference arises due to the practical consideration that the generation cost associated with draft models cannot be overlooked. 

\subsubsection{Offloading}
% FlexGen, Powerinfer, llama.cpp, LM-Offload, (FastDecode)
%\revise{[words polished.]}
Current research investigates the potential of offloading to accommodate the substantial memory demand of LLMs (see Sec.~\ref{sec:efficiency}) in resource-constrained environments. The essence of offloading is to offload part of the storage from the GPU to the CPU when it is free of use. Intuitively, the focus of such kind of research lies in hiding the expensive data movement latency between the GPU and the CPU. 
FlexGen~\cite{sheng2023flexgen} enables the offloading of weights, activations, and the KV cache, and further formulates a graph traversal problem for offloading to maximize the throughput. 
% The weights of the next layer, KV cache/activation load of the next batch, KV cache/activation store of the previous batch, and the computation of the current batch are overlapped.
The data loading of the next batch and the data storing of the previous batch can be overlapped with the computation of the current batch.
Another work llama.cpp~\cite{ggerganov2024llamacpp} also assigns computational tasks to the CPU, mitigating the data transfer overhead at the cost of computing with the low-powered CPU. Powerinfer~\cite{song2023powerinfer} exploits the sparsity in activations using ReLU~\cite{ReLU} in LLMs, and divides the activations into subsets of cold and hot neurons representing the frequency of computation. 
The cold neurons are offloaded to the CPU for both storage and computation in Powerinfer. Leveraging adaptive predictors and sparse operators, Powerinfer significantly improves the computational efficiency with offloading. 
FastDecode~\cite{he2024fastdecode} proposes to offload the storage and the computation of the entire attention operator to the CPU. Since the attention operation is computed on the CPU, the data movement of KV cache is reduced to merely some activations. The number of CPUs is selected to match the workload latency on GPUs so that the bubbles in the heterogeneous pipeline are mitigated. 
 
\subsection{Serving System} % @hongke [0210: v1 done]
% add a table of open-source serving frameworks --> QPS
\label{sec:serving_system}

The optimizations for serving systemworks are dedicated to improving the efficiency in handling asynchronous requests. The memory management is optimized to hold more requests, and efficient batching and scheduling strategies are integrated to enhance the system throughput. Besides, optimizations specific to distributed systems are proposed to exploit distributed computational resources. 

\begin{figure*}[h]
\centering
\tikzset{
    basic/.style  = {draw, text width=2cm, align=center, font=\sffamily, rectangle},
    root/.style   = {basic, rounded corners=2pt, thin, align=center, fill=white,text width=8cm, rotate=90, font=\footnotesize},
    dnode/.style = {basic, thin, rounded corners=2pt, align=center, fill=yellow!30,text width=3.5cm, font=\footnotesize},
    dnode_1/.style = {basic, thin, rounded corners=2pt, align=center, fill=yellow!30,text width=2cm, font=\footnotesize},
    mnode/.style = {basic, thin, rounded corners=2pt, align=center, fill=blue!10,text width=3.5cm, font=\footnotesize},
    mnode_1/.style = {basic, thin, rounded corners=2pt, align=center, fill=blue!10,text width=2cm, font=\footnotesize}, 
    snode/.style = {basic, thin, rounded corners=2pt, align=center, fill=npurple,text width=3.5cm, font=\footnotesize},
    snode_1/.style = {basic, thin, rounded corners=2pt, align=center, fill=npurple,text width=2cm, font=\footnotesize},
    tnode/.style = {basic, thin, align=left, fill=pink!60, text width=15em, align=center},
    xnode/.style = {basic, thin, rounded corners=2pt, align=center, fill=blue!20,text width=5cm,},
    wnode/.style = {basic, thin, rounded corners=2pt, align=left, fill=white,text width=5.8cm, font=\footnotesize},
    wnode_1/.style = {basic, thin, rounded corners=2pt, align=left, fill=white,text width=3.3cm, font=\footnotesize},
    wnode_2/.style = {basic, thin, rounded corners=2pt, align=left, fill=white,text width=6cm, font=\footnotesize},
    %edge from parent/.style = {draw=black, edge from parent fork right}
    %edge from parent/.style = {draw=black, edge from parent fork down}
}
\begin{forest} 
for tree={
    grow=east,
    growth parent anchor=east,
    parent anchor=east,
    child anchor=west,
    edge path={\noexpand\path[\forestoption{edge},->, >={latex}] 
         (!u.parent anchor) -- +(5pt,0pt) |- (.child anchor)
         \forestoption{edge label};}
}
% l sep is used for arrow distance
[Serving System, snode
    [Distributed Systems, snode
        [{Splitwise~\cite{patel2023splitwise}, 
        TetriInfer~\cite{hu2024tetriinfer}, 
        DistServe~\cite{zhong2024distserve},
        SpotServe~\cite{miao2023spotserve},
        Infinite-LLM~\cite{lin2024infinitellm}}, wnode_2]
    ]
    [Scheduling, snode
        [{ORCA~\cite{yu2022orca}, vLLM~\cite{kwon2023vllm}, LightLLM~\cite{modeltc2024lightllm}, DeepSpeed-FastGen~\cite{holmes2024deepspeedfastgen}, FastServe~\cite{wu2023fastserve}, VTC~\cite{sheng2024vtc}}, 
        wnode_2]
    ]
    [Batching, snode
        [{ORCA~\cite{yu2022orca}, vLLM~\cite{kwon2023vllm}, Sarathi~\cite{agrawal2023sarathi}, DeepSpeed-FastGen~\cite{holmes2024deepspeedfastgen}, 
        Sarathi-Serve~\cite{agrawal2024sarathiserve}, 
        LightLLM~\cite{modeltc2024lightllm}}, wnode_2]
    ]
    [Memory Management, snode
        [{S$^3$~\cite{jin2023s3}, vLLM~\cite{kwon2023vllm}, LightLLM~\cite{modeltc2024lightllm}, FlashInfer~\cite{ye2024flashinfer}}, wnode_2]
    ]
]
\end{forest}

\caption{Taxonomy of the optimization for LLM serving system.}
\label{fig:serving_framework}
\end{figure*}
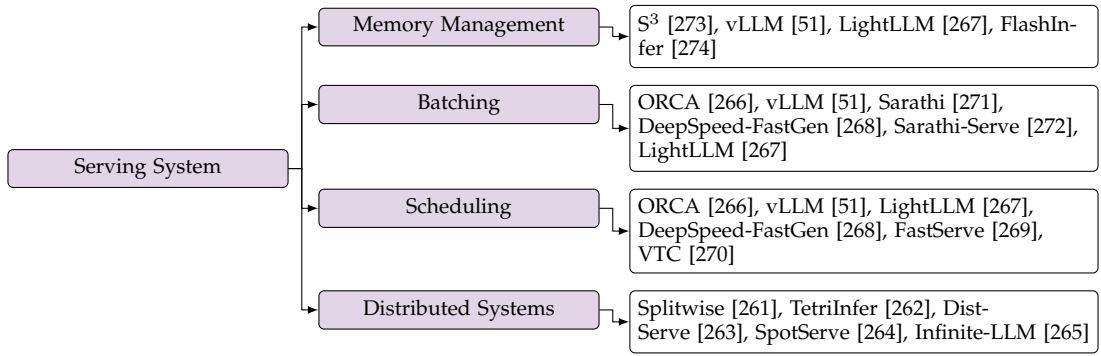

\subsubsection{Memory Management}
% vLLM, LightLLM, S^3, Pensieve
The storage of KV cache dominates the memory usage in LLM serving,  especially when the context length is long (see Sec.~\ref{sec:efficiency}). Since the generation length is uncertain, it is challenging to allocate the space for KV cache storage in advance. Earlier implementations~\cite{FasterTransformer} usually allocate storage space in advance based on the preset maximum length of each request. However, in instances where request generation is terminated early, this approach incurs significant wastage of storage resources. To address the issue, S$^3$~\cite{jin2023s3} proposes to predict an upper bound of the generation length for each request, in order to reduce the waste of the pre-allocated space. However, the static way of KV cache memory allocation still fails when no such large contiguous space exists. To deal with the fragmented storage, vLLM~\cite{kwon2023vllm} proposes to store the KV cache in a paged manner following the operating system. vLLM first allocates a memory space as large as possible and divides it equally into multiple physical blocks. When a request comes, vLLM dynamically maps the generated KV cache to the pre-allocated physical blocks in a discontinuous fashion. In this way, vLLM significantly reduces storage fragmentation and achieves a higher throughput in LLM serving. On the basis of vLLM, LightLLM~\cite{modeltc2024lightllm} uses a more fine-grained KV cache storage to cut down the waste happening with the irregular boundary. Instead of a block, LightLLM treats the KV cache of a token as a unit, so that the generated KV cache always saturates the pre-allocated space.

Current optimized service systems commonly employ this paged approach to manage the KV cache storage, thereby mitigating the waste of redundant KV cache memory. However, the paged storage leads to irregular memory access in the attention operator. For the attention operator using the paged KV cache, this necessitates the consideration of the mapping relationship between the virtual address space of the KV cache and its corresponding physical address space. To enhance the efficiency of the attention operator, the loading pattern of the KV cache must be tailored to facilitate contiguous memory access. For instance, in the case of the PagedAttention by vLLM~\cite{kwon2023vllm}, the storage of the head size dimension is structured as a 16-byte contiguous vector for K cache, while FlashInfer~\cite{ye2024flashinfer} orchestrates diverse data layouts for the KV cache, accompanied by an appropriately designed memory access scheme. The optimization of the attention operator in conjunction with paged KV cache storage remains a forefront challenge in the advancement of serving systems.

\subsubsection{Continuous Batching}
The request lengths in a batch can be different, leading to low utilization when shorter requests are finished and longer requests are still running. Due to the asynchronous nature of requests in serving scenarios, there exists an opportunity that such periods of low utilization could be mitigated. The continuous batching technique is proposed to leverage the opportunity by batching new requests once some old requests are finished. ORCA~\cite{yu2022orca} is the first to utilize the continuous batching technique in LLM serving. The computation of each request encompasses multiple iterations, with each iteration representing either a prefilling step or a decoding step. The author suggests that different requests can be batched at the iteration level. The work implements iteration-level batching in linear operators, concatenating different requests together in the sequence dimension. Hence, the spare storage and computational resources corresponding to the completed requests are promptly released. Following ORCA, vLLM~\cite{kwon2023vllm} extends the technique to the attention computation, enabling requests with different KV cache lengths to be batched together. Sarathi~\cite{agrawal2023sarathi}, DeepSpeed-FastGen~\cite{holmes2024deepspeedfastgen} and Sarathi-Serve~\cite{agrawal2024sarathiserve} further introduce a split-and-fuse method to batch together prefilling requests and decoding requests. Specifically, this method first splits the long prefilling request in the sequence dimension, and then batches it together with multiple short decoding requests. The split-and-fuse method balances the workloads among different iterations, and significantly reduces the tail latency via removing the stalls from new requests. LightLLM~\cite{modeltc2024lightllm} also adopts the split-and-fuse method.

The split-and-fuse technology operates on the premise that requests during the prefilling stage can be partitioned into discrete chunks. Chunked-prefill methodology involves segmenting prefilling requests along the sequence dimension, thereby preventing the potential bottlenecks for other requests. This strategy capitalizes on the auto-regressive characteristics inherent in LLMs, where attention computation only relies on prior tokens. Consequently, the mathematical equivalence of chunked-prefill technology is guaranteed, positioning it as a leading approach for reducing request latency in LLM serving.

\begin{table*}[tb]
\caption{Comparison of multiple open-source inference engines and serving systems. "-" denotes no serving support. Note that the scheduling method of TensorRT-LLM is not open-sourced.}
\label{tab:serving}
\begin{center}
\resizebox{0.95\textwidth}{!}
{
\begin{tabular}{c|cccc|c|ccc|c}%cp{1.6cm}<{\centering}p{1.6cm}<{\centering}p{1.6cm}<{\centering}p{1.6cm}<{\centering}p{1.6cm}<{\centering}}
\toprule

\multirow{2}{*}[-1ex]{Model} & \multicolumn{4}{c|}{Inference Optimization} & \multirow{2}{*}[-0.5ex]{\makecell[c]{Inference \\ (token/s)}} & \multicolumn{3}{c|}{Serving Optimization} & \multirow{2}{*}[-0.5ex]{\makecell[c]{Serving \\ (req/s)}} \\
\cmidrule(lr){2-5} \cmidrule(lr){7-9}
 & Attention & Linear & Graph & Speculative Decoding & & Memory & Batching & Scheduling & \\
\midrule

HuggingFace~\cite{huggingface2024transformers} & & & & \checkmark & 38.963 & - & - & -
 & - \\
DeepSpeed~\cite{deepspeed} & \checkmark & & \checkmark & & 80.947 & blocked & split-and-fuse & decode prioritized 
 & 6.78 \\
vLLM~\cite{kwon2023vllm} & \checkmark & & & \checkmark & 90.052 & paged & continuous batching & prefill prioritized & 7.11 \\
OpenPPL~\cite{OpenPPL} & \checkmark & & \checkmark & & 81.169 & - & - & - & - \\
FlashDecoding++~\cite{hong2024flashdecoding} & \checkmark & \checkmark & \checkmark & & 106.636 & - & - & - & - \\
LightLLM~\cite{modeltc2024lightllm} & \checkmark & & & & 73.599 & token-wise & split-and-fuse & prefill prioritized & 10.29\\
TensorRT-LLM~\cite{tensorrt-llm} & \checkmark & \checkmark & \checkmark & \checkmark & 92.512 & paged & continuous batching & - & 5.87\\

\bottomrule
\end{tabular}
}
\end{center}
\end{table*}

\subsubsection{Scheduling Strategy}
In LLM serving, the job length of each request exhibits variability, and hence the order of executing requests significantly impacts the throughput of the serving system. The head-of-line blocking~\cite{wu2023fastserve} happens when long requests are accorded priority. Specifically, memory usage grows rapidly in response to long requests, impeding subsequent requests when the system exhausts its memory capacity.
The pioneering work ORCA~\cite{yu2022orca} and open-source systems, including vLLM~\cite{kwon2023vllm} and LightLLM~\cite{modeltc2024lightllm}, employ the simple first-come-first-serve (FCFS) principle to schedule requests. DeepSpeed-FastGen~\cite{holmes2024deepspeedfastgen} gives priority to the decoding requests to enhance the performance. FastServe~\cite{wu2023fastserve} proposes a preemptive scheduling strategy to optimize the head-of-line blocking problem, achieving low job completion time (JCT) in LLM serving. FastServe employs a multi-level feedback queue (MLFQ) to prioritize the requests with the shortest remaining time. Since the auto-regressive decoding approach poses unknown request lengths, FastServe predicts the length first and utilizes a skip-join fashion to find the proper priority for each request. Unlike previous work, VTC~\cite{sheng2024vtc} discusses the fairness in LLM serving. VTC introduces a cost function based on token numbers to measure fairness among clients, and further proposes a fair scheduler to ensure fairness.

\subsubsection{Distributed Systems}
% 增加了ExeGPT, Llumnix, LoongServe
In order to achieve high throughput, LLM services are commonly deployed on distributed platforms. Recent works have additionally focused on optimizing the performance of such inference services by exploiting distributed characteristics. Notably, observing that the computations of prefilling and decoding have interference with each other, splitwise~\cite{patel2023splitwise}, TetriInfer~\cite{hu2024tetriinfer} and DistServe~\cite{zhong2024distserve} demonstrate the efficiency of disaggregating the prefilling and the decoding steps of a request. In this way, the two distinct steps are processed independently based on their characteristics. ExeGPT~\cite{oh2024exegpt} also adopts such disaggregated architecture, and proposes different strategies with controllable variables to maximize system throughput under certain latency constraints. Llumnix~\cite{sun2024llumnix} reschedules the requests at runtime for different serving objectives including load balancing, de-fragmentation, and prioritization. SpotServe~\cite{miao2023spotserve} is designed to provide LLM service on clouds with preemptible GPU instances. SpotServe efficiently handles challenges including dynamic parallel control and instance migration, and also utilizes the auto-regressive nature of LLMs to achieve token-level state recovery. Moreover, Infinite-LLM~\cite{lin2024infinitellm} parallels different parts of the sequence in the attention operator across the data center, to address the challenges when serving extremely long contexts. LoongServe~\cite{wu2024loongserve} proposes the elastic sequence parallelism to manage the elastic resource demand at the iteration level, reducing the data movement of KV cache via elaborately designed scheduling.

\subsection{Hardware Accelerator Design} 
\label{sec:hardware}
Previous research efforts~\cite{li2020ftrans, ham2021elsa, fan2022butterfly} have focused on optimizing Transformer architectures, particularly enhancing the attention operator, often employing sparse methods to facilitate FPGA deployment. The FACT~\cite{qin2023fact} accelerator achieves superior energy efficiency compared to the NVIDIA V100 GPU through mixed-precision quantization for linear operators and algorithm-hardware co-design, yet these approaches are not tailored for generative LLMs. 

Recent work like ALLO~\cite{chen2023allo} highlights FPGA advantages in managing the memory-intensive decoding stage and emphasizes the importance of model compression techniques for LLMs' efficient FPGA deployment. Conversely, DFX~\cite{hong2022dfx} focuses on decoding stage optimizations but lacks model compression methods, limiting scalability to larger models and longer inputs (up to 1.5B model and 256 tokens). ALLO builds on these insights, further offering a library of High-level Synthesis (HLS) kernels that are composable and reusable. ALLO's implementation demonstrates superior generation speed-up compared to DFX in the prefilling stage, achieving enhanced energy efficiency and speedup over the NVIDIA A100 GPU during decoding. 

FlightLLM~\cite{flightllm} also leverages these insights, introducing a configurable sparse digital signal processor (DSP) chain for various sparsity patterns with high computational efficiency. It proposes an always-on-chip decode scheme with mixed-precision support to enhance memory bandwidth utilization. FlightLLM achieves 6.0$\times$ higher energy efficiency and 1.8$\times$ better cost efficiency than the NVIDIA V100S GPU for Llama2-7B models, with 1.2$\times$ higher throughput than the NVIDIA A100 GPU during decoding. 

\subsection{Comparison of LLM Frameworks}
We compare the performance of multiple LLM frameworks in Table~\ref{tab:serving}. The inference throughput is measured with Llama2-7B (batch size=1, input length=1k, output length=128). The serving performance is the maximum throughput measured on the ShareGPT~\cite{sharegpt2023sharegpt} dataset. Both are derived on a single NVIDIA A100 80GB GPU. Among the mentioned frameworks, DeepSpeed~\cite{deepspeed}, vLLM~\cite{kwon2023vllm}, LightLLM~\cite{modeltc2024lightllm} and TensorRT-LLM~\cite{tensorrt-llm} integrate the serving function to serve asynchronous requests from multiple users. We also list the optimizations for each framework in the table. All the frameworks except HuggingFace implement operator-level or graph-level optimizations to enhance performance, and some of them also support the speculative decoding technique. Note that the speculative decoding technique is off when we measure the inference performance for all frameworks. The results of inference throughput show that FlashDecoding++ and TensorRT-LLM outperform others with optimizations covering predominant operators and the computational graph. From the aspect of serving, all the frameworks use fine-grained and discontiguous storage for KV cache, and apply the continuous batching techniques to improve the system utilization. Unlike vLLM and LightLLM, DeepSpeed prioritizes the decoding requests in scheduling, which means no new request is merged if there are enough existing decoding requests in the batch.

\subsection{Knowledge, Suggestions and Future Direction}
\label{sec:system_summary}

The system-level optimization improves efficiency while bringing no accuracy degradation, hence becoming prevalent in the LLM inference practice. The optimization for inference is also applicable to serving. Recently, the operator optimization has been closely combined with practical serving scenarios, e.g.,, RadixAttention~\cite{zheng2023efficiently} designed specifically for prefix caching, and tree attention ~\cite{miao2023specinfer} to accelerate speculative decoding verification. The iterating of applications and scenarios will continue to put forward new requirements for operator development.

Given the multifaceted objectives inherent in real-world serving systems, such as JCT, system throughput, and fairness, the design of scheduling strategies becomes correspondingly intricate. Within the domain of LLM serving, where the length of requests is indeterminate, extant literature commonly relies on predictive mechanisms to facilitate the design of scheduling strategies. However, the efficacy of current predictors~\cite{hu2024tetriinfer} falls short of ideal standards, indicating the potential for refinement and optimization in serving scheduling strategy development.

\section{Discussions of Key Application Scenarios}
\label{sec:discussion}

Current research endeavors have made significant strides in exploring the boundaries of efficient LLM inference across various optimization levels. However, further studies are warranted to enhance LLM efficiency in practical scenarios. We have provided promising future directions for optimization techniques at the data-level (Sec.~\ref{sec:data_summary}), model-level (Sec.~\ref{sec:model_summary}), and system-level (Sec.~\ref{sec:system_summary}). In this section, we summarize four critical scenarios: agent and multi-model framework, long-context LLMs, edge scenario deployment, and security-efficiency synergy, and provide a broader discussion on them. 
%We hope this discussion can bring some insights for the future studies for the efficiency optimization of LLMs. 

%\subsection{Long-Context LLMs}

\noindent \textbf{Agent and Multi-Model Framework.} As discussed in Sec.~\ref{sec:data_summary}, recent advancements in agent and multi-model frameworks~\cite{xi2023rise,sun2023corex,guo2024large} have significantly improved agents' capabilities to handle complex tasks and human requests by harnessing the powerful abilities of LLMs. These frameworks, while increasing the computational demands of LLMs, introduce more parallelism into the structure of LLMs' output content, thereby creating opportunities for data-level and system-level optimizations such as output organization techniques~\cite{zheng2023efficiently}. Furthermore, these frameworks naturally introduce a new optimization level, i.e., pipeline-level, which holds potential for efficiency enhancements at this level \cite{chen2023frugalgpt}.

In addition, there is a growing research trend~\cite{xie2024large} focused on extending AI agents into the multimodal domain, which often utilize Large Multimodal Models (LMMs) as the core of these agent systems. To enhance the efficiency of these emerging LMM-based agents, designing optimization techniques for LMMs is a promising research direction.

\noindent\textbf{Long-Context LLMs.} Currently, LLMs face the challenge of handling increasingly longer input contexts. However, the self-attention operation, the fundamental component of Transformer-style LLMs, exhibits quadratic complexity in relation to the context length, imposing constraints on maximum context length during both training and inference phases. Various strategies have been explored to address this limitation, including input compression (Sec.~\ref{sec:input_compress}), sparse attention (Sec.~\ref{sec:sparse}), design of low-complexity structures (Sec.~\ref{sec:alternate}), and optimization of attention operators (Sec.~\ref{sec:graph_operator_optimize}). Notably, non-Transformer architectures (Sec.~\ref{sec:alternate}) with sub-quadratic or linear complexity have recently garnered significant interest from researchers. 

Despite their efficiency, the competitiveness of these novel architectures compared to the Transformer architecture across various abilities, such as in-context learning ability and long-range modeling ability, is still under scrutiny~\cite{park2024can,lee2023exploring}. Therefore, exploring the capabilities of these new architectures from multiple angles and addressing their limitations remains a valuable pursuit. Moreover, it is crucial to determine the necessary context lengths for various scenarios and tasks, as well as identify the next-generation architecture that will serve as the foundational backbone for LLMs in the future.

% Additionally, attention operator optimization techniques (Section~\ref{sec:graph_operator_optimize}) aim to reduce memory access costs through kernel fusing or enhance GPU utilization via resource scheduling during attention computation. These methods can ensure model output equivalence and preserve performance integrity. 

\noindent\textbf{Edge Scenario Deployment.} 
While considerable efforts have been directed towards enhancing the efficiency of LLM inference, deploying LLMs onto extremely resource-constrained edge devices like mobile phones presents ongoing challenges. Recently, numerous researchers~\cite{biderman2023pythia,bai2023qwen,tang2024rethinking,li2023textbooks,gunasekar2023textbooks,zhang2024tinyllama,zhang2023towards,openlm2023openllama,bellagente2024stable,minicpm2024,liu2024mobilellm} have shown interest in pre-training smaller language models with 1B to 3B parameters. Models of this scale offer reduced resource costs during inference and hold potential for achieving generalization abilities and competitive performance compared to larger models. However, the methods to develop such efficient and powerful smaller language models remain under-explored. 

Several studies have initiated this promising direction. For instance, MiniCPM~\cite{minicpm2024} conducts sandbox experiments to determine optimal pre-training hyper-parameters. PanGu-$\pi$-Pro~\cite{tang2024rethinking} suggests initializing model weights from pre-trained LLMs using metrics and techniques from model pruning. MobileLLM~\cite{liu2024mobilellm} adopts a``deep and thin'' architecture for small model design and proposes weight sharing across different layers to increase the number of layers without additional memory costs. Nevertheless, a performance gap still exists between small and large models, necessitating future studies to narrow this gap. In the future, there is a crucial need for research aimed at identifying the model scale limited in the edge scenarios, and exploring the boundaries of various optimization methods on designing smaller models. 

Beyond designing smaller models, system-level optimization offers a promising direction in LLM deployment. A notable recent project, MLC-LLM~\cite{mlc-llm}, successfully deploys the LLaMA-7B model on mobile phones. MLC-LLM primarily employs compilation techniques like fusion, memory planning, and loop optimization to enhance latency and reduce memory cost during inference. Additionally, adopting the cloud-edge collaboration techniques, or designing more sophisticated hardware accelerators can also help deploy LLMs onto edge devices. 

\noindent \textbf{Security-Efficiency Synergy.} In addition to task performance and efficiency, security is also a crucial factor that must be considered in LLM applications~\cite{yao2024survey,li2024personal}. Current research primarily focuses on efficiency optimization without adequately addressing security considerations. Therefore, it is critical to investigate the interplay between efficiency and security and determine whether the current optimization techniques compromise the security of LLMs. If these techniques negatively impacts LLMs' security, a promising direction would involve developing new optimization methods or refining the existing ones to achieve a better trade-off between LLMs' efficiency and security.

\section{Conclusion}
\label{sec:conclusion}

% In this survey, we provide a systematic review of efficient LLMs, an important area of research
% aimed at democratizing LLMs. We start with motivating the necessity for efficient LLMs. Guided
% by a taxonomy, we review algorithm-level and system-level efficient techniques for LLMs from
% model-centric and data-centric perspectives respectively. Furthermore, we review LLM frameworks
% with specific optimizations and features crucial for efficient LLMs. We believe that efficiency will
% play an increasingly important role in LLMs and LLMs-oriented systems. We hope this survey
% could enable researchers and practitioners to quickly get started in this field and act as a catalyst to
% inspire new research on efficient LLMs.

Efficient LLM inference focuses on reducing the computational, memory access, and memory costs during LLM inference processes, aiming to optimize efficiency metrics such as latency, throughput, storage, power, and energy. This survey offers a comprehensive review of efficient LLM inference research, presenting insights, recommendations, and future directions for key techniques. Initially, we introduce a hierarchical taxonomy encompassing data-, model-, and system-level optimizations. Subsequently, guided by this taxonomy, we meticulously examine and summarize studies at each level and sub-field. For well-established techniques like model quantization and efficient serving systems, we conduct experiments to evaluate and analyze their performance. Based on these analyses, we offer practical suggestions and identify promising research avenues for practitioners and researchers in the field.

\section*{Acknowledgements}
This work was supported by National Natural Science Foundation of China (No. 62325405, 62104128, U19B2019, U21B2031, 61832007, 62204164), Tsinghua EE Xilinx AI Research Fund, and Beijing National Research Center for Information Science and Technology (BNRist). We thank for all the support from Infinigence-AI. 
We thank Xiangsheng Shi, Zinan Lin, Xinhao Yang, Hongyi Wang, Linfeng Zhang, Yulin Wang, Xuemin Sun, Saiqian Zhang for their valuable suggestions on the paper. We thank Shengxiang Wang, Qiuli Mao for providing the efficiency profiling data of quantized operators. 
%We thank ... @hk

\bibliography{
top,
refs/intro,
refs/data_level,
refs/spec,refs/kd,
refs/structure_optim,
refs/moe,
refs/quantization,
refs/attention,
refs/operator,
refs/serving,
refs/alternate,
refs/sparse,
refs/discussion
}
\bibliographystyle{IEEEtran}

% \appendix
% \section{Appendix}

\end{document}